\documentclass{article}%

 \pdfoutput=1
\usepackage{graphicx}
\usepackage{psfrag}
\usepackage{subfigure}
\usepackage{url}
\usepackage{stfloats}
\usepackage{amsmath}
\usepackage{amsfonts}
\usepackage{amssymb}
\usepackage{graphicx}
\usepackage{subfigure}
\usepackage{epstopdf}
\usepackage{vmargin}
\usepackage{tikz}
\usepackage{amsfonts}

\graphicspath{{./}{./figures/}}
\tikzstyle{bluecircle}=[circle,
minimum size=.05cm,
draw=blue!100,
fill=blue!100,
]
\tikzstyle{redcircle}=[circle,
minimum size=.1cm,
draw=red!100,
fill=red!100,
font=\small
]
\tikzstyle{greencircle}=[circle,
minimum size=.01cm,
draw=green!100,
fill=green!100,
font=\small
]
\tikzstyle{blackcircle}=[circle,
minimum size=.01cm,
draw=black!100,
fill=black!100,
font=\small
]

\graphicspath{{./}{./figures/}}

\newcommand{\new}{}

\providecommand{\U}[1]{\protect\rule{.1in}{.1in}}
\setmarginsrb{2.5cm}{2cm}{2.5cm}{3cm}{0cm}{0cm}{0cm}{1cm}
\graphicspath{{./}{./figures/}}
\tikzstyle{bluecircle}=[circle,
minimum size=.05cm,
draw=blue!100,
fill=blue!100,
]
\tikzstyle{redcircle}=[circle,
minimum size=.1cm,
draw=red!100,
fill=red!100,
font=\small
]
\tikzstyle{greencircle}=[circle,
minimum size=.01cm,
draw=green!100,
fill=green!100,
font=\small
]
\tikzstyle{blackcircle}=[circle,
minimum size=.01cm,
draw=black!100,
fill=black!100,
font=\small
]
\begin{document}

\title{Bayesian nonparametric image segmentation using a generalized Swendsen-Wang algorithm}
\author{Richard Xu$^{1}$, Fran\c{c}ois Caron$^{2}$, Arnaud Doucet$^{2}$\\$^{1}$Faculty of Engineering and Information Technology, \\University of Technology of Sydney, Australia.\\$^{2}$Department of Statistics, Oxford University, United Kingdom}
\maketitle

\begin{abstract}
 Unsupervised image segmentation aims at clustering the set of
pixels of an image into spatially homogeneous regions. We introduce here
a class of Bayesian nonparametric models to address this problem.
These models are based on a combination of a Potts-like spatial smoothness
component and a prior on partitions which is used to control both
the number and size of clusters. This class of models is flexible
enough to include the standard Potts model and the more recent Potts-Dirichlet
Process model \cite{Orbanz2008}. More importantly, any prior on partitions
can be introduced to control the global clustering structure so that
it is possible to penalize small or large clusters if necessary. Bayesian
computation is carried out using an original generalized Swendsen-Wang
algorithm. Experiments demonstrate that our method is competitive
in terms of RAND\ index compared to popular image segmentation methods,
such as mean-shift, and recent alternative Bayesian nonparametric
models.
\end{abstract}

\section{Introduction\label{sec:intro}}

Sophisticated statistical models were introduced early on to address
unsupervised image segmentation tasks. The seminal 1984 paper of \cite{Geman1984} popularized Ising-Potts models and more
generally Markov Random Fields (MRF) as well as Markov chain Monte
Carlo (MCMC) methods in this area.\ There has been an increasing
interest in such approaches ever since \cite{Higdon1998,Winkler2003,Barbu2005,Barbu2007}.
A key problem of MRF type approach is that they typically require
specifying the number of clusters beforehand. It is easy conceptually
to assign a prior to this number but then Bayesian inference becomes
computationally demanding as the partition function of the MRF\ is
analytically intractable \cite{Green2002}. It is additionally difficult
to design efficient reversible jump MCMC algorithms in this context
\cite{Green1995}. Recently, a few Bayesian nonparametric (BNP) models
for image segmentation have been proposed which bypass these problems\ \cite{Duan2007,Orbanz2008,Sudderth2009,Du2009,Ghosh2011}.

In this paper we propose an original class of BNP models which generalizes
the approach pioneered by \cite{Orbanz2008}.\ Our model combines
a spatial smoothness component to ensure that each data point is more
likely to have the same label as other spatially close data and a
partition prior to control the overall number and size of clusters.\ This
model is flexible enough to encompass the Potts model, the Potts-Dirichlet
Process (Potts-DP) model \cite{Orbanz2008} but additionally it allows
us to introduce easily prior information which prevents the creation
of small and/or large clusters.

Bayesian inference in this context is not analytically tractable and
requires the use of MCMC\ techniques. It is possible to derive a
simple single-site Gibbs sampler to perform Bayesian computation as
in \cite{Geman1984,Orbanz2008,Ghosh2011} but the mixing properties
of such samplers are poor.\ A popular alternative to single-site
Gibbs sampling for Potts models is the Swendsen-Wang (SW) algorithm
\cite{Swendsen1987} which originates from \cite{Fortuin1972}. In
an image segmentation context where the Potts model is unobserved,
SW can also mix poorly but a generalized version of it has been developed
to overcome this shortcoming \cite{Edwards1988,Higdon1998,Barbu2005,Barbu2007}.
We develop here an original Generalized SW (GSW) algorithm that is
reminiscent of split-merge samplers for Dirichlet process mixtures~\cite{Dahl2005} \cite{Jain2004} \cite{Hughes2012}.
For a particular setting of the BNP model parameters, our GSW is actually
an original split-merge sampler for Dirichlet process mixtures.

We demonstrate our BNP model and the associated GSW sampler on a standard
database of different natural scene types. We focus here on a truncated
version of the Potts-DP model penalizing low-size clusters. Experimentally
this model provides visually better segmentation results than its non-truncated
version and performs similarly to some recently proposed BNP\ alternatives.
From a computational point of view, the GSW\ allows us to better
explore high posterior probability regions compared to single-site
Gibbs.

The rest of this paper is organized as follows. In Section \ref{section:statisticalmodel},
we introduce a general statistical image segmentation model and discuss
various specific settings of interest.\ Section \ref{Section:generalizedSwendsenWang}
is devoted to Bayesian computation and we detail the GSW sampler.
We report our experimental results in Section \ref{Section:Experiments}.

\section{Statistical model\label{section:statisticalmodel}}

\subsection{Likelihood Model}

We model an observed image not as a collection of pixels but as a
collection of super-pixels, which correspond to small blocks of contiguous
textually-alike pixels \cite{Ren2003,Ghosh2011}. Unlike normal image
pixels that enjoy regular lattice with their neighbours, the super-pixels
on the other hand form irregular lattices with their neighbouring
super-pixels.

The $n$ super-pixels, called sites, $\mathbf{y}:=\left(y_{1},y_{2},...,y_{n}\right)$
constituting an image are assumed conditionally independent given
some latent variables $\mathbf{x}:=\left(x_{1},x_{2},...,x_{n}\right)$
with
\begin{equation}
\left.y_{i}\right\vert x_{i}\sim f(\cdot|x_{i})\label{eq:likelihood}
\end{equation}
where $\mathbf{x}$ can take a number of different values $k\leq n$
called cluster locations denoted $\mathbf{u}:=\left(u_{1},...,u_{k}\right)$.
These cluster locations are assumed to be statistically independent;
i.e. $u_{j}\overset{\text{i.i.d}}{\sim}\mathbb{G}_{0}$ for $j=1,\ldots,k$
where $\mathbb{G}_{0}$ is a probability measure with no atomic component.

The choice of $f\left(\left.\cdot\right\vert \cdot\right)$ and $\mathbb{G}_{0}$
is application dependent. In the experiments discussed in section
\ref{Section:Experiments}, $y_{i}$ is summarized by a histogram,
$f\left(\left.\cdot\right\vert \cdot\right)$ is a multinomial distribution,
$x_{i}$\ the associated multinomial parameters and $\mathbb{G}_{0}$
a finite Dirichlet distribution.

We associate to each site $i$ an allocation variable $z_{i}$ satisfying
$x_{i}=u_{z_{i}}$ and we denote $\mathbf{z}:=\left(z_{1},...,z_{n}\right)$.
Let $\Pi:=\Pi(\mathbf{z})$ be the random partition of $[n]:=\{1,\ldots,n\}$
defined by equivalence classes for the equivalence relation $z_{i}=z_{j}$.
The partition $\Pi=\{A_{1},\ldots,A_{k}\}$ is an unordered collection
of disjoint nonempty subsets $A_{j}$ of $[n]$, $j=1,\ldots,k$,
where $\cup_{j}A_{j}=[n]$ and $k\leq n$ is the number of subsets
for partition $\Pi$. Given the partition $\Pi$, the marginal likelihood
of the observations $\mathbf{y}$, integrating out cluster locations,
is given by
\begin{equation}
p(\mathbf{y}|\Pi)=\prod_{j=1}^{k}p(\mathbf{y}_{A_{j}})\label{eq:marglik}
\end{equation}
where $\mathbf{y}_{A_{j}}:=\{y_{i};i\in A_{j}\}$ and
\begin{equation}
p(\mathbf{y}_{A_{j}})=\int\prod_{i\in A_{j}}f(y_{i}|u_{j})\mathbb{G}_{0}(u_{j})du_{j}.\label{eq:margliklocal}
\end{equation}
We assume further on that $p(\mathbf{y}_{A_{j}})$ is known analytically;
e.g. $\mathbb{G}_{0}$ is a conjugate prior for $f$.

\subsection{Potts-Partition Model}

Our model combines a Potts-type spatial smoothness component and a
partition model. We review briefly the Potts model and partition models
before discussing how they can be combined in a simple way. We then
present examples of special interest.

\subsubsection{Potts model}

A standard approach to statistical image segmentation consists of
assigning a Potts prior distribution on $\mathbf{z}$ which introduces
some spatial smoothness in the clustering \cite{Higdon1998,Winkler2003}.
In this case, the allocation variables can only take a prespecified
number $K$ of different values and we set
\begin{equation}
P(\mathbf{z})\propto\exp\left(\sum_{i\leftrightarrow j}\beta_{ij}\mathbf{1}_{z_{i}=z_{j}}\right)
\end{equation}
where we write `$i\leftrightarrow j$' if the super-pixels $i$ and
$j$ are neighbours on a prespecified neighbouring structure, $\mathbf{1}_{z_{i}=z_{j}}=1$
if $z_{i}=z_{j}$ and $0$ otherwise. We set $\beta_{ij}>0$ to enforce
that two neighbours are more likely to have the same label. To simplify
notation, we will write
\[
P(\mathbf{z})\propto\exp\left(\sum_{i<j}\beta_{ij}\mathbf{1}_{z_{i}=z_{j}}\right)
\]
and set $\beta_{ij}=0$ if $i$ is not a neighbour of $j$. The Potts
model induces the following distribution over the partition $\Pi$
\begin{equation}
P(\Pi)\propto\left\{ \begin{array}{ll}
\frac{K!}{(K-k)!}\exp\left(\sum_{i<j}\beta_{ij}\mathbf{1}_{z_{i}=z_{j}}\right) & \text{if }1\leq k\leq K,\\
0 & \text{otherwise.}
\end{array}\right.\label{eq:Pottspartition}
\end{equation}

\subsubsection{Partition model}

We review a general class of partition models where the prior distribution
on partitions can be expressed in terms of an {\new exchangeable
probability function (EPF)} $g$~\cite{Pitman1995}; that is
\begin{equation}
P(\Pi=\{A_{1},\ldots,A_{k}\})=g(|A_{1}|,\ldots,|A_{k}|)\label{eq:PEFG}
\end{equation}
where $\left\vert A_{i}\right\vert $ denotes the size of the cluster
$A_{i}$ and $g$ is a symmetric function of its arguments, i.e.
\[
g(m_{1},\ldots,m_{k})=g(m_{\sigma(1)},\ldots,m_{\sigma(k)})
\]
for any permutation $\sigma$ of $k=2,3,...$ . The {\new EPF}
$g$ implicitly tunes the prior distribution on the overall number
of clusters $k$ and sizes of the clusters. A good overview of EPF for clustering can be found in \cite{Lau2006}.

\begin{figure}[ptb]
\begin{centering}
\includegraphics[width=8cm]{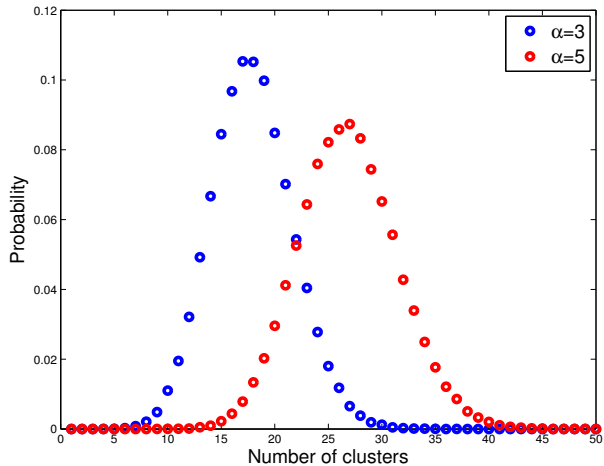}
\par\end{centering}

\caption{Distribution on the number of clusters for the Dirichlet Process partition
model with $n=1000$ and scale parameters $\alpha=3$ and $\alpha=5$.}

\label{fig:distclustDP}
\end{figure}

Let us denote $\mathbf{m:=}(m_{1},\ldots,m_{k})$ then, if we only
allow for a maximum number $K$ of clusters, we can select
\begin{equation}
g(\mathbf{m})\propto\left\{ \begin{array}{ll}
\frac{K}{(K-k)!} & \text{if }1\leq k\leq K\\
0 & \text{otherwise}
\end{array}\right.\label{eq:partitionpartPotts}
\end{equation}
which favours large values of $k$ but does not penalize cluster sizes
or
\begin{equation}
g(\mathbf{m})\propto\left\{ \begin{array}{ll}
\frac{K!}{(K-k)!}\prod_{j=1}^{k}\Gamma(\alpha+m_{j}) & \text{if }1\leq k\leq K\\
0 & \text{otherwise}
\end{array}\right.\label{eq:partitionfiniteDirichlet}
\end{equation}
where $\alpha>0$ which is the finite Dirichlet partition model. If
we do not limit the number of clusters, a very popular partition model
is the Dirichlet process partition model where for $k\geq1$
\begin{equation}
g(\mathbf{m})\propto\alpha^{k}\prod_{j=1}^{k}\Gamma(m_{j})\label{eq:partitionDirichlet}
\end{equation}
with $\Gamma$ the standard gamma function. The properties of this
distribution over partitions are well understood, see e.g. \cite{Lau2006}.
The parameter $\alpha$ tunes the number of clusters in the partition
as displayed in Figure~\ref{fig:distclustDP}, the mean number of
clusters being approximately $\alpha\log n$.

The Dirichlet process partition model can be further generalized to
the two parameter Poisson-Dirichlet partition model~\cite{Pitman1995}
given by
\begin{equation}
g(\mathbf{m})\propto\lbrack\theta+\alpha]_{\theta}^{k-1}\prod_{j=1}^{k}[1-\theta]_{1}^{m_{j}-1}\label{eq:partitionPitmanYor}
\end{equation}
where $[x]_{b}^{a}=x(x+b)\ldots(x+(a-1)b)$ and $(\alpha,\theta)$
verify either $\alpha>-\theta$ and $0\leq\theta<1$ or $\theta<0$
and $\alpha=-L\theta$ for some $L\in\mathbb{N}^{\ast}$. For $\theta=0$,
we obtain the Dirichlet process partition model.

All the previous models have been used extensively to address general
clustering tasks. In the specific context of image segmentation, it
can be of interest to exclude low size clusters. This is easily possible
by restricting the support of $g(\mathbf{m})$ to clusters of minimum
size $T_{\text{min}}$ so that for the {\new Dirichlet process}
prior, we have {\new
\begin{equation}
g(\mathbf{m})\propto\left\{ \begin{array}{ll}
\alpha^{k}\prod_{j=1}^{k}\Gamma(m_{j}) & \text{if }m_{j}\geq T_{\min}\text{ for all }j\\
0 & \text{otherwise.}
\end{array}\right.\label{eq:partitiontruncatedPitmanYor}
\end{equation}
}



\subsubsection{Combining Potts and partition models}

Our proposed model combines a spatial smoothness component of the
form
\begin{equation}
M(\Pi)=\exp\left(\sum_{i<j}\beta_{ij}\mathbf{1}_{z_{i}=z_{j}}\right)\label{eq:Potts}
\end{equation}
akin to the Potts model with a {\new EPF-type} model (\ref{eq:PEFG})
through
\begin{equation}
P(\Pi=\{A_{1},\ldots,A_{k}\})\propto M(\Pi)\times g(|A_{1}|,\ldots,|A_{k}|).\label{eq:PottsPPM}
\end{equation}
Clearly if we set $\beta_{ij}=0$ for all $(i,j)$ then one recovers
the classical priors for clustering allowing us to control the number
and size of clusters whereas $M(\Pi)$ ensures that spatially close
super-pixels are more likely to be in the same cluster.

If we select $g$ as (\ref{eq:partitionpartPotts}), we are back to
the standard Potts model given in (\ref{eq:Pottspartition}) whereas
if we select $g$ as the Dirichlet partition model (\ref{eq:partitionDirichlet})
then the proposed partition model (\ref{eq:PottsPPM}) corresponds
to the Potts-DP model of \cite{Orbanz2008}.

\begin{figure}[ptb]
\begin{centering}
\includegraphics[width=8cm]{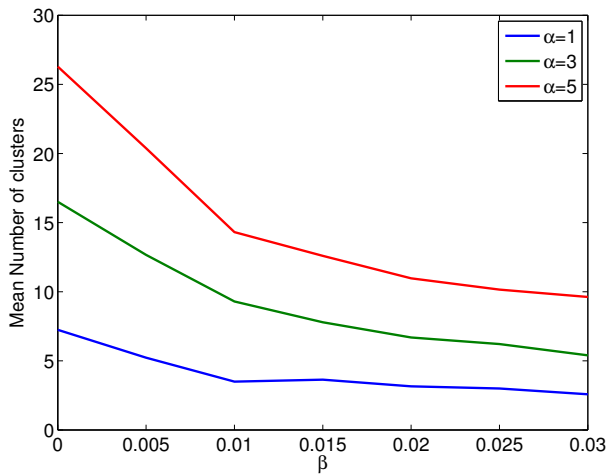}
\par\end{centering}

\caption{Mean number of clusters on an image with $n=1099$ sites for the Potts-Dirichlet
model with respect to the Potts parameter $\beta$ and for different
values of the Dirichlet process parameter $\alpha$. Results are obtained
by simulating from the prior. For $\beta=0$ ones recovers the classical
results associated to the Dirichlet process, i.e. 6.6, 17.2 and 26.5
clusters respectively for $\alpha=1,3,5$.}

\label{fig:simus}
\end{figure}

In Figure \ref{fig:simus}, we display the expected number of clusters
for the Potts-DP model for different values of the Potts parameter
$\beta_{ij}=\beta>0$ for neighbours and a Dirichlet process parameter
$\alpha>0$; the neighbouring structure is described in Section~\ref{Section:Experiments}.

Image segmentation now relies on the posterior over partitions
\begin{equation}
P(\left.\Pi\right\vert \mathbf{y})\propto p(\left.\mathbf{y}\right\vert \Pi)P\left(\Pi\right)\label{eq:posterior}
\end{equation}
where $p(\left.\mathbf{y}\right\vert \Pi)$ is given in (\ref{eq:marglik}).

\section{Generalized Swendsen-Wang algorithm for Bayesian computation\label{Section:generalizedSwendsenWang}}

\subsection{Single-site Gibbs sampler}

A standard strategy to sample approximately from $P(\left.\Pi\right\vert \mathbf{y})$
is to successively update the cluster assignment of each site $i$
given the cluster assignments of the other sites.\ This sampler proceeds
as follows. Let $\Pi_{-i}=\{A_{-i,1},\ldots,A_{-i,k_{-i}}\}$ be the
partition obtained by removing site $i$ from $\Pi$, and $m_{-i,j}$
be the size of $A_{-i,j}$.

The site $i$ will be assigned to cluster $j=1,\ldots,k_{-i}$ with
probability proportional to
\begin{equation}
g(m_{-i,1},\ldots,m_{-i,j}+1\ldots,m_{-i,k_{-i}})\frac{p(\mathbf{y}_{i\cup A_{-i,j}})}{p(\mathbf{y}_{A_{-i,j}})}\prod_{j=1}^{n}\exp(\beta_{ij}\mathbf{1}_{z_{i}=z_{j}})
\end{equation}
and be assigned to a new cluster with probability proportional to
\begin{equation}
g(m_{-i,1},\ldots,m_{-i,k_{-i}},1)p(y_{i}).
\end{equation}

This strategy, used by \cite{Orbanz2008} for the Potts-DP model,
is simple to implement, but exhibits poor mixing properties as the
cluster assignment of a given site is highly correlated with the cluster
assignments of its neighbours due to the spatial smoothness component.

\subsection{Generalized Swendsen-Wang sampler}

We propose here a GSW sampler that allows us to update simultaneously
cluster labels of groups of sites and hence improve the exploration
of the posterior. This algorithm can be interpreted as a generalization
of the technique proposed by \cite{Higdon1998} for standard Potts
models to our generalized Potts-partition model. It includes as special
cases the single-site Gibbs sampler presented previously and the classical
Swendsen-Wang algorithm.

The GSW relies on the introduction of auxiliary binary bond variables
$r_{ij}$ where $r_{ij}=1$ if sites $i$ and $j$ are bonded and
$0$ otherwise. We write $\mathbf{r}=(r_{ij})_{1\leq i<j\leq n}$
and the augmented model is defined by
\begin{equation}
P(\Pi,\mathbf{r})=P(\Pi)p(\mathbf{r}|\Pi)
\end{equation}
where
\[
P(\mathbf{r}|\Pi)=\prod_{1\leq i<j\leq n}P(r_{ij}|\Pi)
\]
with, for $1\leq i<j\leq n$,
\begin{equation}
P(r_{ij}=0|\Pi)=\exp(-\beta_{ij}\delta_{ij}\mathbf{1}_{z_{i}=z_{j}})=q_{ij}\label{eq:condrij}
\end{equation}
where $\delta_{ij}\geq0$; that is two neighbouring sites ($\beta_{ij}\neq0$)
sites with the same cluster assignment are bonded with probability
$1-\exp(-\beta_{ij}\delta_{ij})$. The parameters $\mathbf{\delta=}\left(\delta_{ij}\right)_{1\leq i<j\leq n}$
are hyperparameters of the GSW sampler whose choice will be discussed
later on.

The introduction of this augmented probabilistic model allows us to
sample from the resulting posterior distribution $P(\Pi,\mathbf{r}|\mathbf{y})$
using a block Gibbs strategy which iteratively and successively samples
$\mathbf{r}\sim P(\mathbf{r}|\Pi,\mathbf{y})$ and $\Pi\sim P(\Pi|\mathbf{r},\mathbf{y}).$

The bonds are independent of the data given the partition $\Pi$.
Therefore we have $P(\mathbf{r}|\Pi,\mathbf{y})=P(\mathbf{r}|\Pi)$
and the bond variables $r_{ij}$, $1\leq i<j\leq n$ are updated independently
using (\ref{eq:condrij}). The other conditional distribution $P(\Pi|\mathbf{r},\mathbf{y})$
can be expressed as
\begin{align*}
P(\Pi|\mathbf{r,y}) & \propto P(\Pi)P(\mathbf{r}|\Pi)p(\mathbf{y}|\Pi)\\
 & =g(\mathbf{m})\prod_{j=1}^{k}p(\mathbf{y}_{A_{j}})\prod_{1\leq i<j\leq n}(1-q_{ij})^{r_{ij}}\exp(-\beta_{ij}\mathbf{1}_{z_{i}=z_{j}})^{\delta_{ij}(1-r_{ij})}\prod_{1\leq i<j\leq n}\exp(\beta_{ij}\mathbf{1}_{z_{i}=z_{j}})\\
 & =g(\mathbf{m})\prod_{j=1}^{k}p(\mathbf{y}_{A_{j}})\prod_{1\leq i<j\leq n}\left[\exp(\beta_{ij}\delta_{ij}\mathbf{1}_{z_{i}=z_{j}})-1\right]^{r_{ij}}\left[\exp(\beta_{ij}\mathbf{1}_{z_{i}=z_{j}})\right]^{1-\delta_{ij}(1-r_{ij})-\delta_{ij}r_{ij}}\\
 & =g(\mathbf{m})\prod_{j=1}^{k}p(\mathbf{y}_{A_{j}})\prod_{1\leq i<j\leq n}\left[\exp(\beta_{ij}\delta_{ij}\mathbf{1}_{z_{i}=z_{j}})-1\right]^{r_{ij}}\left[\exp(\beta_{ij}(1-\delta_{ij})\mathbf{1}_{z_{i}=z_{j}})\right]
\end{align*}
The bond variables $r_{ij}$ induce groups of sites which have the
same cluster label, as the term $\prod_{1\leq i<j\leq n}\left[\exp(\beta_{ij}\delta_{ij}\mathbf{1}_{z_{i}=z_{j}})-1\right]^{r_{ij}}$
implies that the conditional distribution only assigns positive probability
mass to partitions where bonded sites are in the same cluster. Let
$C_{1},\ldots,C_{p}$ be the groups of sites, or spin-clusters, induced
by the bonds. We denote by $\Pi_{-\ell}=\{A_{-\ell,1},\ldots,A_{-\ell,k_{-\ell}}\}$
the partition obtained by removing sites $i\in C_{\ell}$ from $\Pi$
and $m_{-\ell,j}$ the size of $A_{-\ell,j}$. Note that this notation
differs slightly from the one introduced in the previous section on
single-site Gibbs sampling.

We then successively update the cluster assignment of each spin-cluster
$C_{\ell}$ given the cluster assignments of the other spin-clusters.\ The
spin-cluster $C_{\ell}$ is assigned to cluster $j=1,\ldots,k_{-\ell}$
with probability proportional to
\begin{equation}
g(m_{-\ell,1},\ldots,m_{-\ell,j}+|C_{\ell}|\ldots,m_{-\ell,k_{-\ell}})\frac{p(\mathbf{y}_{C_{\ell}\cup A_{-\ell,j}})}{p(\mathbf{y}_{A_{-\ell,j}})}\prod_{\{(i,j)|i\in C_{\ell},r_{ij}=0\}}\exp(\beta_{ij}(1-\delta_{ij})\mathbf{1}_{z_{i}=z_{j}})\label{eq:conditional1}
\end{equation}
and to a new cluster with probability proportional to
\begin{equation}
g(m_{-\ell,1},\ldots,m_{-\ell,k_{-\ell}},|C_{\ell}|)p(\mathbf{y}_{C_{\ell}}).\label{eq:conditional2}
\end{equation}
\bigskip{}

As an example, for the Potts-DP model, (\ref{eq:conditional1}) becomes
\[
\frac{\Gamma(m_{-\ell,j}+|C_{\ell}|)}{\Gamma(m_{-\ell,j})}\frac{p(\mathbf{y}_{C_{\ell}\cup A_{-\ell,j}})}{p(\mathbf{y}_{A_{-\ell,j}})}\prod_{\{(i,j)|i\in C_{\ell},r_{ij}=0\}}\exp(\beta_{ij}(1-\delta_{ij})\mathbf{1}_{z_{i}=z_{j}})
\]
while (\ref{eq:conditional2}) corresponds to
\begin{equation}
\alpha\Gamma(|C_{\ell}|)p\left(\mathbf{y}_{C_{\ell}}\right).
\end{equation}

\bigskip{}

The difference between the original SW and this generalized version
is the term
\[
\prod_{\{(i,j)|i\in C_{\ell},r_{ij}=0\}}\exp(\beta_{ij}(1-\delta_{ij})\mathbf{1}_{z_{i}=z_{j}})
\]
which only depends on the cluster assignments of the sites which are
neighbours of the group $C_{\ell}$. The algorithm reduces to single
site Gibbs sampling if $\delta_{ij}=0$ and to the classical SW algorithm
if $\delta_{ij}=1$.

\bigskip{}

\bigskip{}

\begin{figure}[h]
\begin{centering}
\subfigure{\begin{tikzpicture}[scale=.6]
\draw (0,0) grid (5,5);
\foreach \c in {(0,0),(1,0),(1,1),(2,1),(2,0),(0,1),(4,2),(4,3),(4,4),(4,5),(5,2),(5,3),(5,5),(3,4)}
    \node[greencircle]()at\c {};
\foreach \c in {(1,2),(1,3),(2,2),(2,3),(3,2),(3,3),(5,4)}
    \node[bluecircle]()at\c {};
\foreach \c in {(0,2),(0,3),(0,4),(0,5),(1,5),(1,4),(2,5),(2,4),(3,5),(3,0),(3,1),(4,0),(4,1),(5,0),(5,1)}
    \node[redcircle]()at\c {};
\end{tikzpicture}}\hskip.2in \subfigure{\begin{tikzpicture}[scale=.6]
\draw[thick] (0,0) grid (2,1);
\draw[thick] (0,2) grid (0,4);
\draw[thick] (3,0) grid (5,1);
\draw[thick] (0,5) grid (3,5);
\draw[thick] (1,4) grid (2,5);
\draw[thick] (1,2) grid (3,2);
\draw[thick] (1,2) grid (1,3);
\draw[thick] (3,2) grid (3,3);
\draw[thick] (4,2) grid (5,2);
\draw[thick] (4,3) grid (5,3);
\draw[thick] (4,3) grid (4,5);
\draw[thick] (3,4) grid (4,4);
\draw[thick] (4,5) grid (5,5);
\foreach \c in {(0,0),(1,0),(1,1),(2,1),(2,0),(0,1),(4,2),(4,3),(4,4),(4,5),(5,2),(5,3),(5,5),(3,4)}
    \node[greencircle]()at\c {};
\foreach \c in {(1,2),(1,3),(2,2),(2,3),(3,2),(3,3),(5,4)}
    \node[bluecircle]()at\c {};
\foreach \c in {(0,2),(0,3),(0,4),(0,5),(1,5),(1,4),(2,5),(2,4),(3,5),(3,0),(3,1),(4,0),(4,1),(5,0),(5,1)}
    \node[redcircle]()at\c {};
\end{tikzpicture}}
\hskip.2in \subfigure{\begin{tikzpicture}[scale=.6]
\draw[thick] (0,0) grid (2,1);
\draw[thick] (0,2) grid (0,4);
\draw[thick] (3,0) grid (5,1);
\draw[thick] (0,5) grid (3,5);
\draw[thick] (1,4) grid (2,5);
\draw[thick] (1,2) grid (3,2);
\draw[thick] (1,2) grid (1,3);
\draw[thick] (3,2) grid (3,3);
\draw[thick] (4,2) grid (5,2);
\draw[thick] (4,3) grid (5,3);
\draw[thick] (4,3) grid (4,5);
\draw[thick] (3,4) grid (4,4);
\draw[thick] (4,5) grid (5,5);
\draw[thick,rounded corners=5pt] (-.4,-.4) -| (2.4,1.4) -| (-.4,-.4);
\draw[thick,rounded corners=5pt] (2.6,-.4) -| (5.4,1.4) -| (2.6,-.4);
\draw[thick,rounded corners=5pt] (-.4,1.6) -| (0.4,4.4) -| (-.4,1.6);
\draw[thick,rounded corners=5pt] (0.6,1.6) -| (3.4,3.4) -| (2.6,2.4) -| (1.4,3.4) -| (0.6,1.6);
\draw[thick,rounded corners=5pt] (-.4,5.4) -| (3.4,4.6) -| (2.4,3.6) -| (0.6,4.6) -| (-.4,5.4);
\draw[thick,rounded corners=5pt] (1.6,2.6) -| (2.4,3.4) -| (1.6,2.6);
\draw[thick,rounded corners=5pt] (3.6,1.6) -| (5.4,2.4) -| (3.6,1.6);
\draw[thick,rounded corners=5pt] (3.6,2.6) -| (5.4,3.4) -| (4.4,4.6) -| (5.4,5.4)  -| (3.6,4.4)  -| (2.6,3.6) -| (3.6,2.6);
\draw[thick,rounded corners=5pt] (4.6,3.6) -| (5.4,4.4) -| (4.6,3.6);
\foreach \c in {(0,0),(1,0),(1,1),(2,1),(2,0),(0,1),(4,2),(4,3),(4,4),(4,5),(5,2),(5,3),(5,5),(3,4)}
    \node[greencircle]()at\c {};
\foreach \c in {(1,2),(1,3),(2,2),(2,3),(3,2),(3,3),(5,4)}
    \node[bluecircle]()at\c {};
\foreach \c in {(0,2),(0,3),(0,4),(0,5),(1,5),(1,4),(2,5),(2,4),(3,5),(3,0),(3,1),(4,0),(4,1),(5,0),(5,1)}
    \node[redcircle]()at\c {};
\end{tikzpicture}}\hskip.2in \subfigure{\begin{tikzpicture}[scale=.6]
\draw[thick,rounded corners=5pt] (-.4,-.4) -| (2.4,1.4) -| (-.4,-.4);
\draw[thick,rounded corners=5pt] (2.6,-.4) -| (5.4,1.4) -| (2.6,-.4);
\draw[thick,rounded corners=5pt] (-.4,1.6) -| (0.4,4.4) -| (-.4,1.6);
\draw[thick,rounded corners=5pt] (0.6,1.6) -| (3.4,3.4) -| (2.6,2.4) -| (1.4,3.4) -| (0.6,1.6);
\draw[thick,rounded corners=5pt] (-.4,5.4) -| (3.4,4.6) -| (2.4,3.6) -| (0.6,4.6) -| (-.4,5.4);
\draw[thick,rounded corners=5pt] (1.6,2.6) -| (2.4,3.4) -| (1.6,2.6);
\draw[thick,rounded corners=5pt] (3.6,1.6) -| (5.4,2.4) -| (3.6,1.6);
\draw[thick,rounded corners=5pt] (3.6,2.6) -| (5.4,3.4) -| (4.4,4.6) -| (5.4,5.4)  -| (3.6,4.4)  -| (2.6,3.6) -| (3.6,2.6);
\draw[thick,rounded corners=5pt] (4.6,3.6) -| (5.4,4.4) -| (4.6,3.6);
\foreach \c in {(0,0),(1,0),(1,1),(2,1),(2,0),(0,1),(4,2),(5,2),(2,3)}
    \node[greencircle]()at\c {};
\foreach \c in {(0,2),(0,3),(0,4),(1,2),(1,3),(2,2),(3,2),(3,3),(5,4)}
    \node[bluecircle]()at\c {};
\foreach \c in {(0,5),(1,5),(1,4),(2,5),(2,4),(3,5),(4,3),(4,4),(4,5),(5,3),(5,5),(3,4)}
    \node[redcircle]()at\c {};
\foreach \c in {(3,0),(3,1),(4,0),(4,1),(5,0),(5,1)}
    \node[blackcircle]()at\c {};
\end{tikzpicture} }
\par\end{centering}

\caption{Illustration of the GSW algorithm on a regular lattice graph where
each site has 4 neighbours. (a) The partition $\Pi$ is represented
by colors. (b) Each pair of neighbors $(i,j)$ in the same cluster
is bonded with probability $1-q_{ij}$. (c) This defines a partition
of the $n$ sites into spin-clusters $C_{1},\ldots,C_{p}$. (d) Each
spin-cluster $C_{\ell}$ is successively assigned to an existing or
a new cluster conditionally on the colors of the other spin-cluster
clusters to obtain a new partition $\Pi$ of the $n$ sites.}

\label{fig:swendsen}
\end{figure}
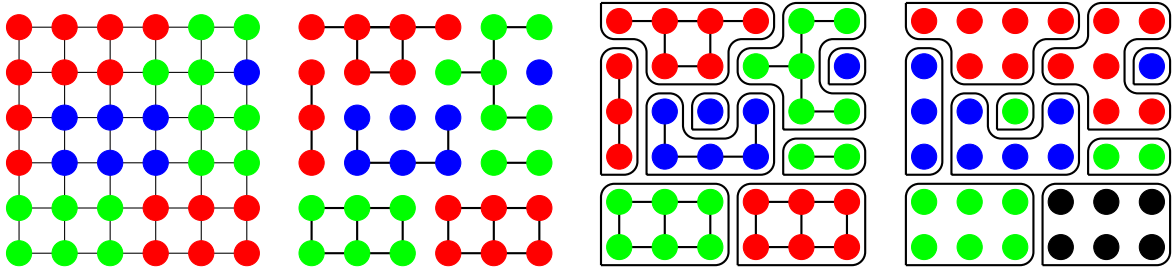

The overall GSW sampler, which is summarized in Figure~\ref{fig:swendsen},
proceeds as follows at each iteration.
\begin{itemize}
\item For each $1\leq i<j\leq n\ $such that $\beta_{ij}\neq0$, sample
the bond variables
\[
r_{ij}\sim\text{Ber}(1-\exp(-\beta_{ij}\delta_{ij}\mathbf{1}_{z_{i}=z_{j}}))
\]
where Ber$\left(\upsilon\right)$ is the Bernoulli distribution of
parameter $\upsilon.$ Let $C_{1},\ldots C_{p}$ denote the corresponding
spin-clusters.
\item For each spin-cluster $\ell=1,\ldots,p$

\begin{itemize}
\item Let $\Pi_{-\ell}=\{A_{-\ell,1},\ldots,A_{-\ell,k_{-\ell}}\}$ be the
partition obtained by removing sites $i\in C_{\ell}$ from $\Pi$,
and $m_{-\ell,j}$ be the size of $A_{-\ell,j}$.\ Then all sites
in the spin-cluster $C_{\ell}$ will be associated to cluster $j=1,\ldots,k_{-\ell}$
with probability proportional to
\begin{equation}
g(m_{-\ell,1},\ldots,m_{-\ell,j}+|C_{\ell}|\ldots,m_{-\ell,k_{-\ell}})\frac{p(\mathbf{y}_{C_{\ell}\cup A_{-\ell,j}})}{p(\mathbf{y}_{A_{-\ell,j}})}\prod_{\{(i,j)|i\in C_{\ell},r_{ij}=0\}}\exp\left(\beta_{ij}(1-\delta_{ij})\mathbf{1}_{z_{i}=z_{j}}\right)
\end{equation}
or be associated to a new cluster $k_{-\ell}+1$ with probability
proportional to
\begin{equation}
g(m_{-\ell,1},\ldots,m_{-\ell,k_{-\ell}},|C_{\ell}|)p(\mathbf{y}_{C_{\ell}})
\end{equation}

\end{itemize}
\end{itemize}

Whatever being the hyperparameters $\mathbf{\delta=}\left(\delta_{ij}\right)_{1\leq i<j\leq n}$,
the Markov transition kernel of the GSW\ sampler admits $P(\Pi,\mathbf{r}|\mathbf{y})$
as invariant distribution. A first simple choice, that is made in
this article, is to set $\delta_{ij}=\lambda>0$ for all $(i,j)$
such that $\beta_{ij}\neq0$. Another choice is to set $\delta_{ij}$
based on the observations. For example, we can set
\begin{equation}
\delta_{ij}=\lambda\exp(-\tau d(y_{i},y_{j}))\label{eq:auxiliaryparameters}
\end{equation}
where $d(\cdot,\cdot)$ is some distance measure and $\lambda,\tau$
are some positive tuning parameters. We also tried this setting, but
this did not improve the results significantly. More sophisticated
choices have been proposed by \cite{Barbu2005,Barbu2007} in the context of the standard Potts model. The authors
report significant improvements but we have not pursued this approach
here.

\section{Experiments\label{Section:Experiments}}

\subsection{Dataset, preprocessing and likelihood}

The dataset and preprocessing steps adopted here follow essentially
\cite{Ghosh2011} \cite{Ghosh2012}. We used two different datasets. The first dataset is from \emph{labelme}
toolbox' \cite{russell2008labelme} ``8 scene categories\textquotedblright{},
which is comprised of eight categories of natural outdoor images.
All images used contain 256x256 pixels. {\new The second dataset
is the BSB300 \cite{MartinFTM01} which contain images of variable
sizes.} Sampling-based segmentation methods could be prohibitively
slow if we were associated to each single pixel location a different
site. Therefore, mimicking the steps taken in the dependent Chinese
Restaurant Process (dd-CRP) and its Hierarchical (regional) variant
(rdd-CRP) \cite{Ghosh2011}, we first group image pixels into so-called,
super-pixels, in which around 60, colour/textually-alike pixels are grouped
to form a single super-pixel. The super-pixel representation is a frequently used techniques to pre-group pixels in image processing literatures. The computation is relatively fast. It reduces the amount of sites one has to perform from a full image size to merely around $1000$ sites.  In our work, we used a standard super-pixel toolbox \cite{mori2004recovering}, Although it's not a central theme of our research, but we anticipate that by using a more state-of-the-art super pixel generation algorithm, it should improve our segmentation result even further.

The observation $y_{i}$ is then constructed by forming a histogram
of the pixels within each super pixel. Instead of using a simple 8-bin
histogram as in \cite{Orbanz2008}, we followed the technique of \cite{Ghosh2011},
in which colour information was used to construct a 120-bin representation
via a clustering procedure. In order to provide a fair comparison,
the same histogram pre-processing is used for all the sampling-based
methods described in this paper, namely, dd-CRP, rdd-CRP and our algorithm.
The difference, however, between our work and the work of \cite{Ghosh2011} is
that we do not use texton histograms. This is done deliberately to
perform a fair comparison with the mean-shift method, which uses purely
luminance information.

We optimized the parameters of both mean-shift and rdd-CRP so as to
maximize the rand index \cite{hubert1985comparing} by using a training
set of images from the LabelMe dataset. The optimal rdd-CRP parameters
we obtained were $\alpha=10^{-10}$ and $\gamma=10^{-2}$. For mean-shift,
we used a spatial bandwidth of 20, a range bandwidth of 15 and a minimum
region size of 1500.

To compute $p(\mathbf{y}_{C_{l}})$, we use for $f\left(\left.\cdot\right\vert \cdot\right)$
a multinomial distribution and $\mathbb{G}_{0}$ a 120-dimensional
Dirichlet distribution with concentration vector $\pi=\phi\bar{\mathbf{y}}$,
where $\bar{\mathbf{y}}$ is the normalised sum of all $n$ data histograms,
and $\sum_{d=1}^{120}\bar{\mathbf{y}}_{d}=1$. We set $\phi=50$ for
the concentration parameter.

\subsection{Evaluation of the generalized Swendsen-Wang algorithm}

\begin{figure}[ptb]
\subfigure{\includegraphics[width=0.5\textwidth]{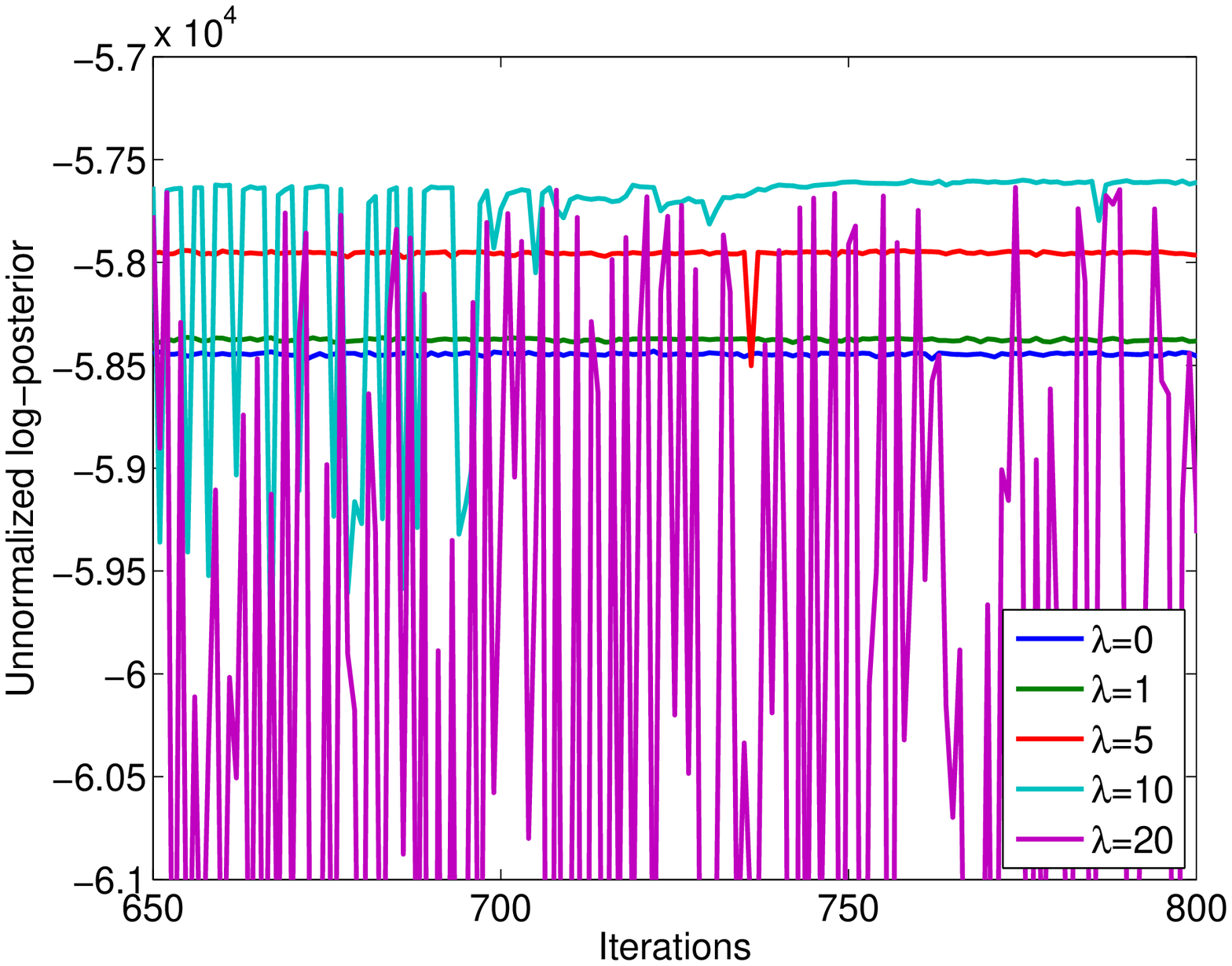}}
\subfigure{\includegraphics[width=0.5\textwidth]{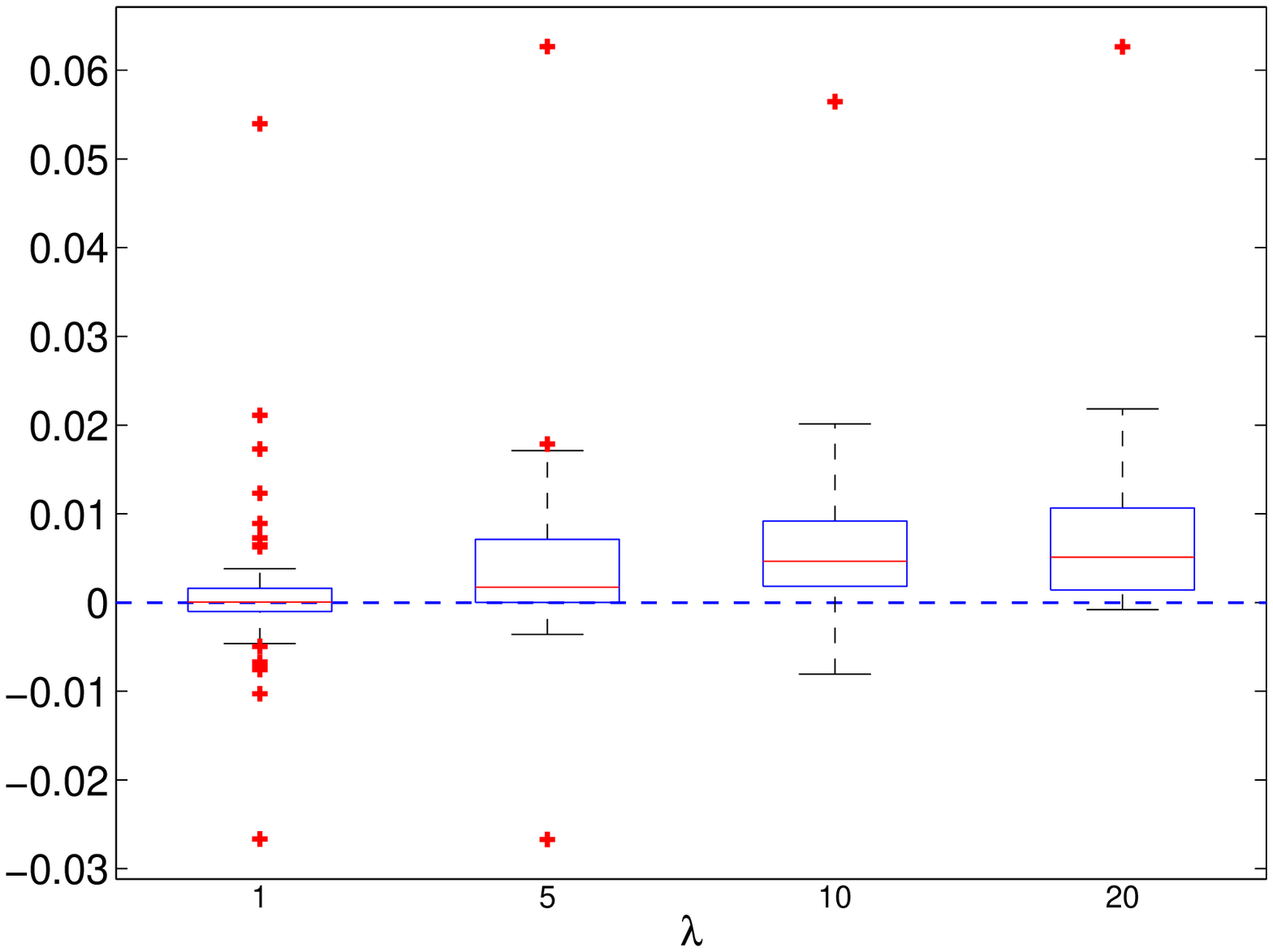}}\caption{Comparison of the GSW algorithm for various values of $\lambda$:
$\lambda=0$ (single-site Gibbs), $\lambda=1$ (classical SW), $\lambda=5,10,20$.
(a) Typical run of the GSW sampler for various values of $\lambda$.
Increasing the value of $\lambda$ leads to a better exploration of
the posterior distribution. Low values ($\lambda=0,1$) typically
get stuck in a local maximum. Values above $\lambda=10$ lead to very
large fluctuations. (b) Percentage of increase in the maximum of the
log-posterior w.r.t. single-site Gibbs ($\lambda=0$) for 1000 iterations
and over 48 images.}

\label{fig:genSWcomp}
\end{figure}

We display the performance of the GSW algorithms for the Potts-DP
model with $\alpha=3$ and $\beta=0.02$ and different values of $\lambda$
in Figure~\ref{fig:genSWcomp}. Increasing the value of $\lambda$
allows us to better explore the posterior distribution, see Figure~\ref{fig:genSWcomp}(a).
When using GSW as a stochastic search algorithm to get a Maximum A
Posteriori (MAP) estimate, better MAP\ estimates are obtained on
average as $\lambda$ increases until about 10-20, as shown in Figure~\ref{fig:genSWcomp}(b).
Using too high a value of $\lambda$ is inefficient as it slows down
the convergence of the Markov chain to its stationary distribution.
We found experimentally that $\lambda=10$ provides on average good
and stable results.

Similar conclusions were reached when using different values of $\alpha$
and $\beta$ and for the truncated Potts-DP\ model discussed in (\ref{eq:partitiontruncatedPitmanYor}).

\subsection{Comparison to other methods}

\begin{figure}[ptb]
\begin{centering}
\includegraphics[width=0.15\textwidth]{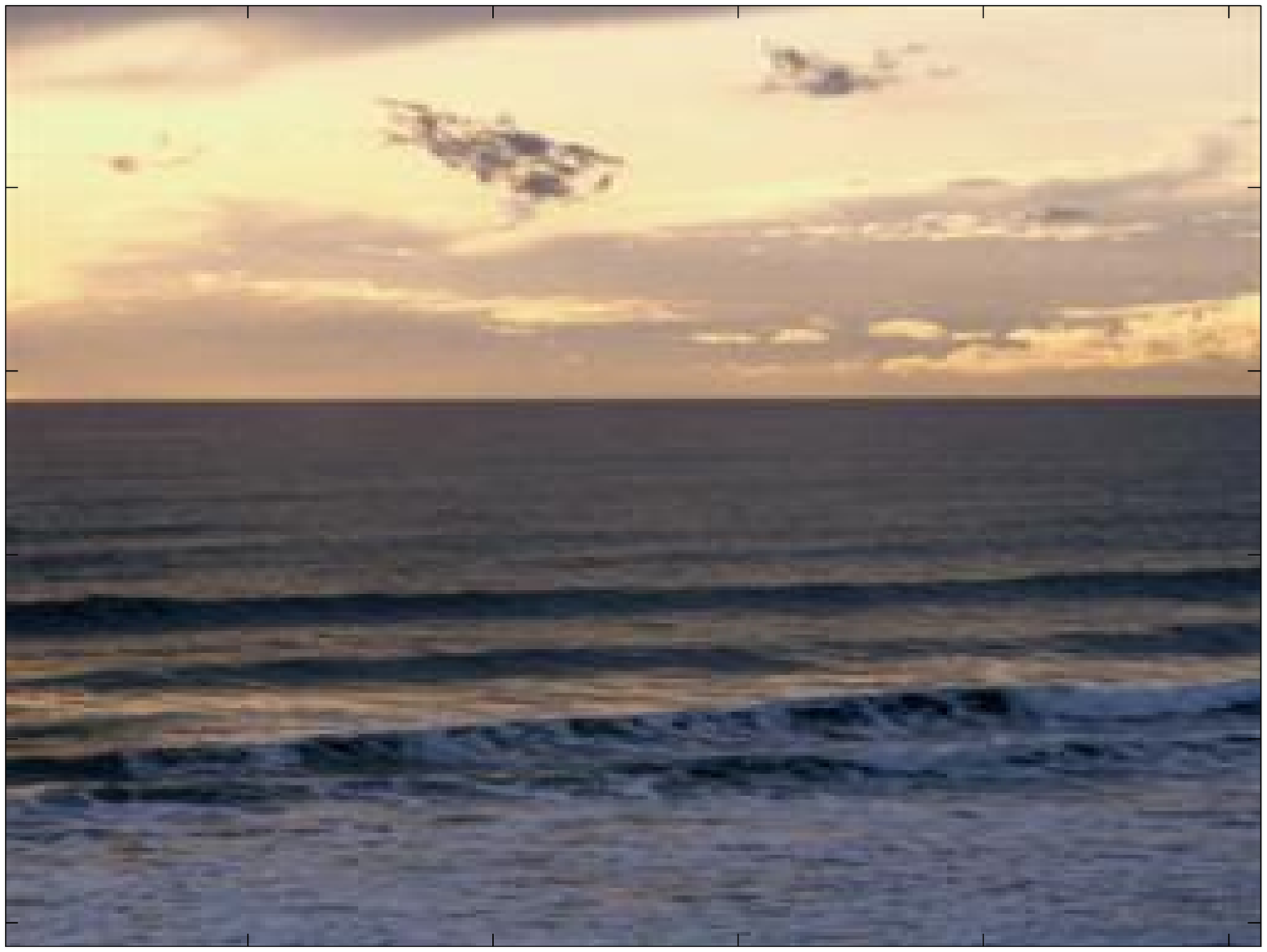}
\includegraphics[width=0.15\textwidth]{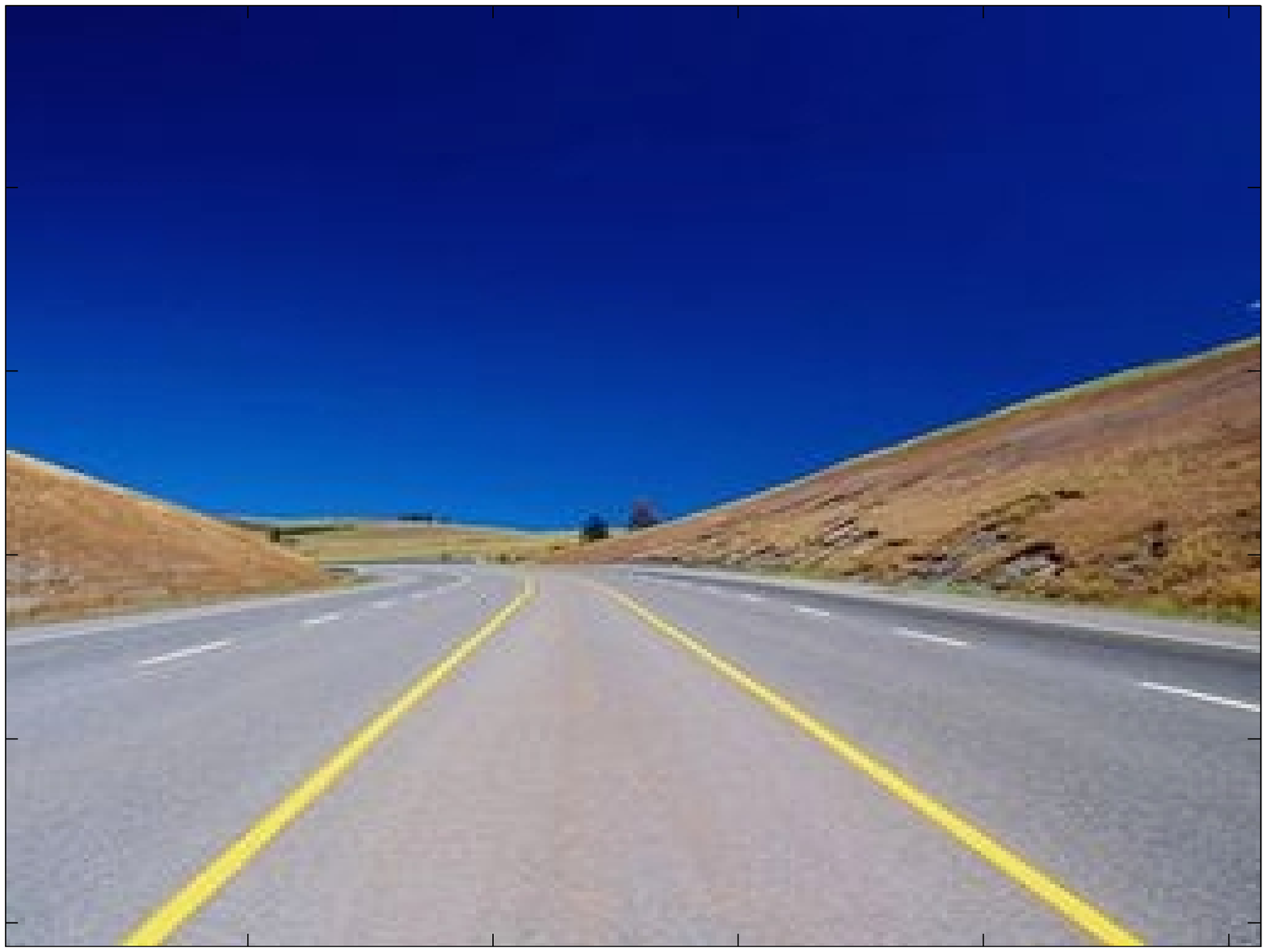}
\includegraphics[width=0.15\textwidth]{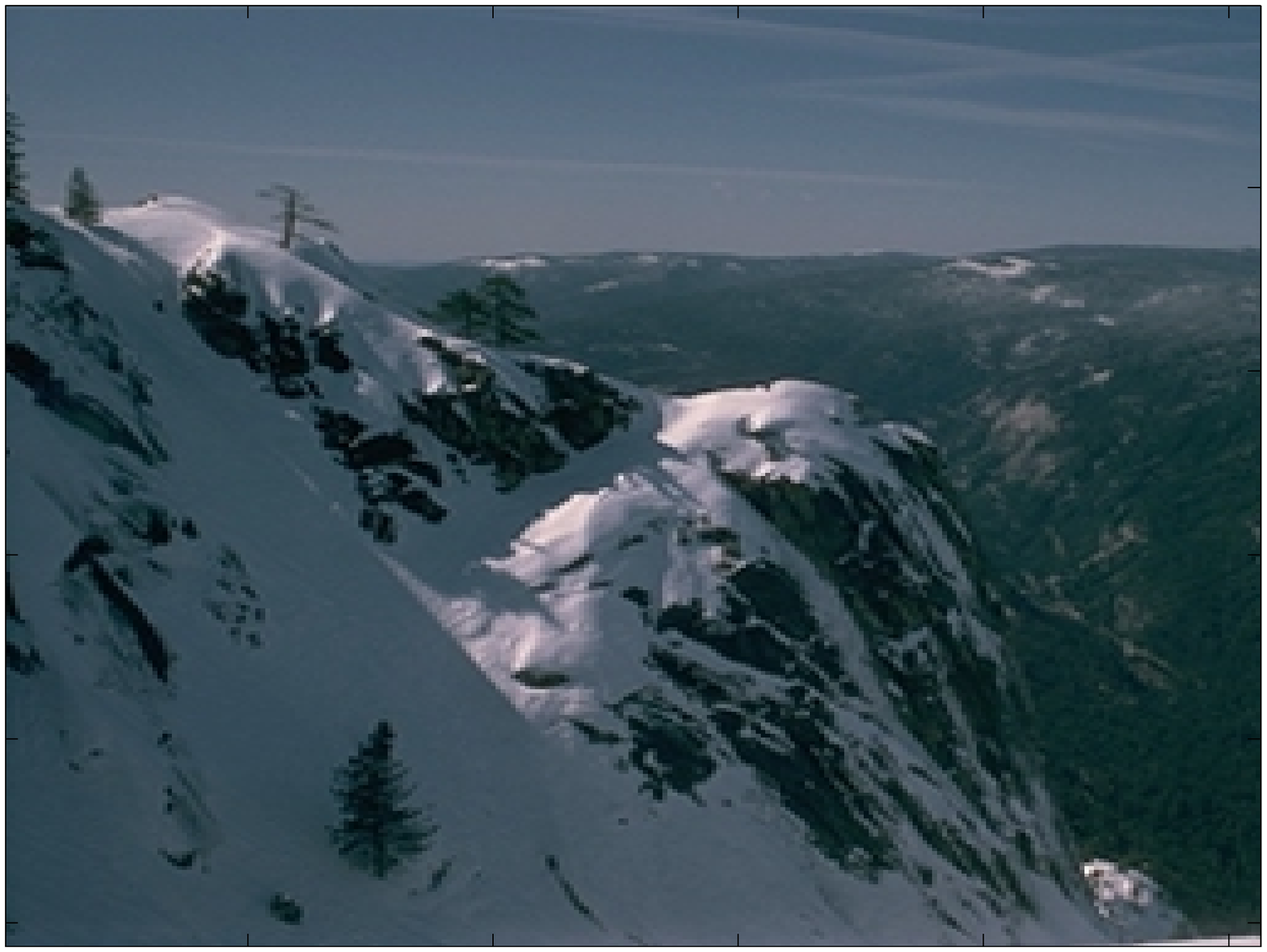}
\includegraphics[width=0.15\textwidth]{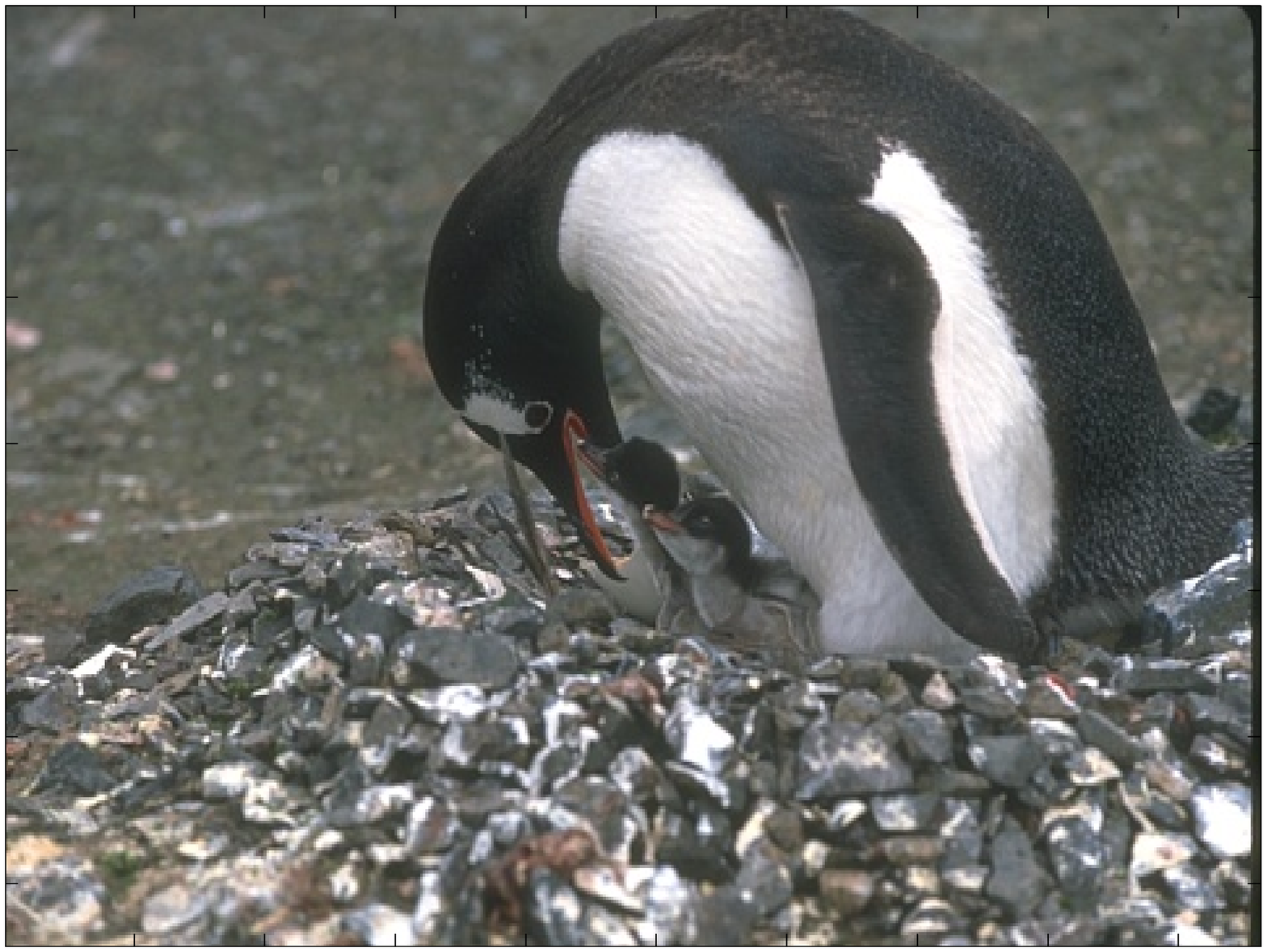}
\includegraphics[width=0.15\textwidth]{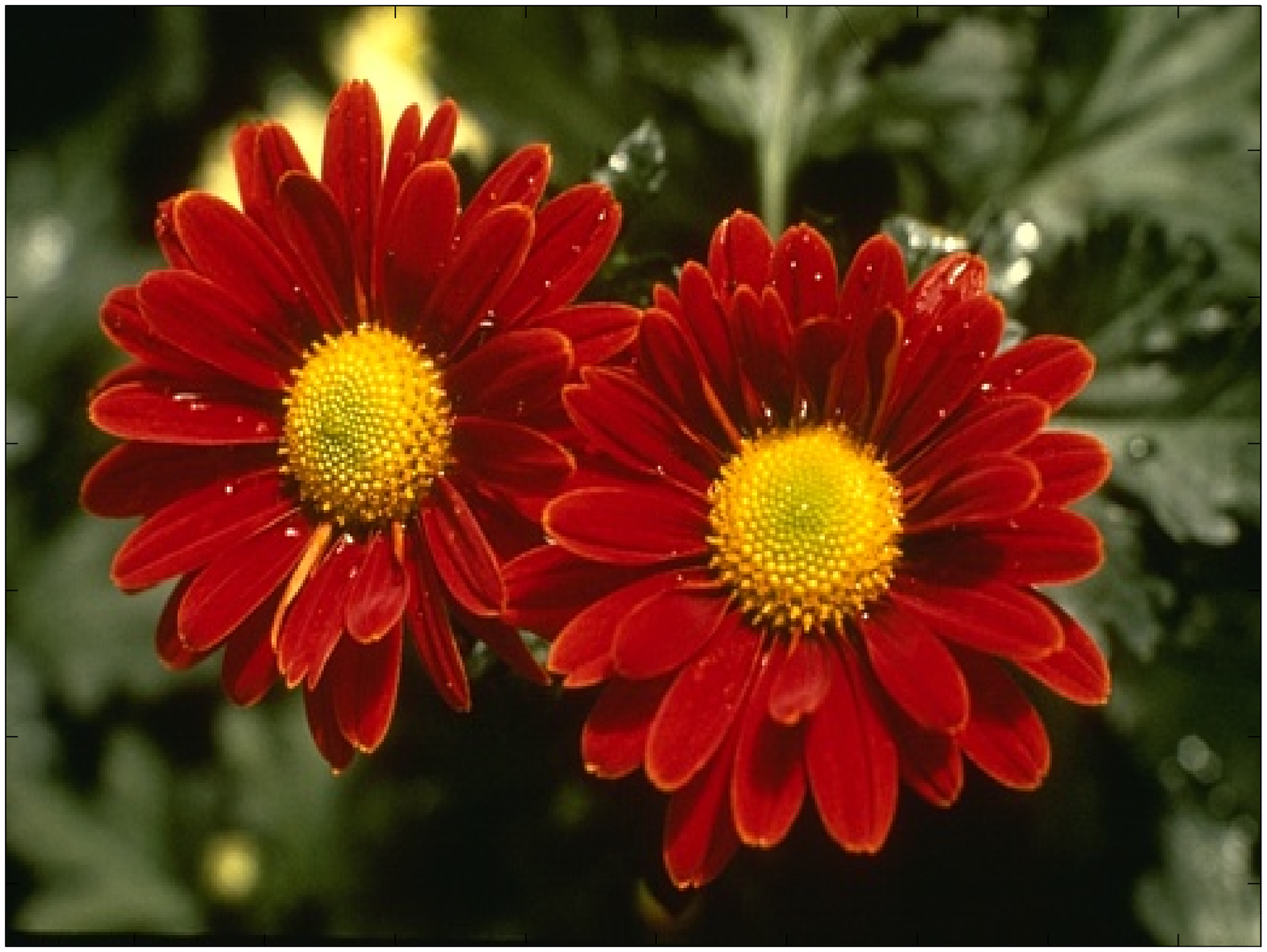}
\includegraphics[width=0.15\textwidth]{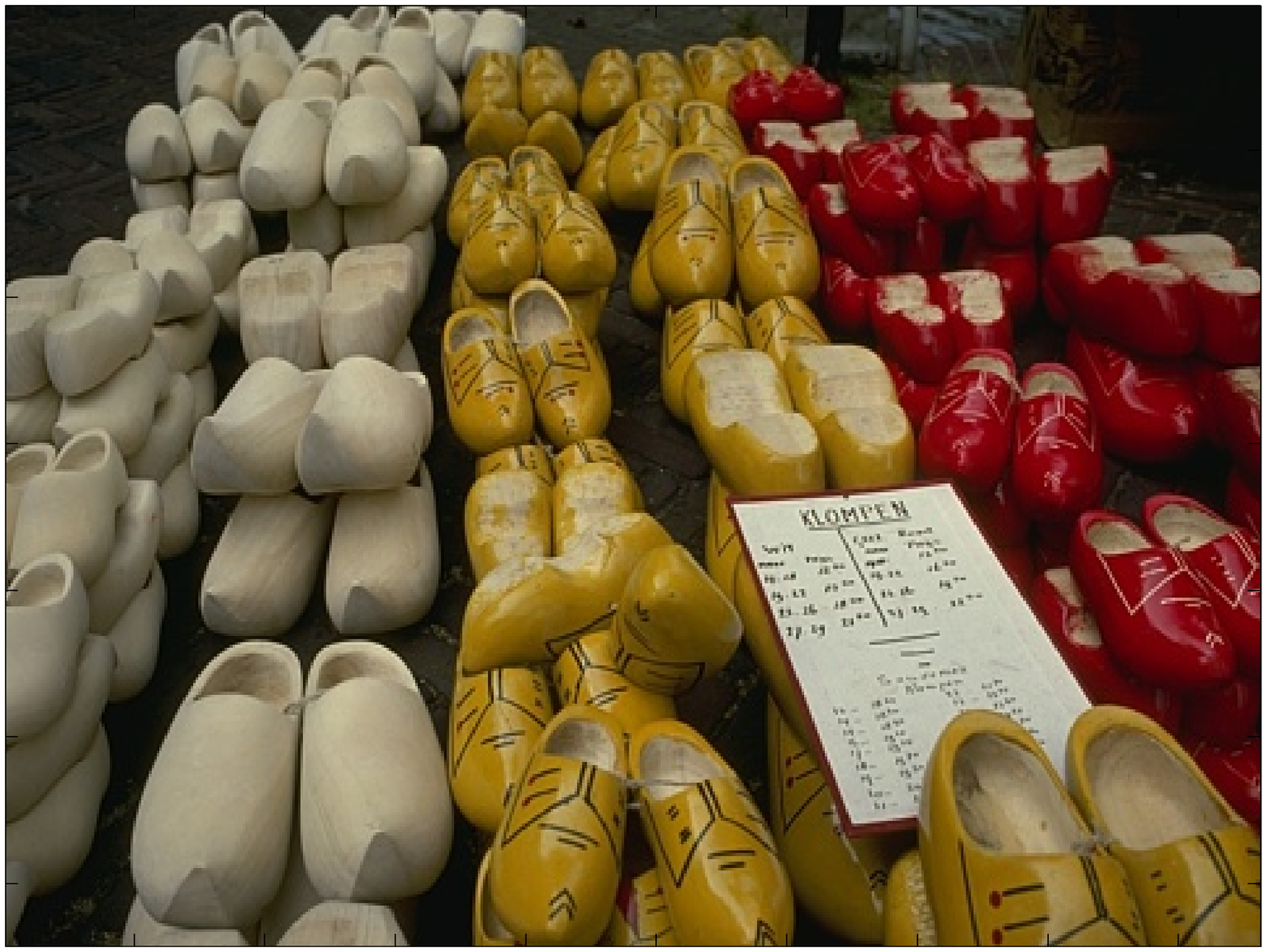}\\
 \includegraphics[width=0.15\textwidth]{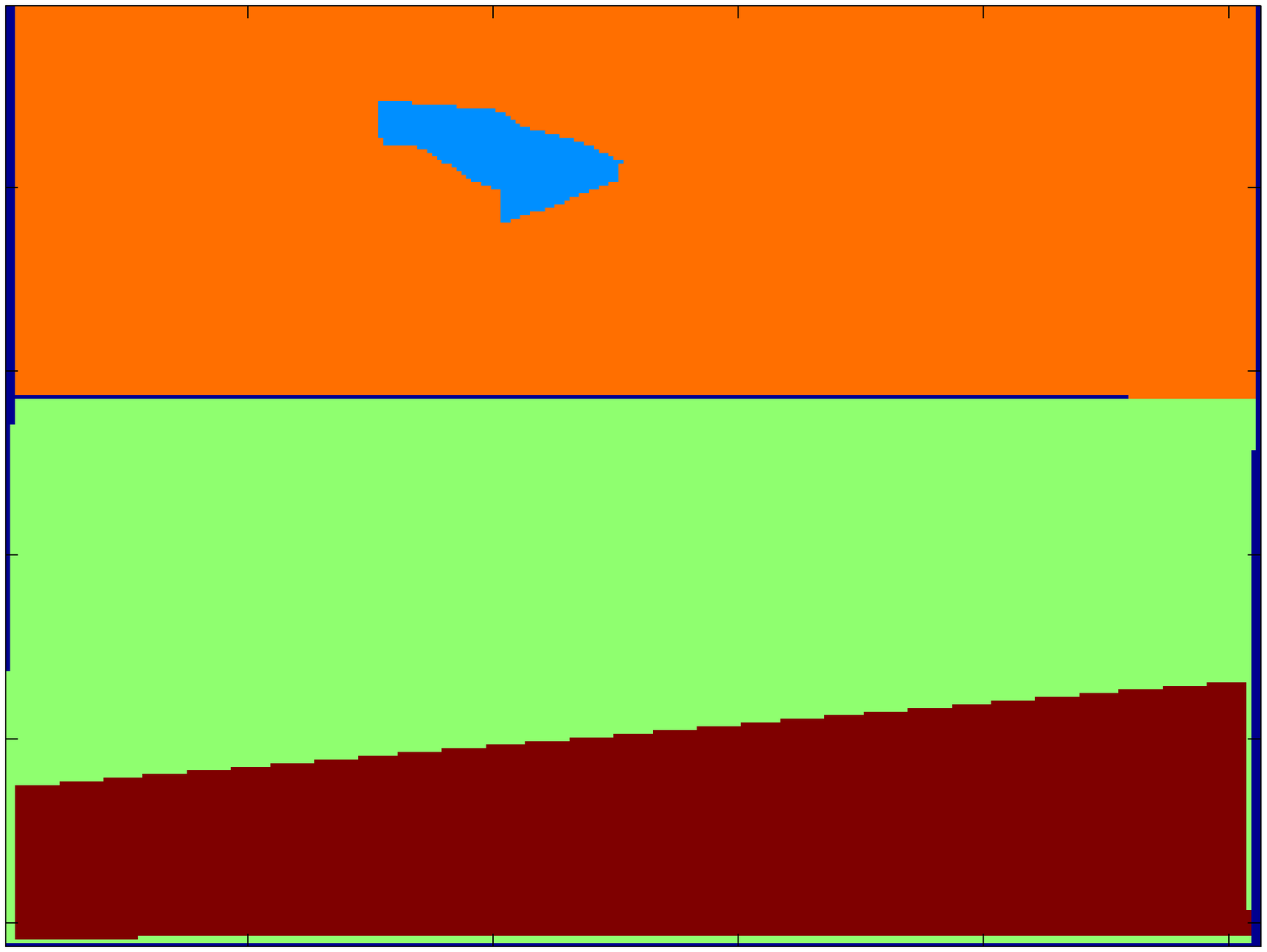}
\includegraphics[width=0.15\textwidth]{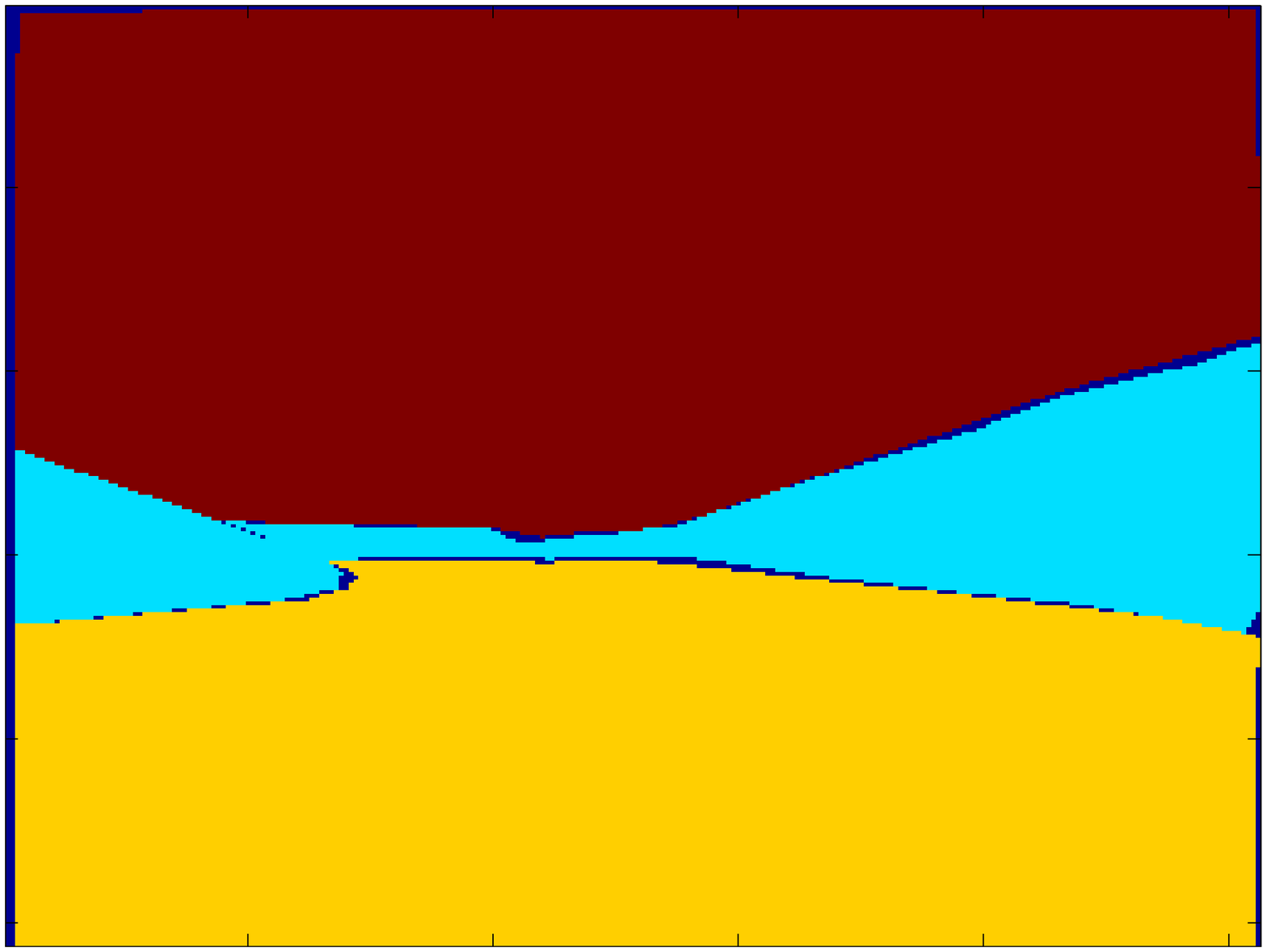}
\includegraphics[width=0.15\textwidth]{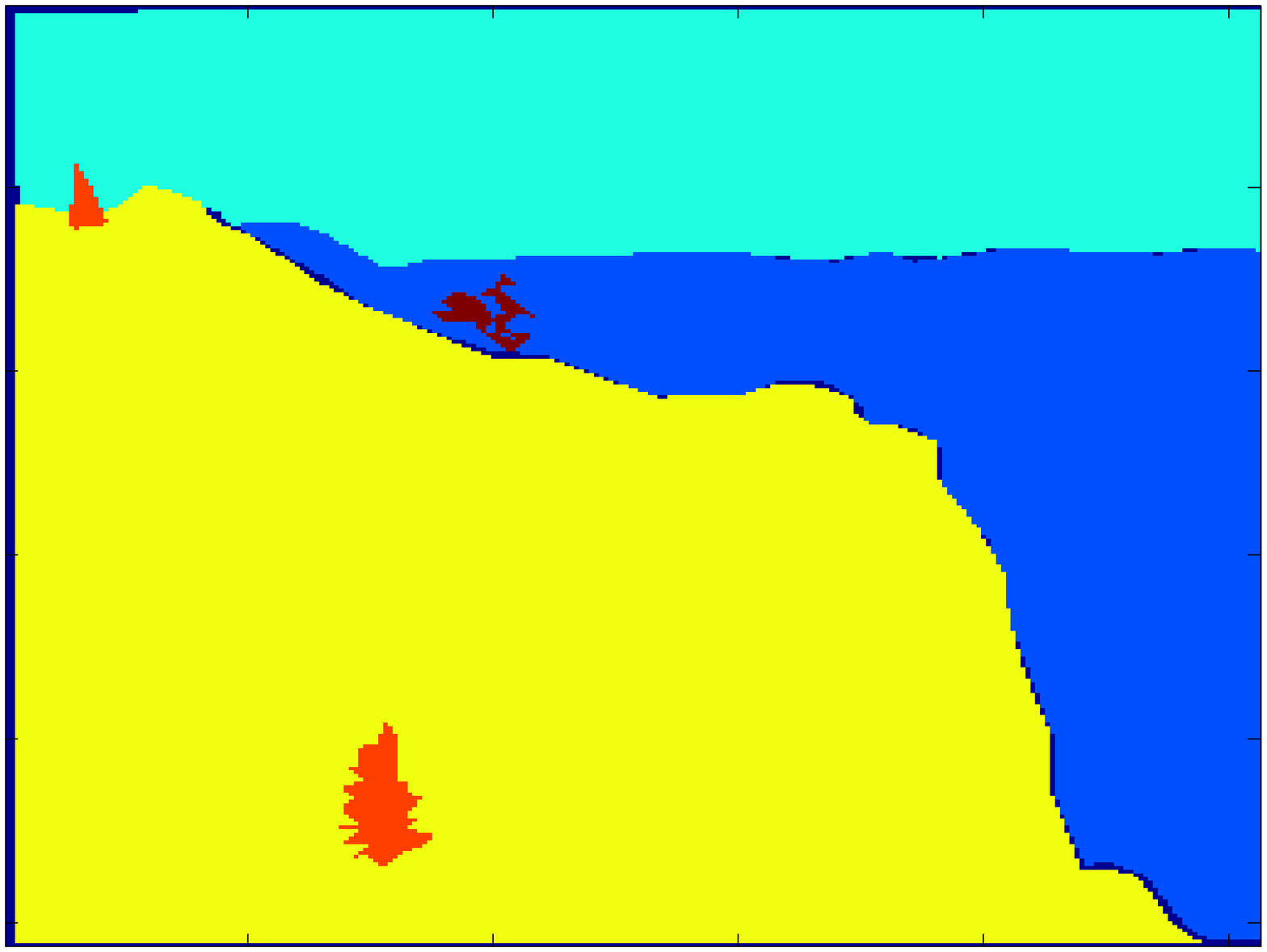}
\includegraphics[width=0.15\textwidth]{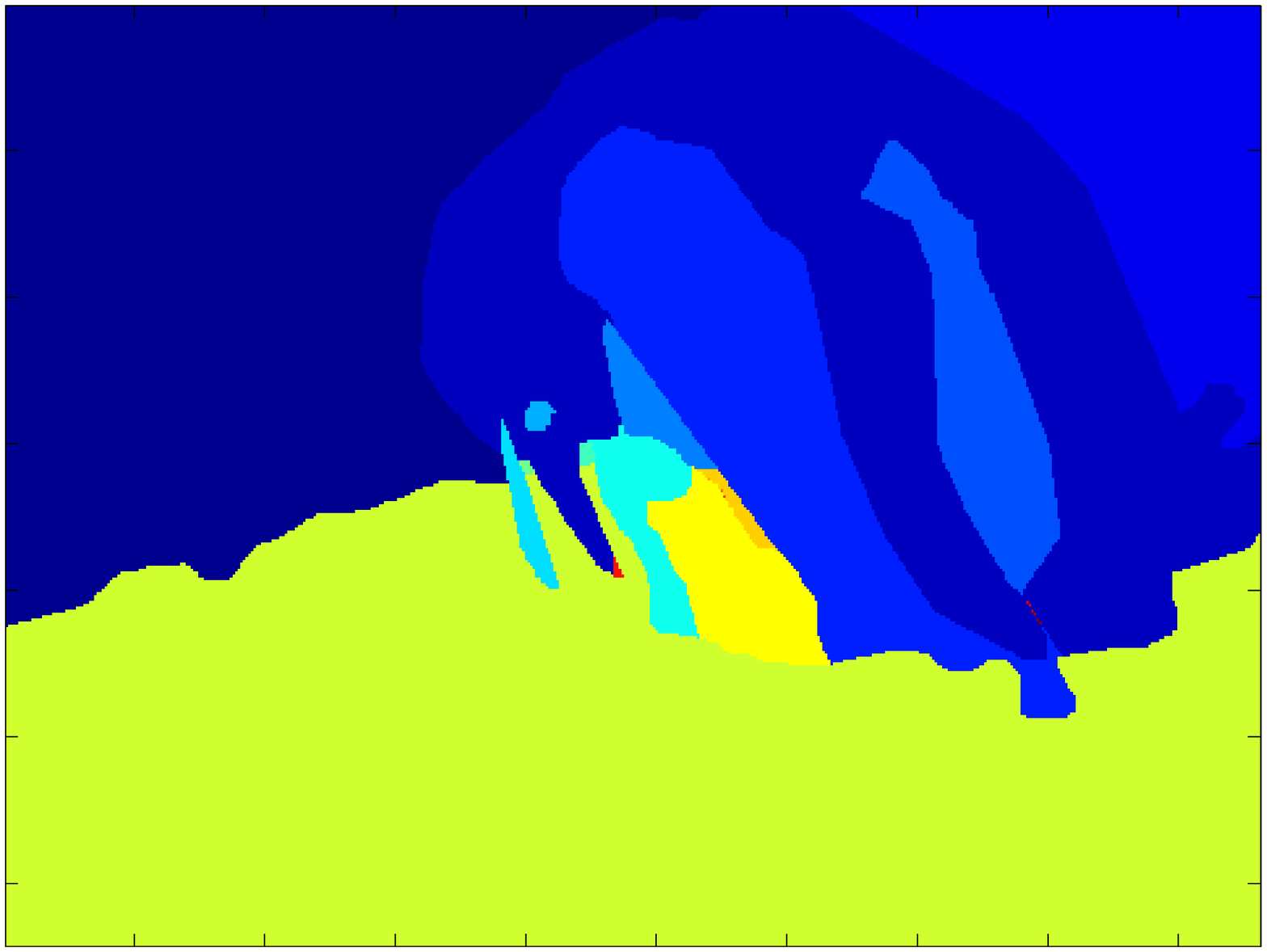}
\includegraphics[width=0.15\textwidth]{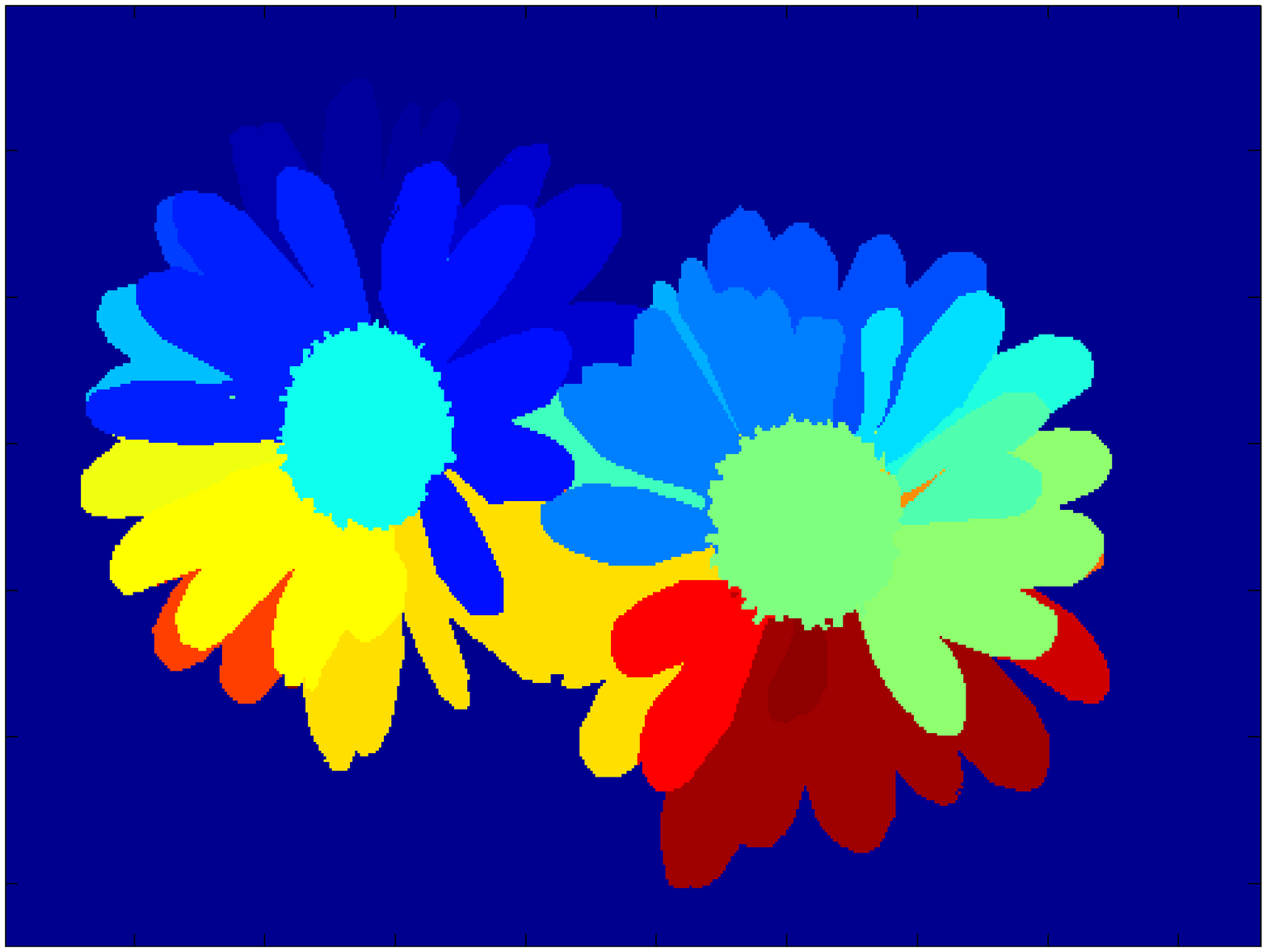}
\includegraphics[width=0.15\textwidth]{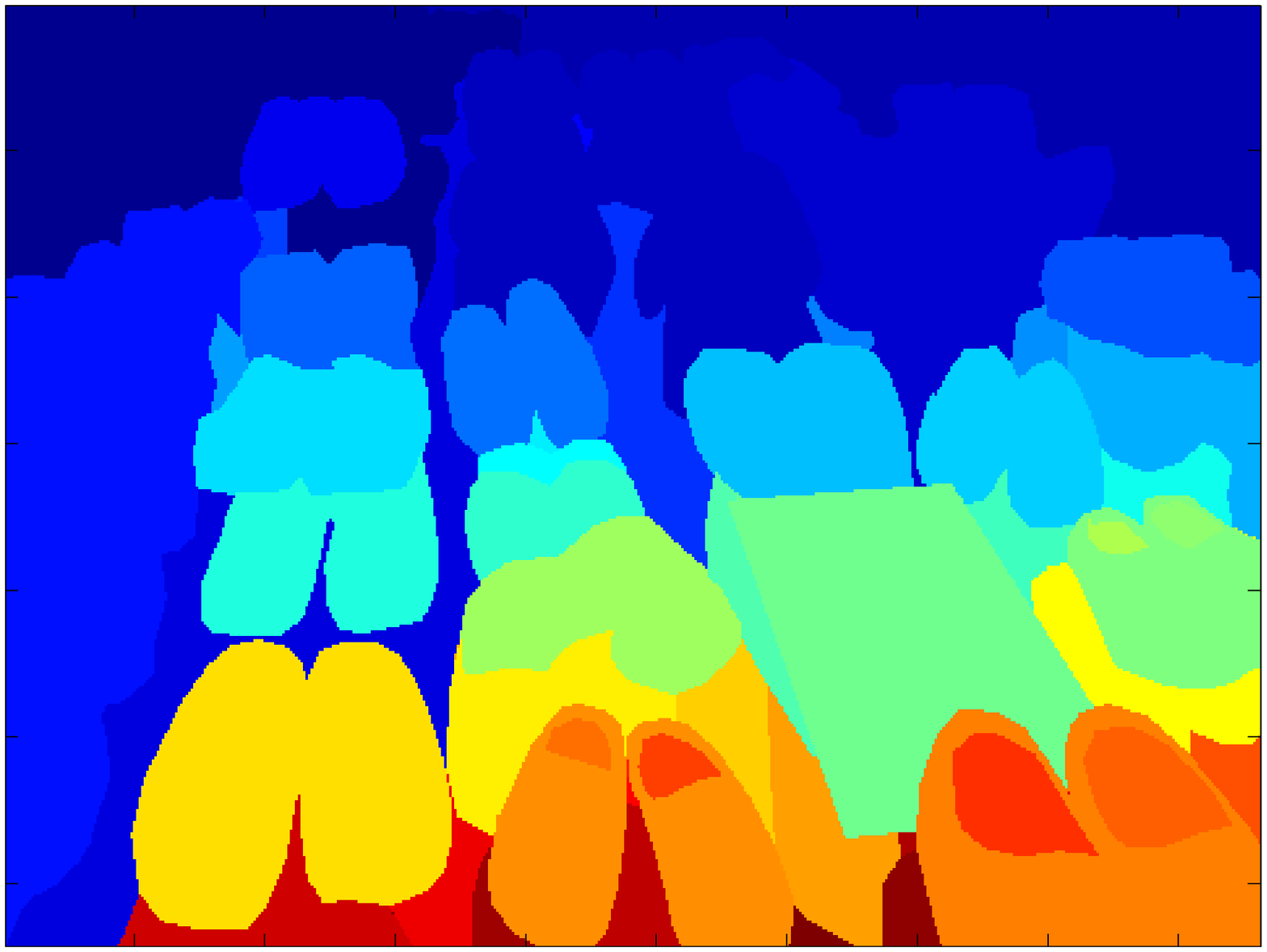}\\
 \includegraphics[width=0.15\textwidth]{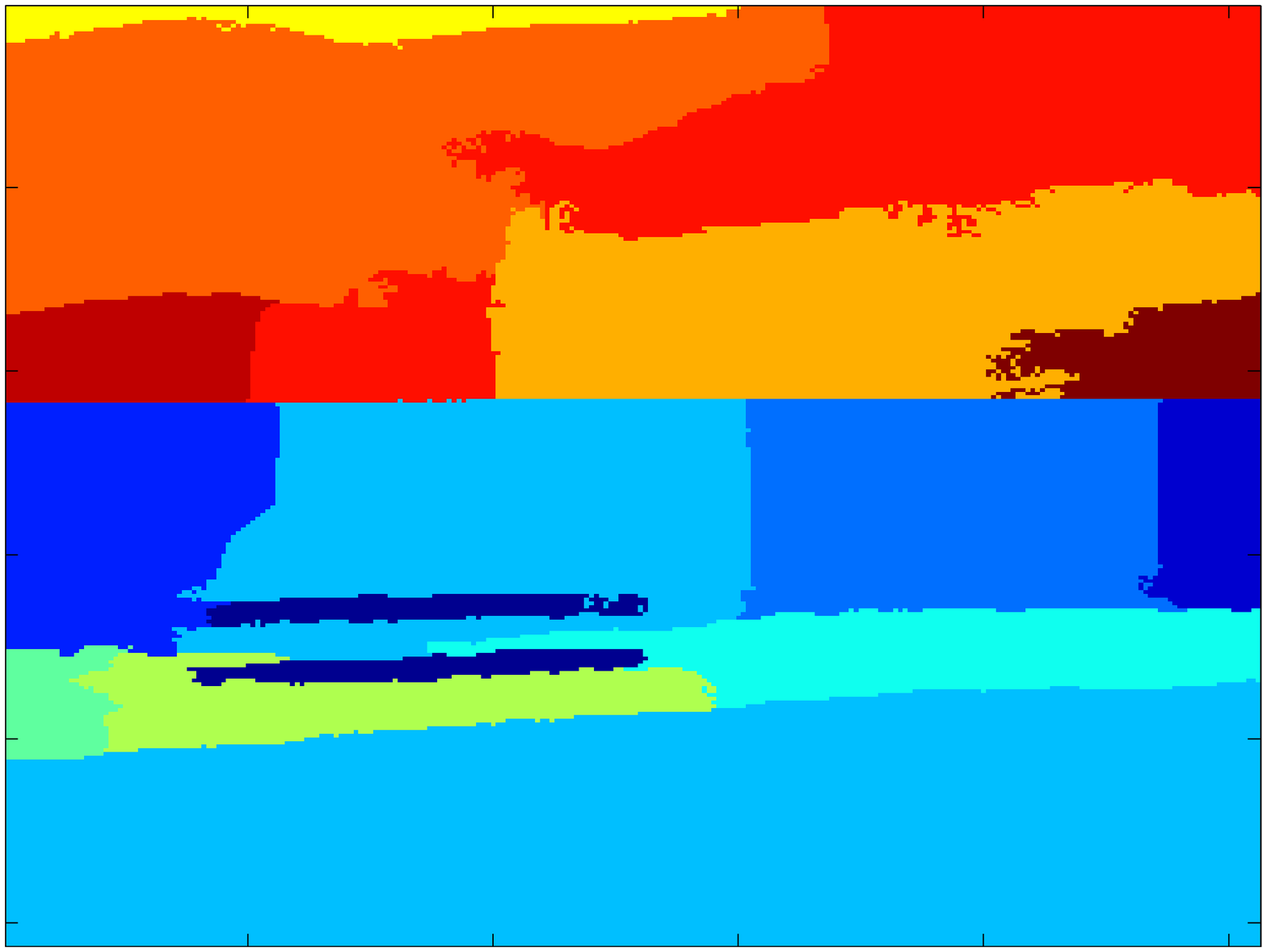}
\includegraphics[width=0.15\textwidth]{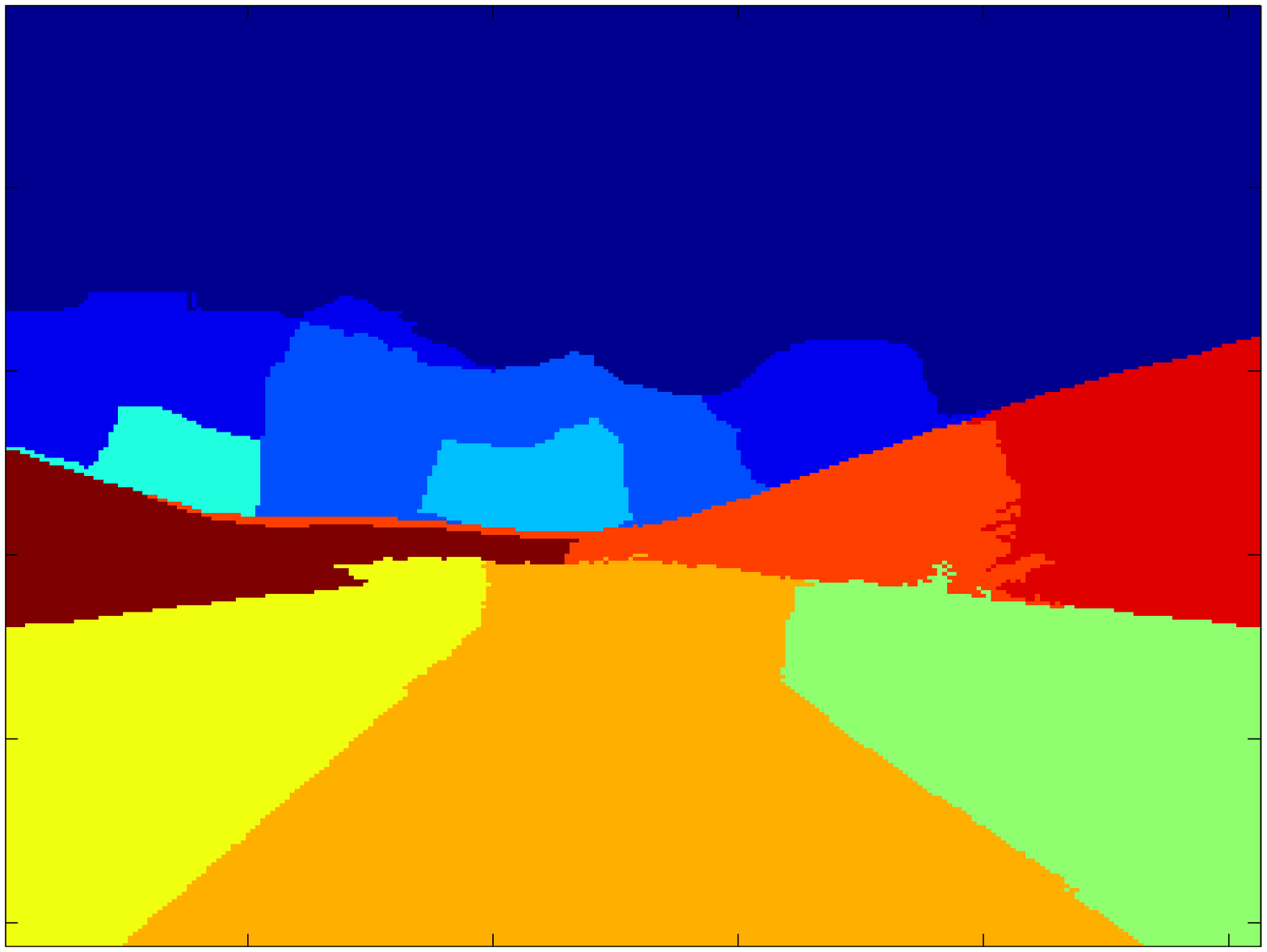}
\includegraphics[width=0.15\textwidth]{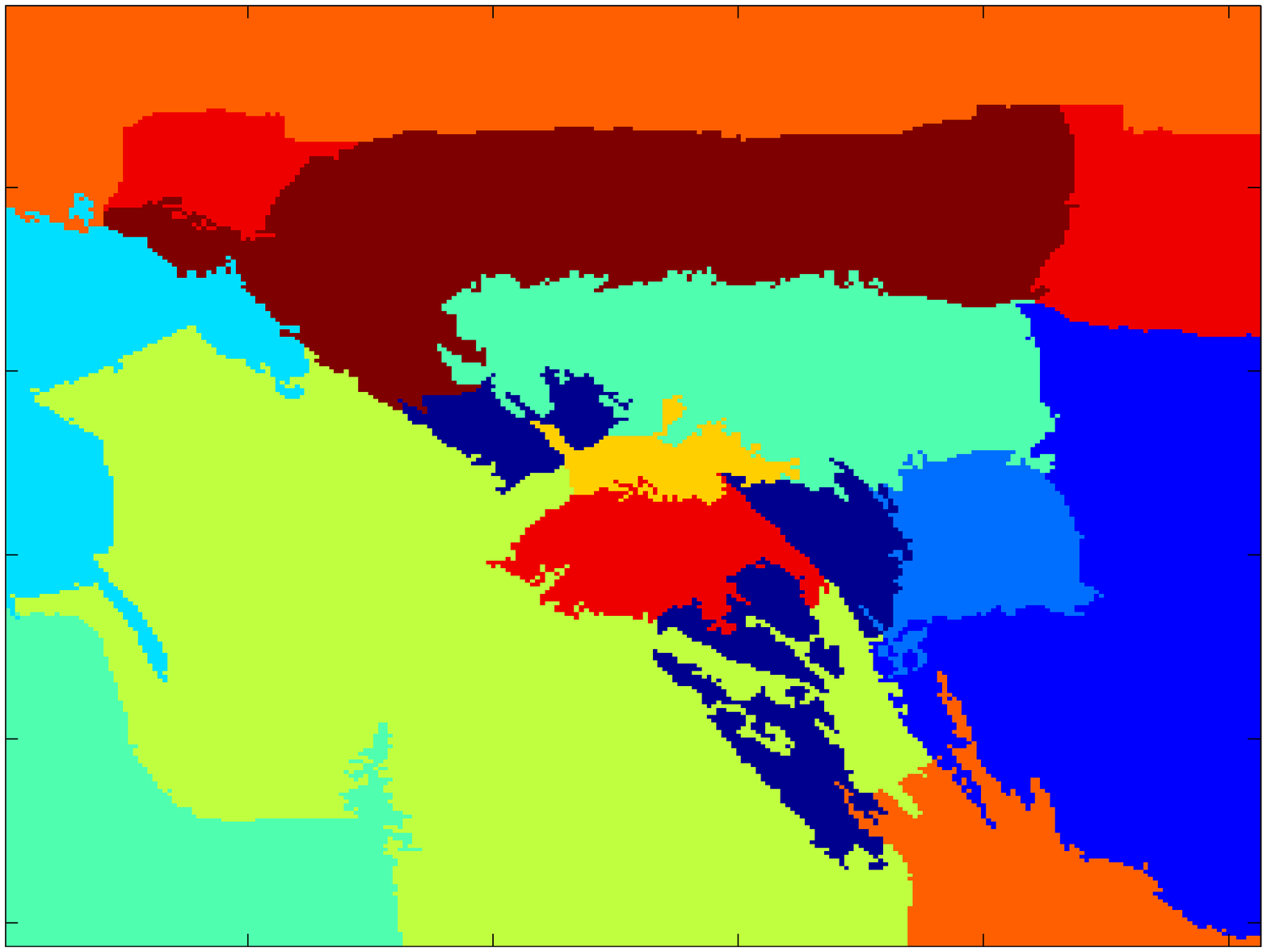}
\includegraphics[width=0.15\textwidth]{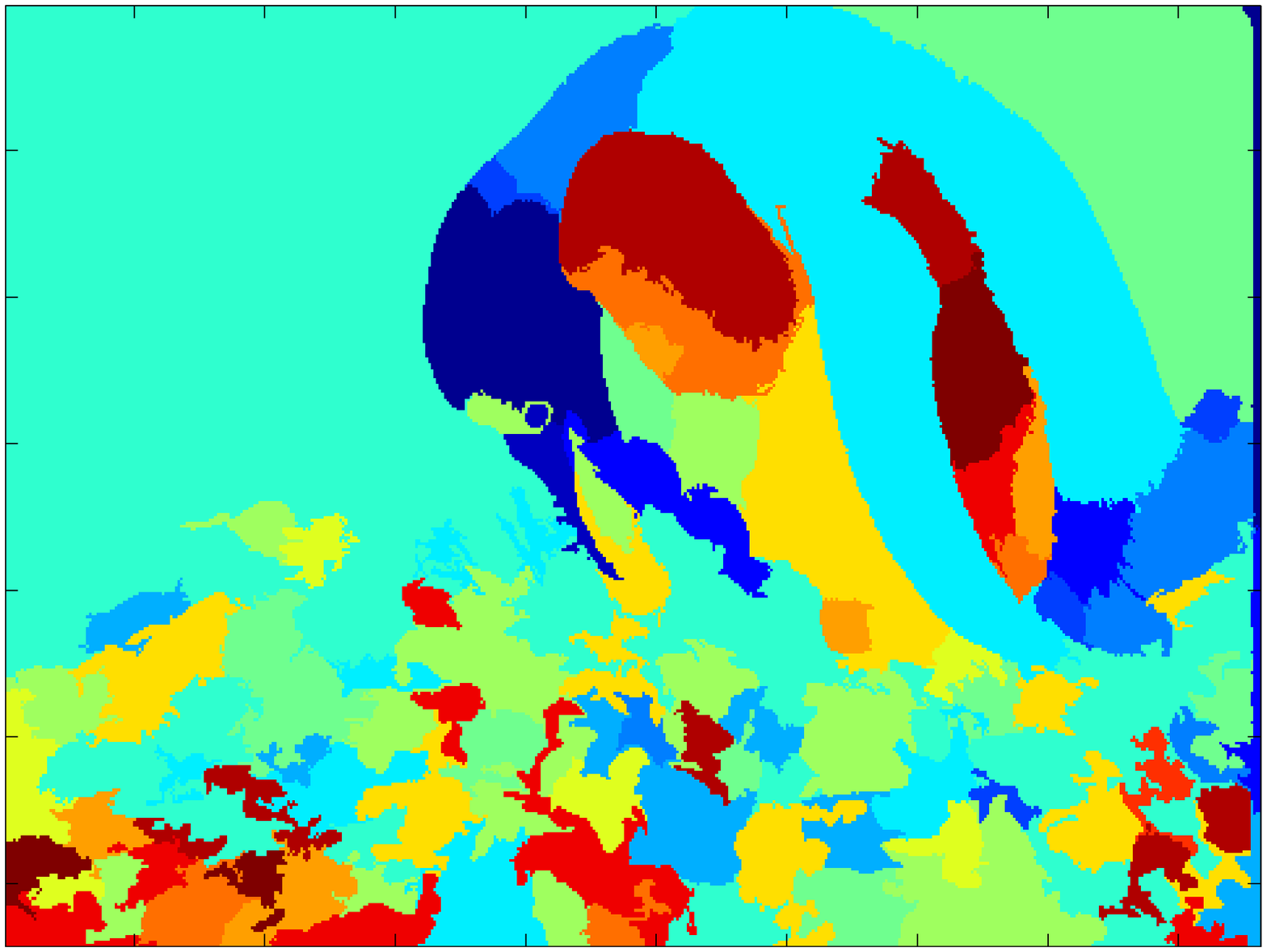}
\includegraphics[width=0.15\textwidth]{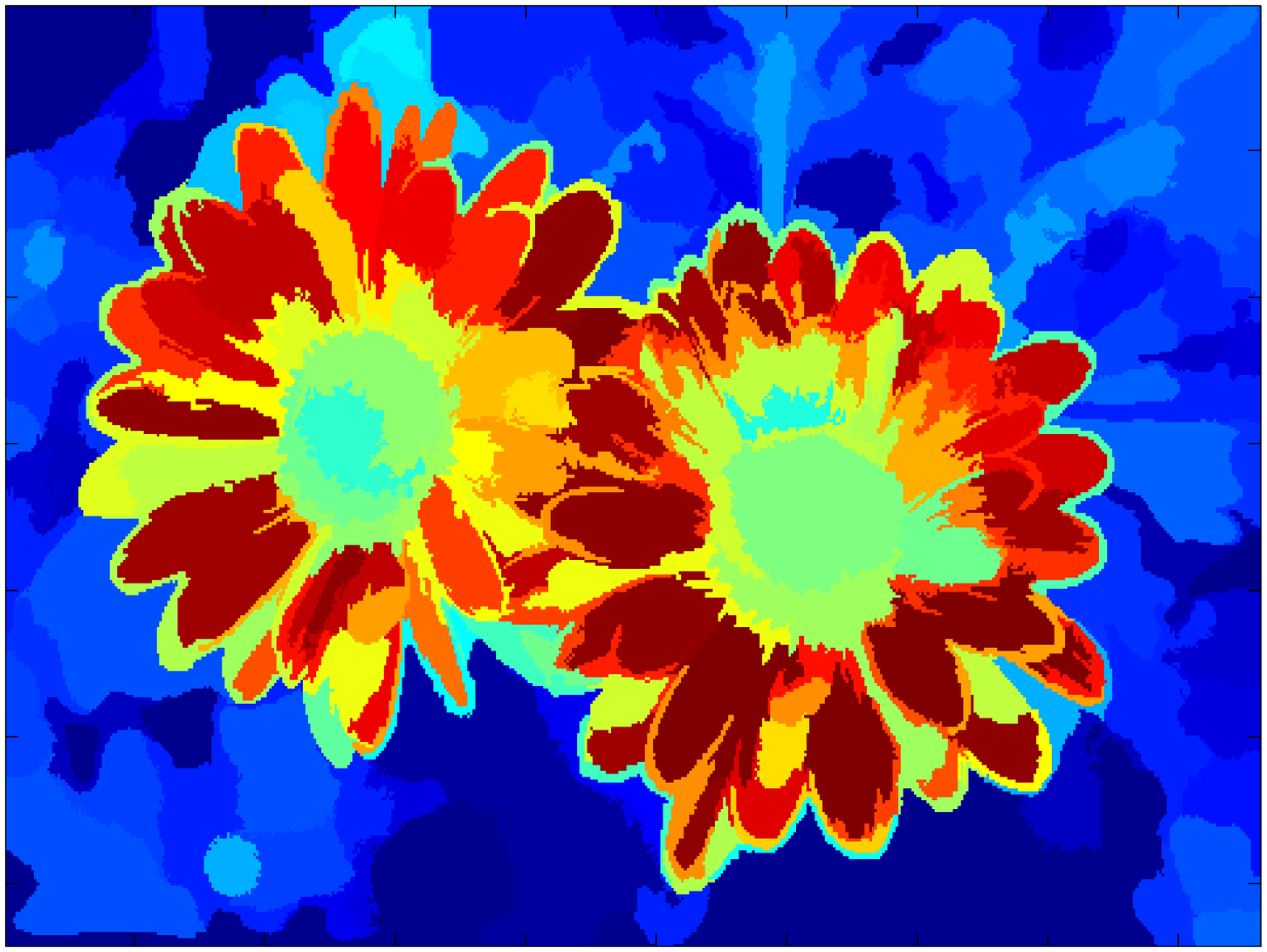}
\includegraphics[width=0.15\textwidth]{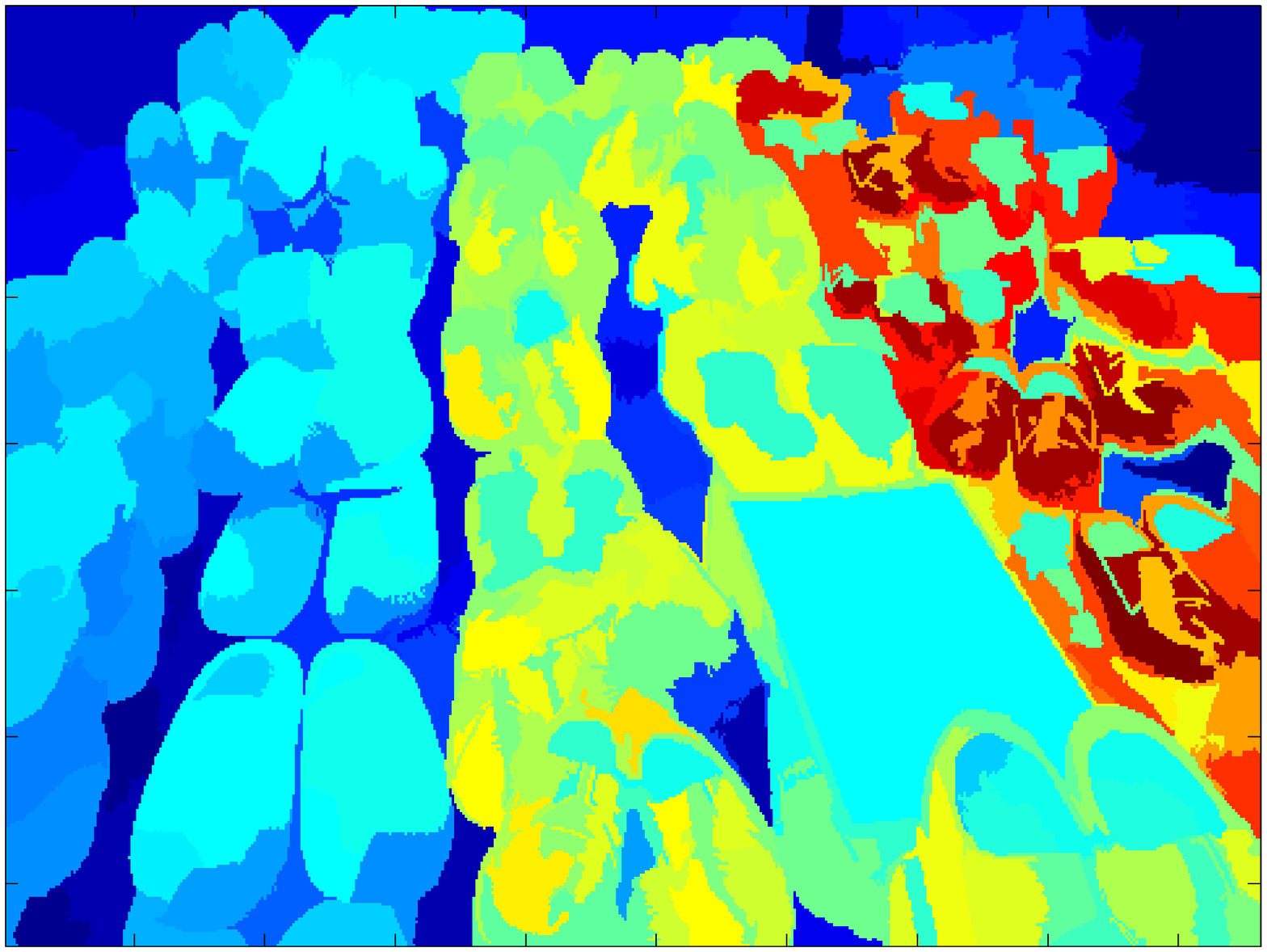}\\
 \includegraphics[width=0.15\textwidth]{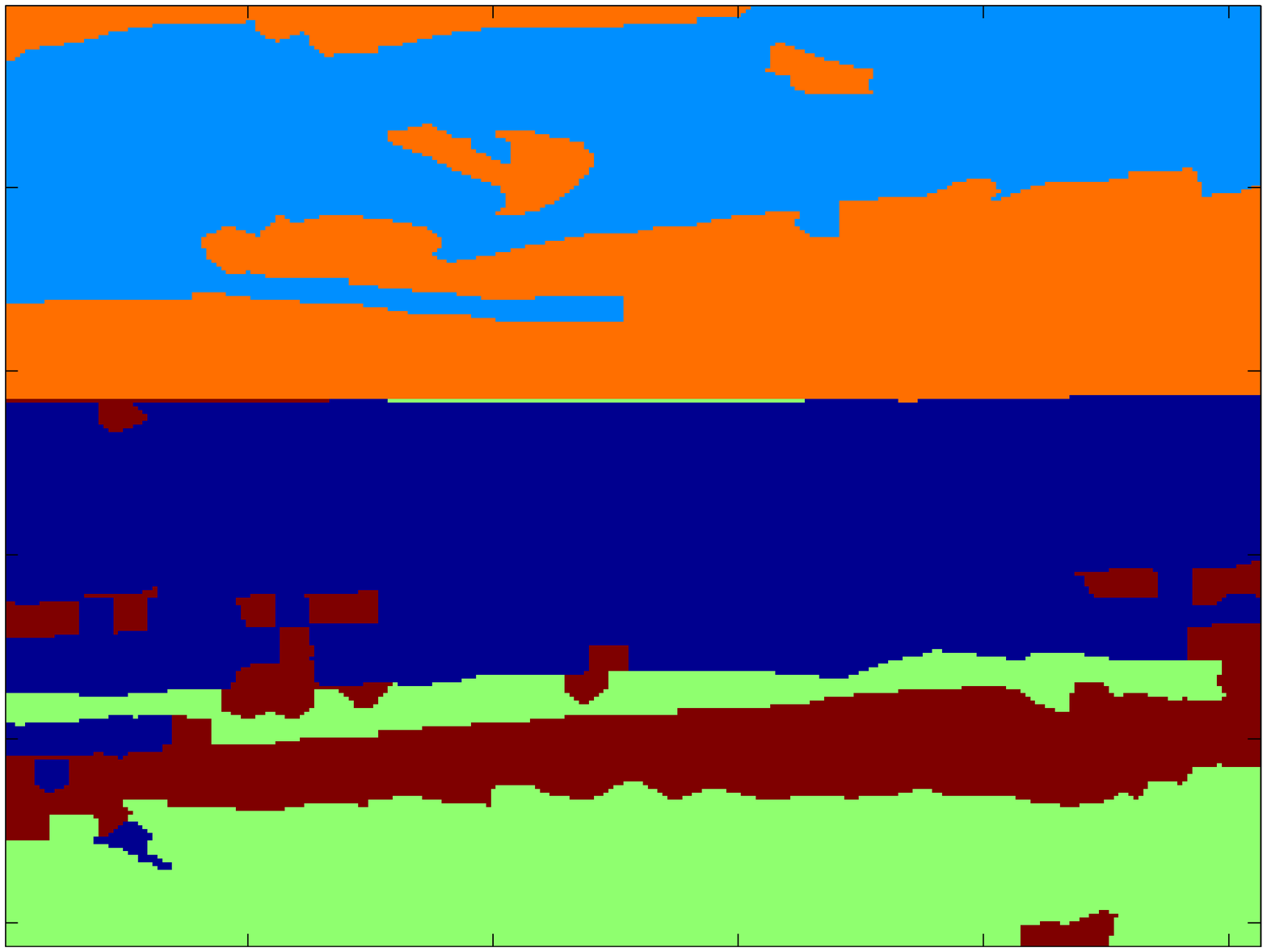}
\includegraphics[width=0.15\textwidth]{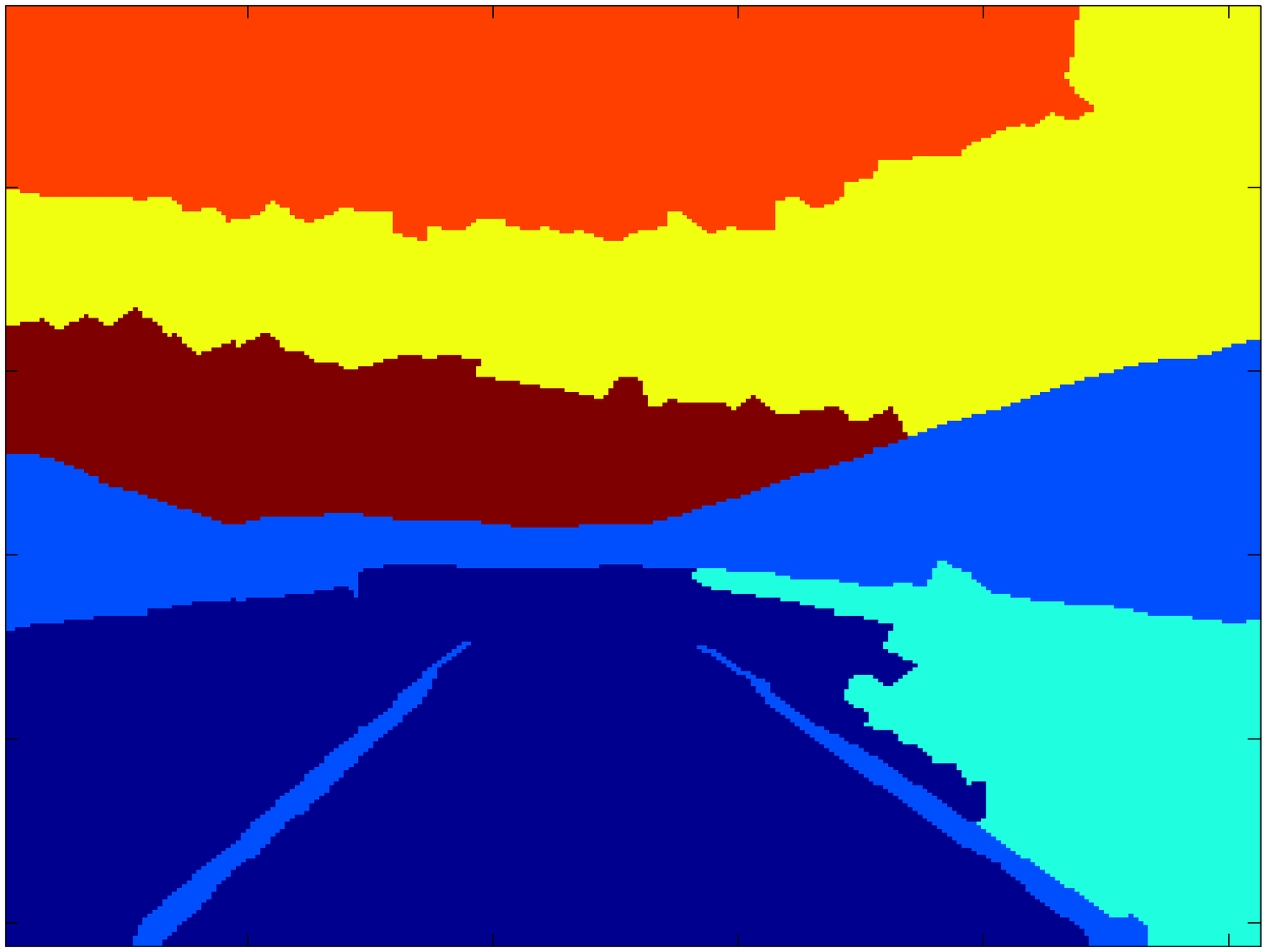}
\includegraphics[width=0.15\textwidth]{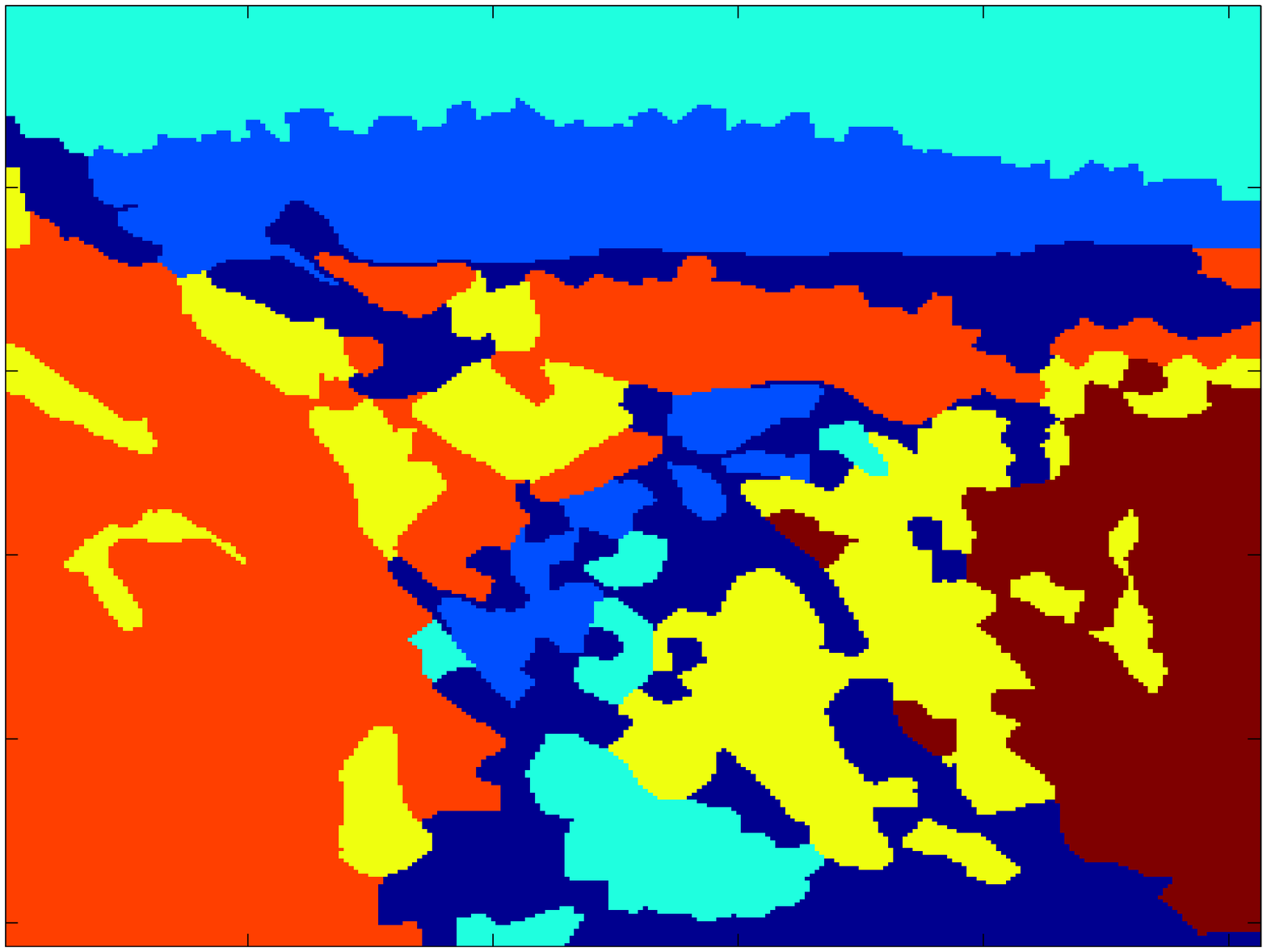}
\includegraphics[width=0.15\textwidth]{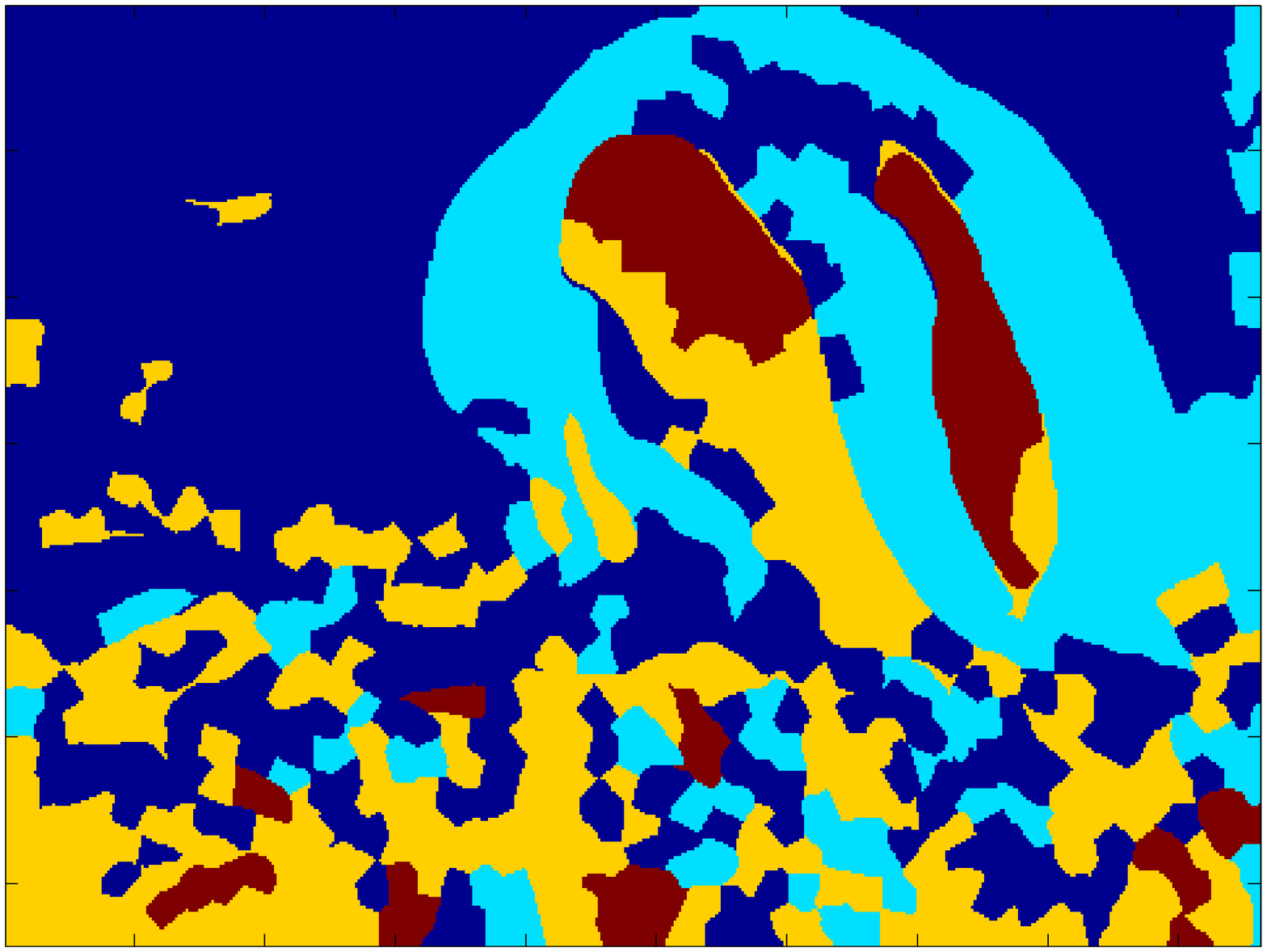}
\includegraphics[width=0.15\textwidth]{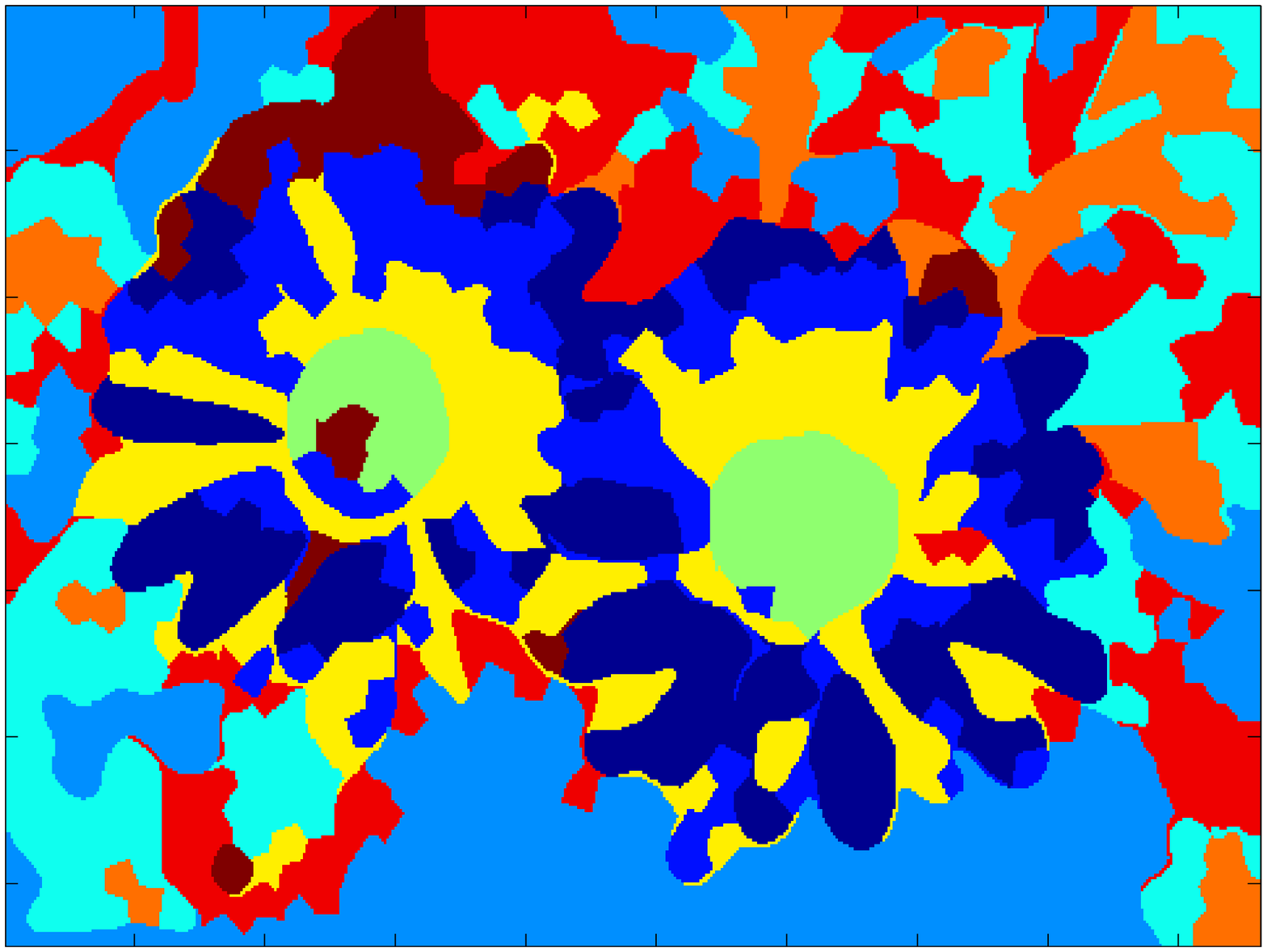}
\includegraphics[width=0.15\textwidth]{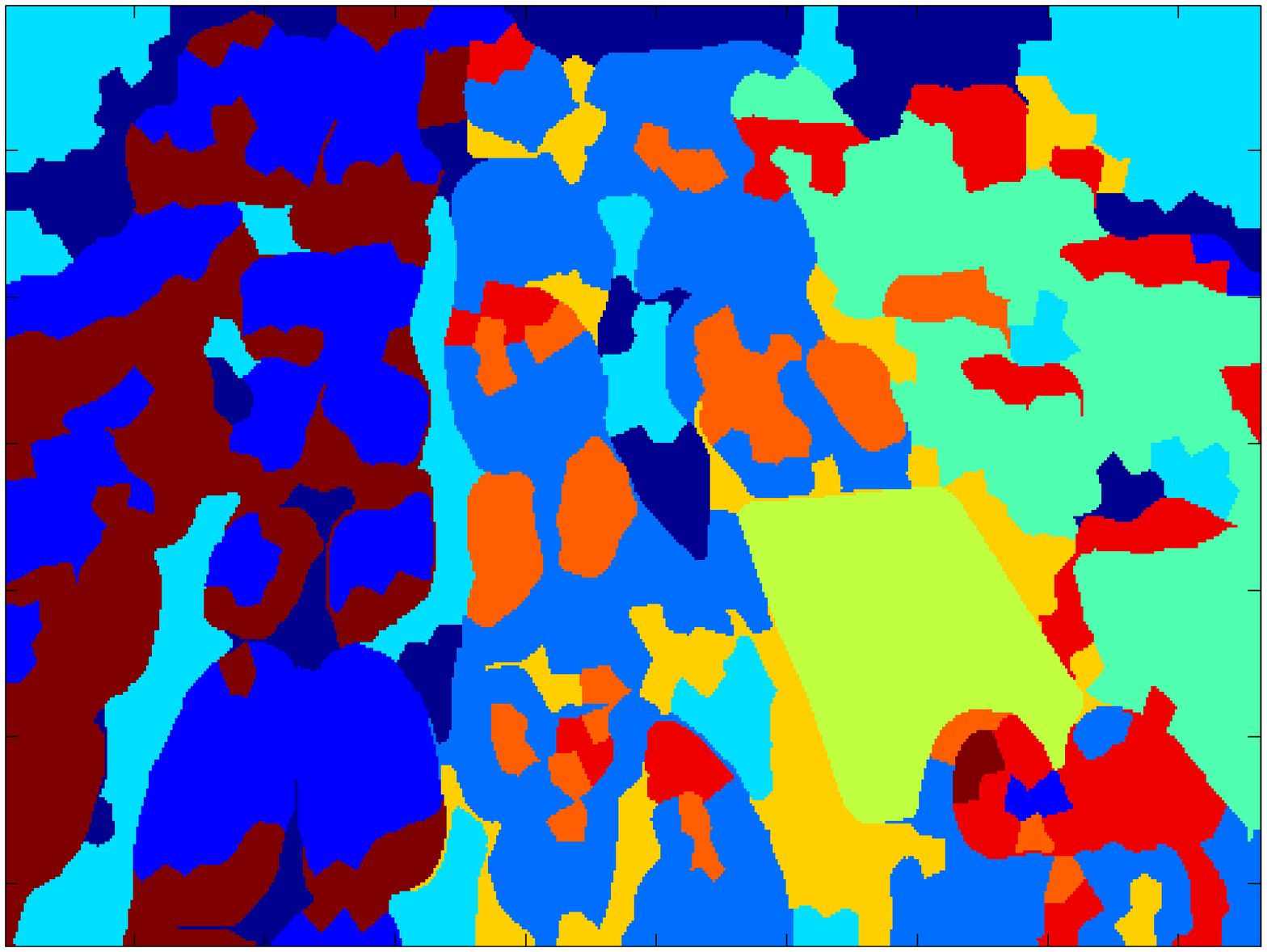}\\
 \includegraphics[width=0.15\textwidth]{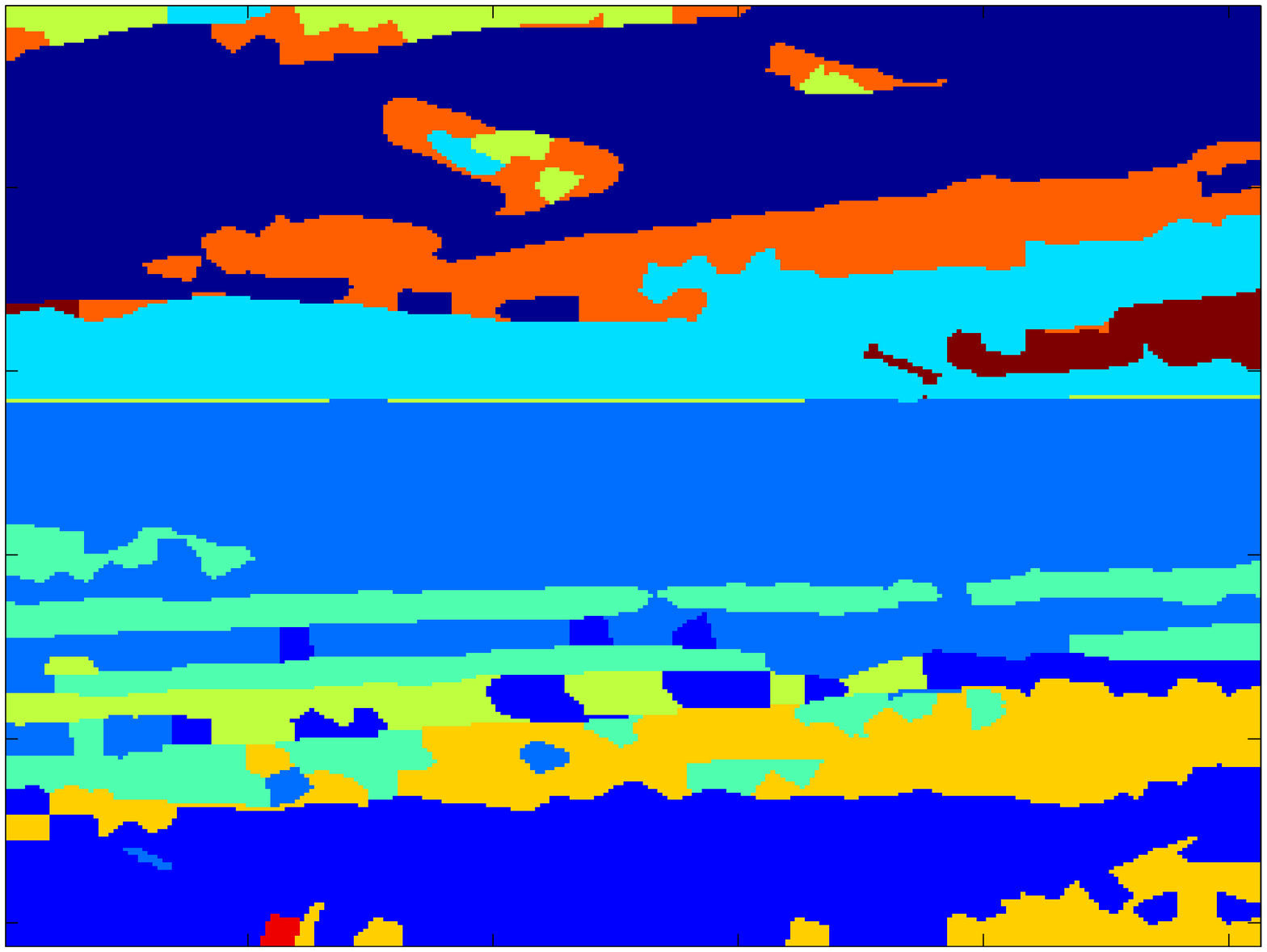}
\includegraphics[width=0.15\textwidth]{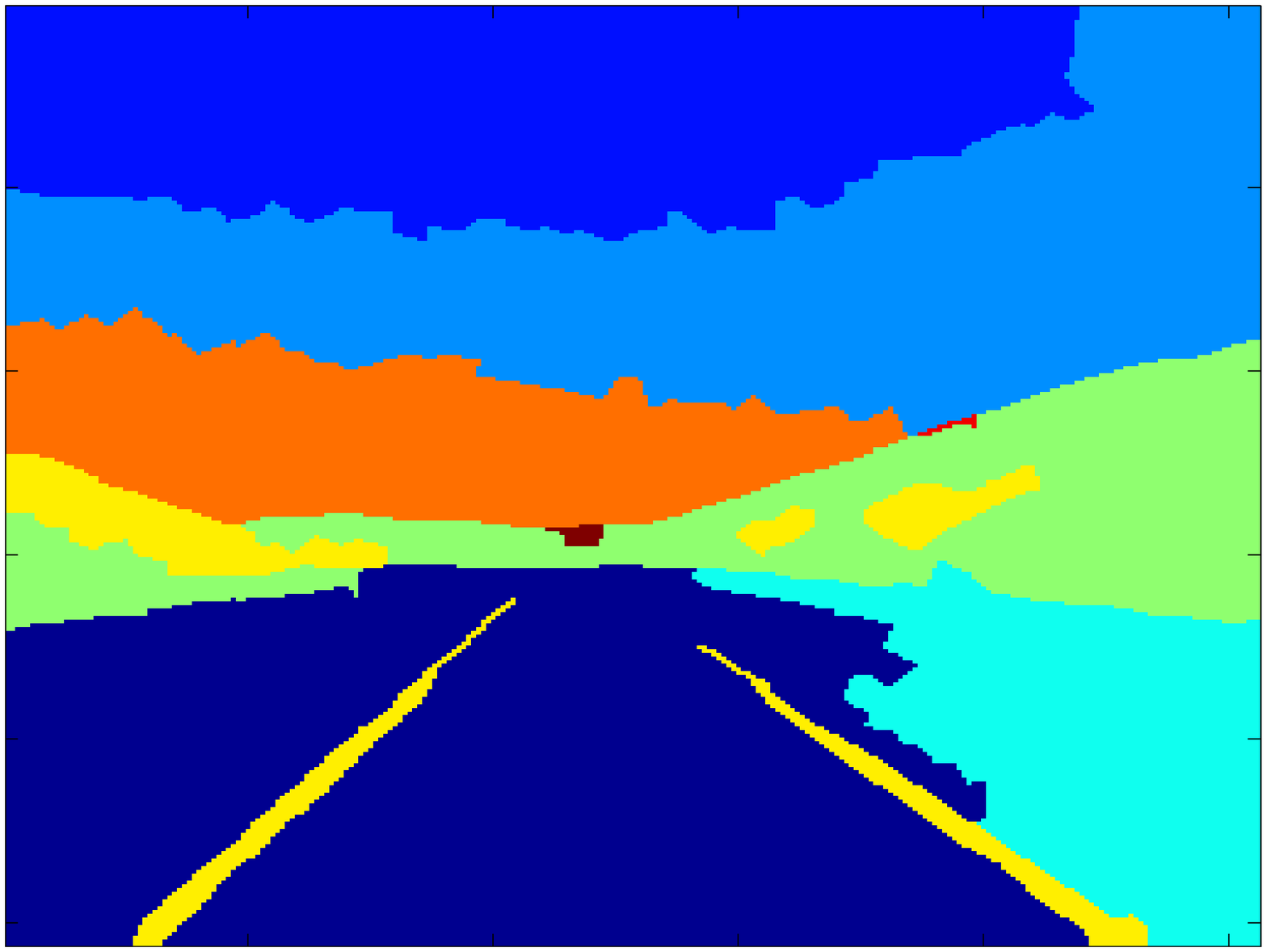}
\includegraphics[width=0.15\textwidth]{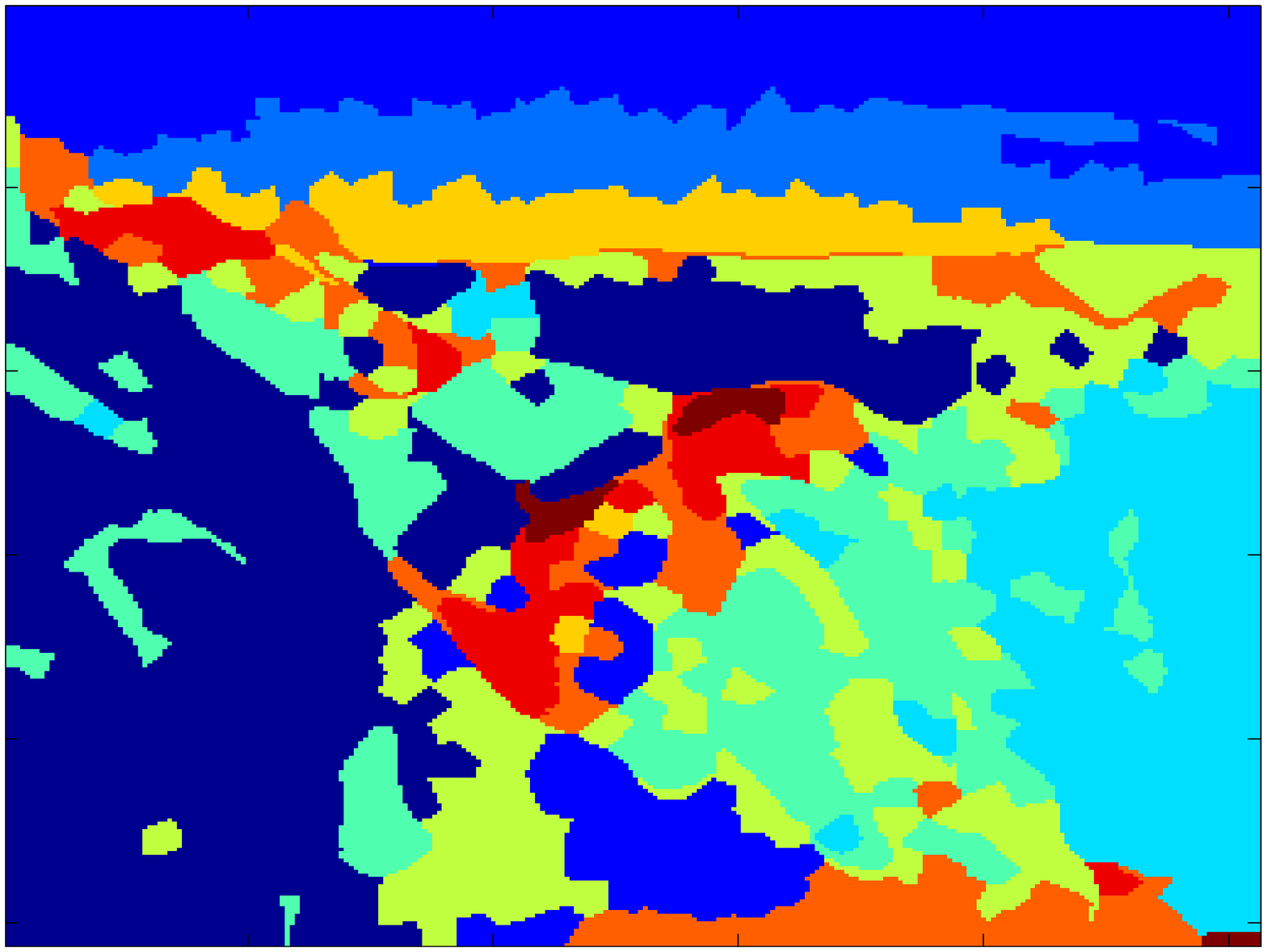}
\includegraphics[width=0.15\textwidth]{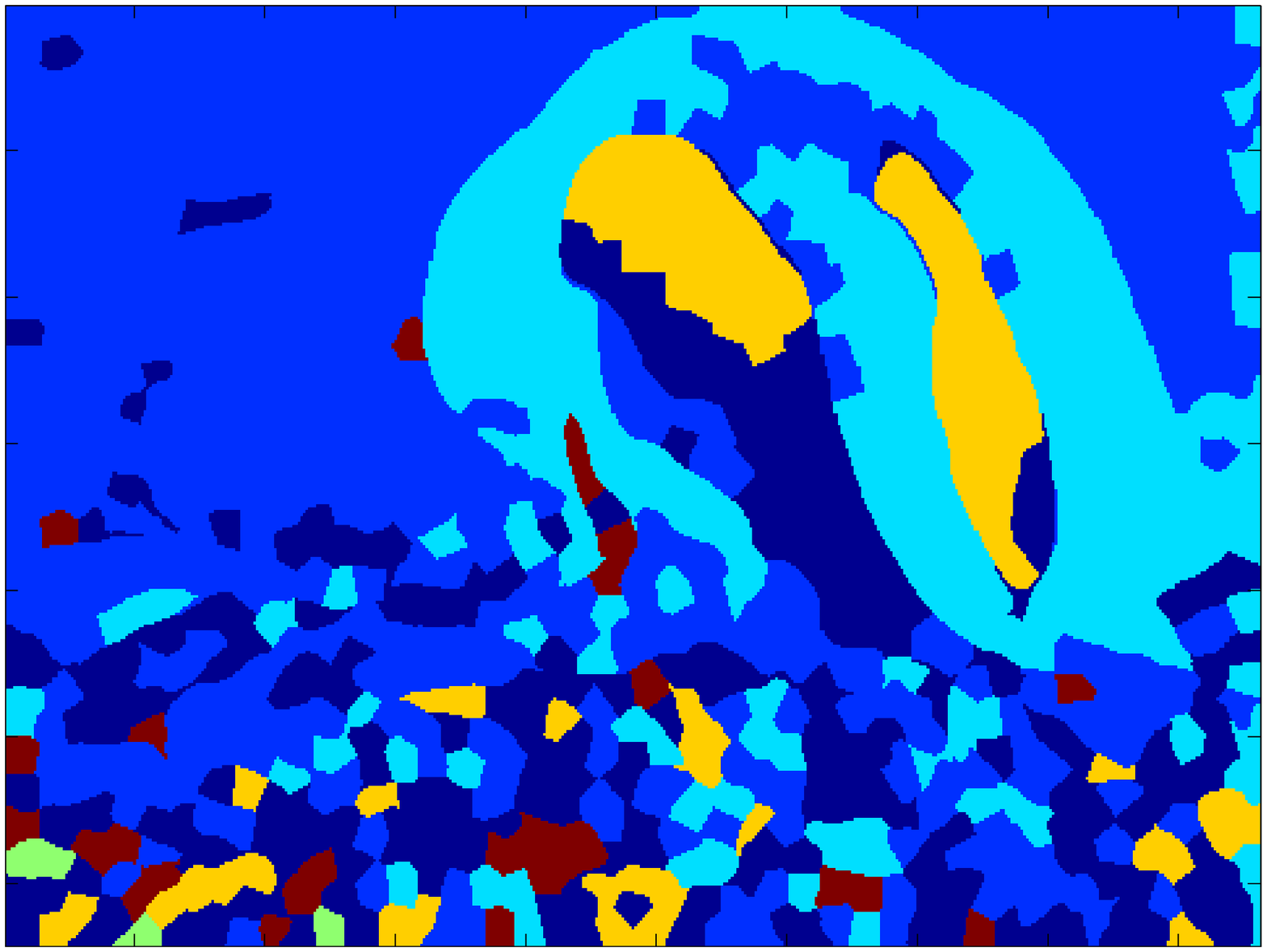}
\includegraphics[width=0.15\textwidth]{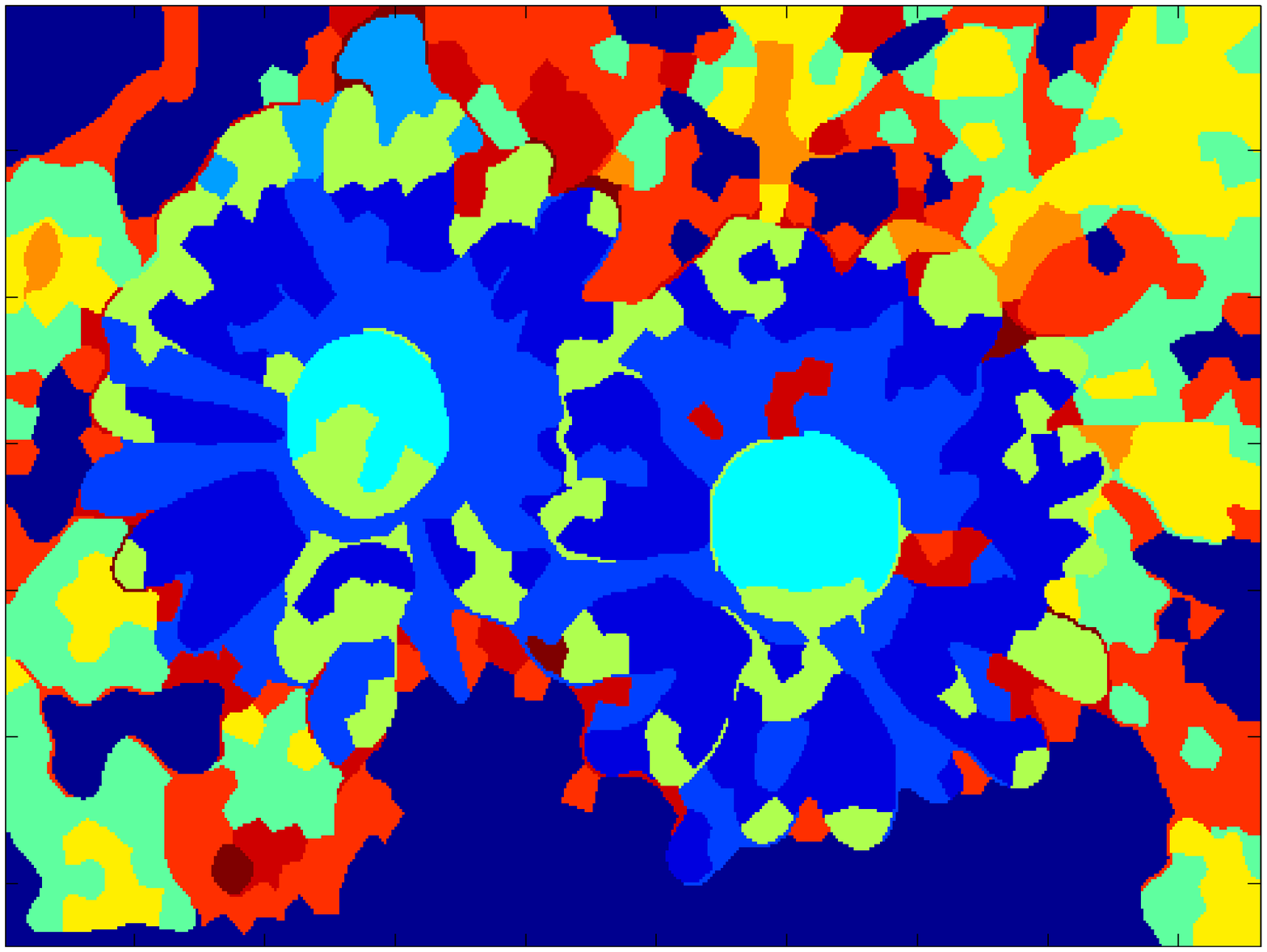}
\includegraphics[width=0.15\textwidth]{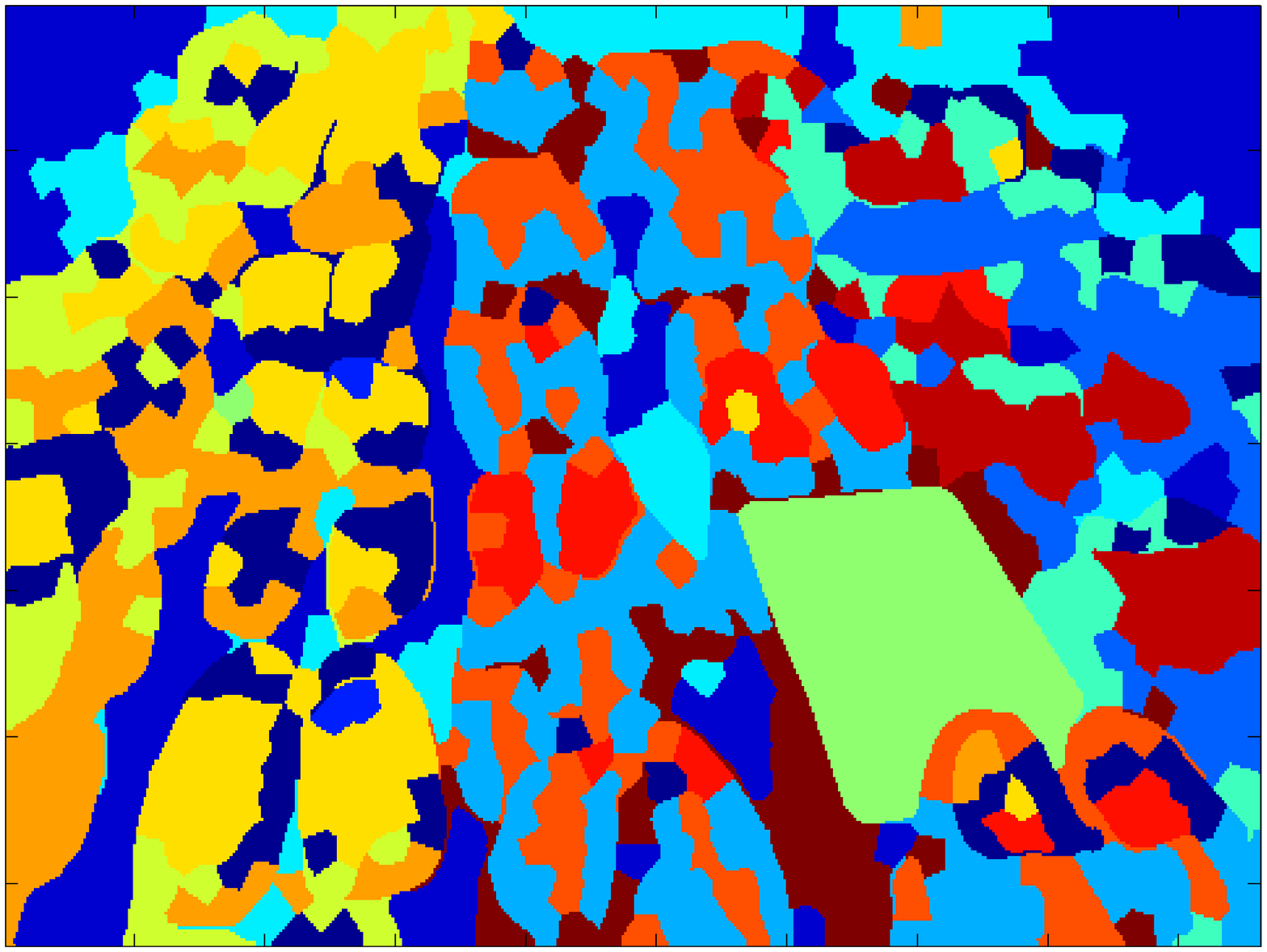}\\
 \includegraphics[width=0.15\textwidth]{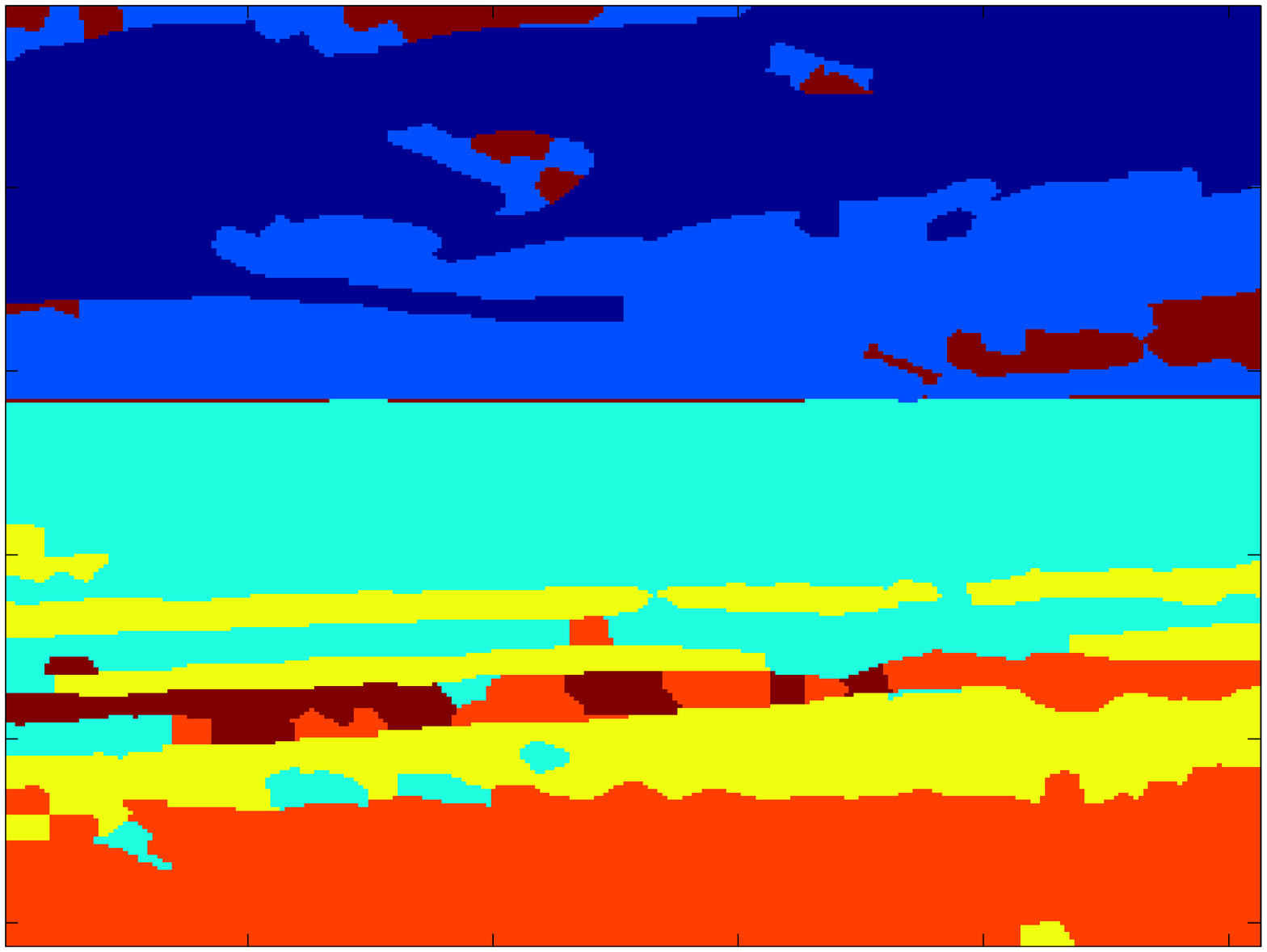}
\includegraphics[width=0.15\textwidth]{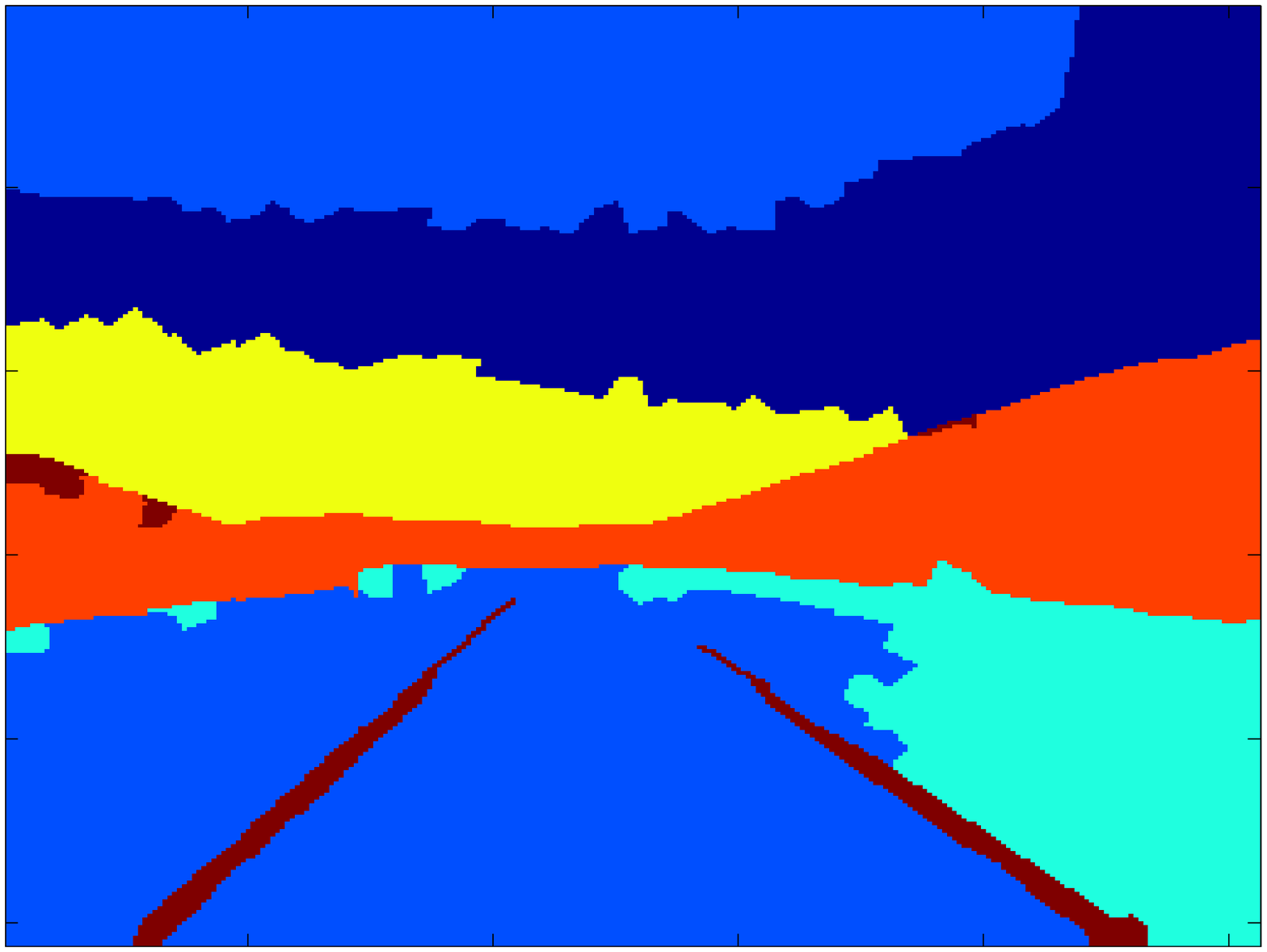}
\includegraphics[width=0.15\textwidth]{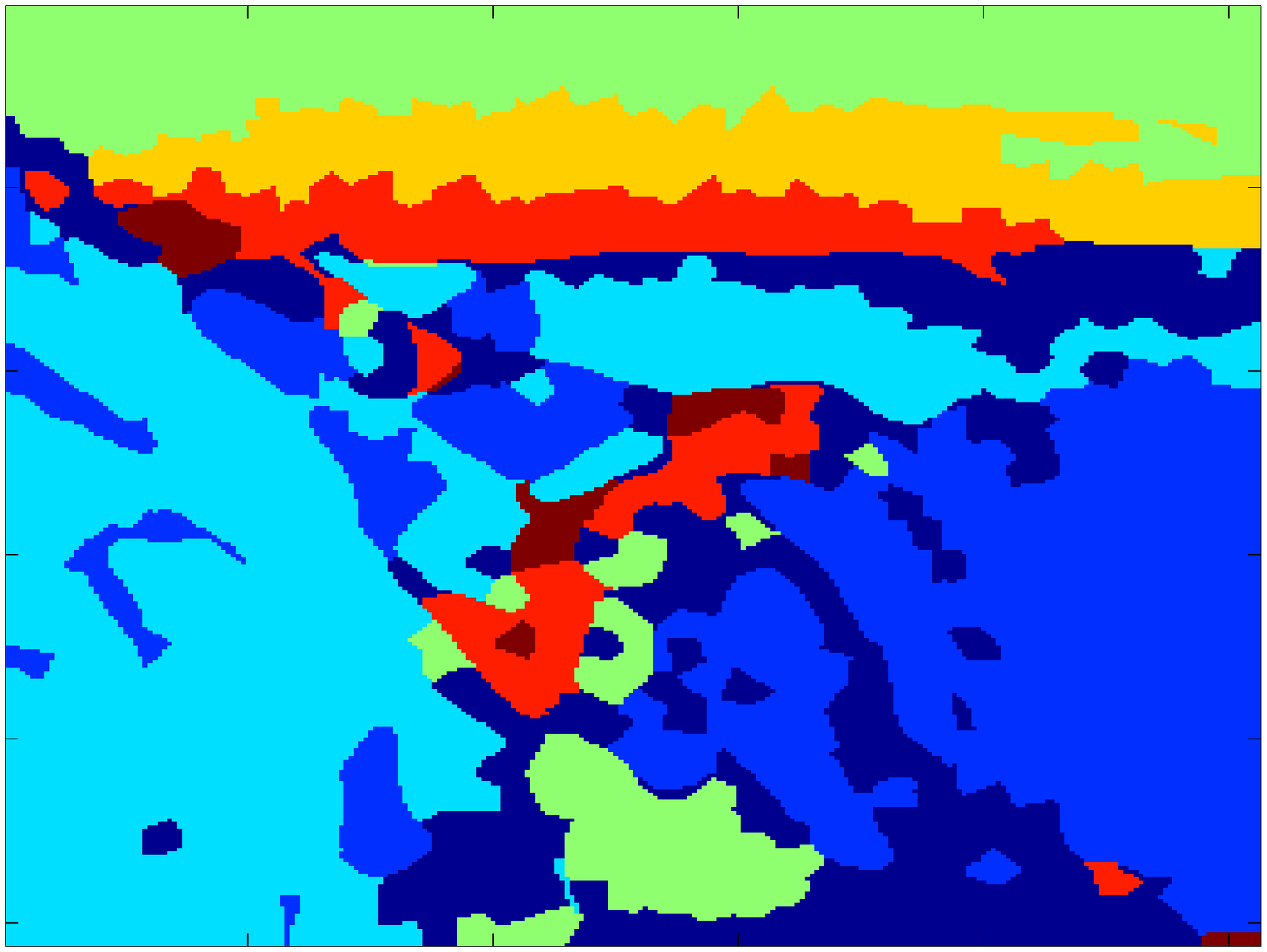}
\includegraphics[width=0.15\textwidth]{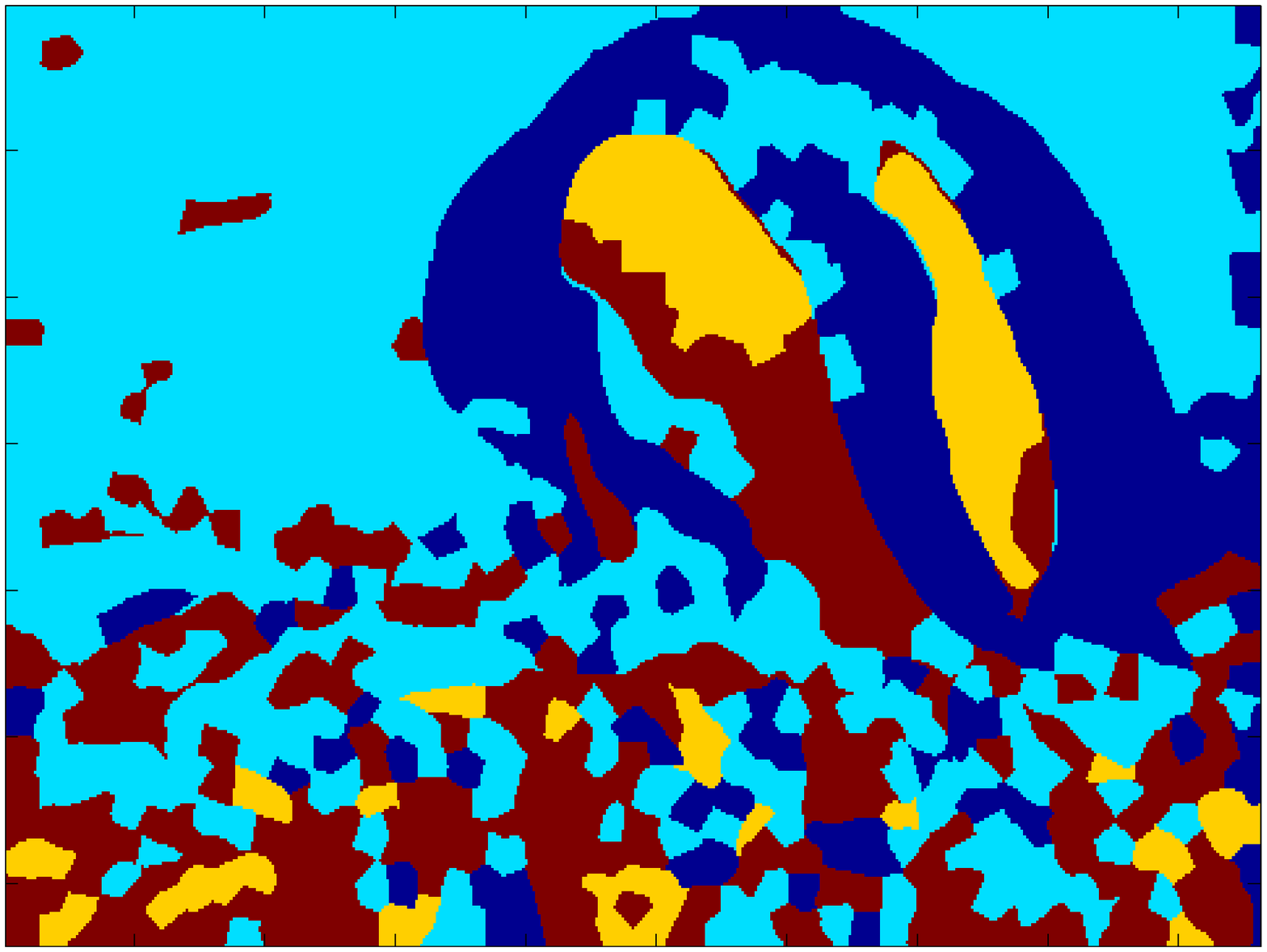}
\includegraphics[width=0.15\textwidth]{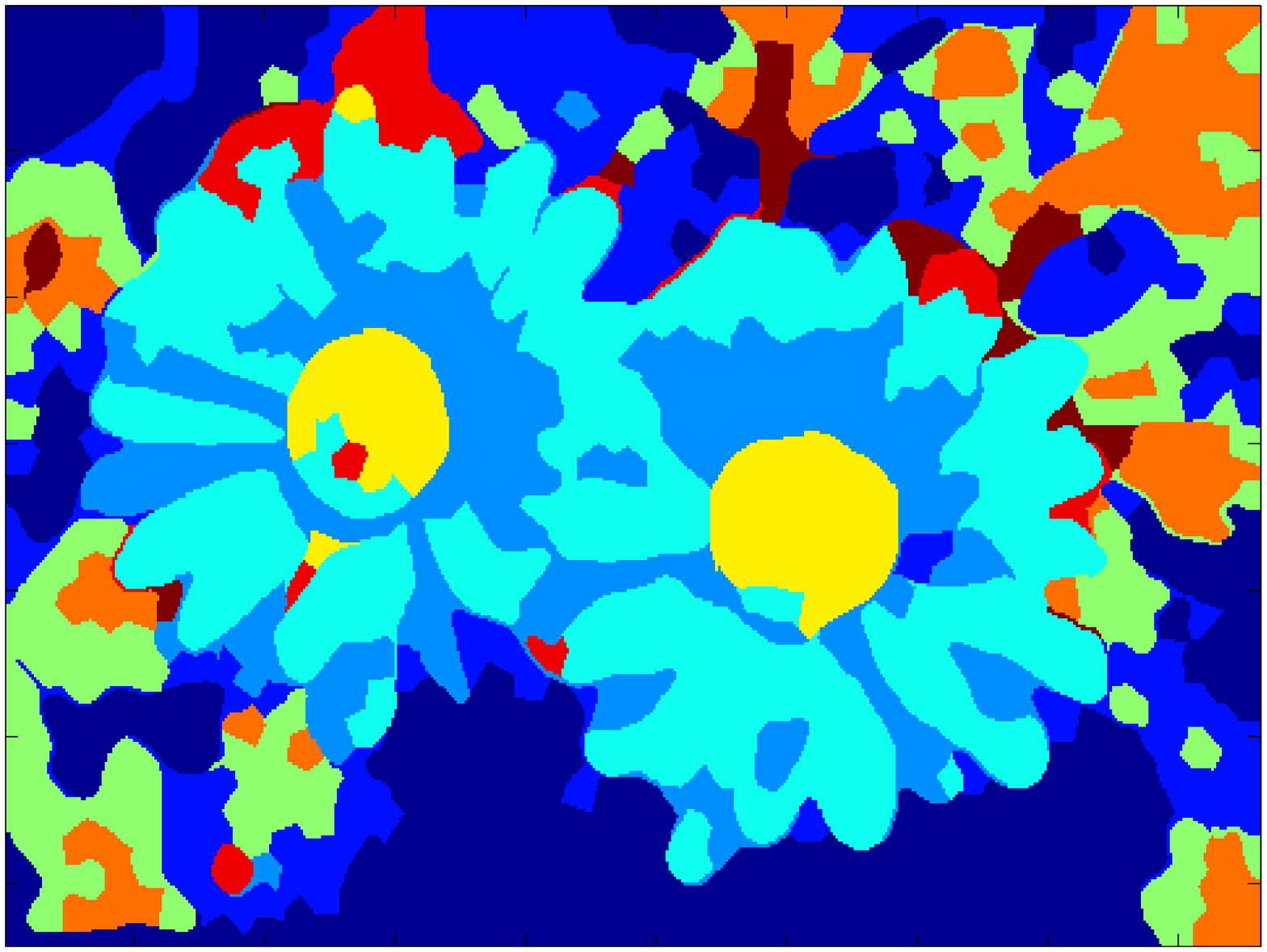}
\includegraphics[width=0.15\textwidth]{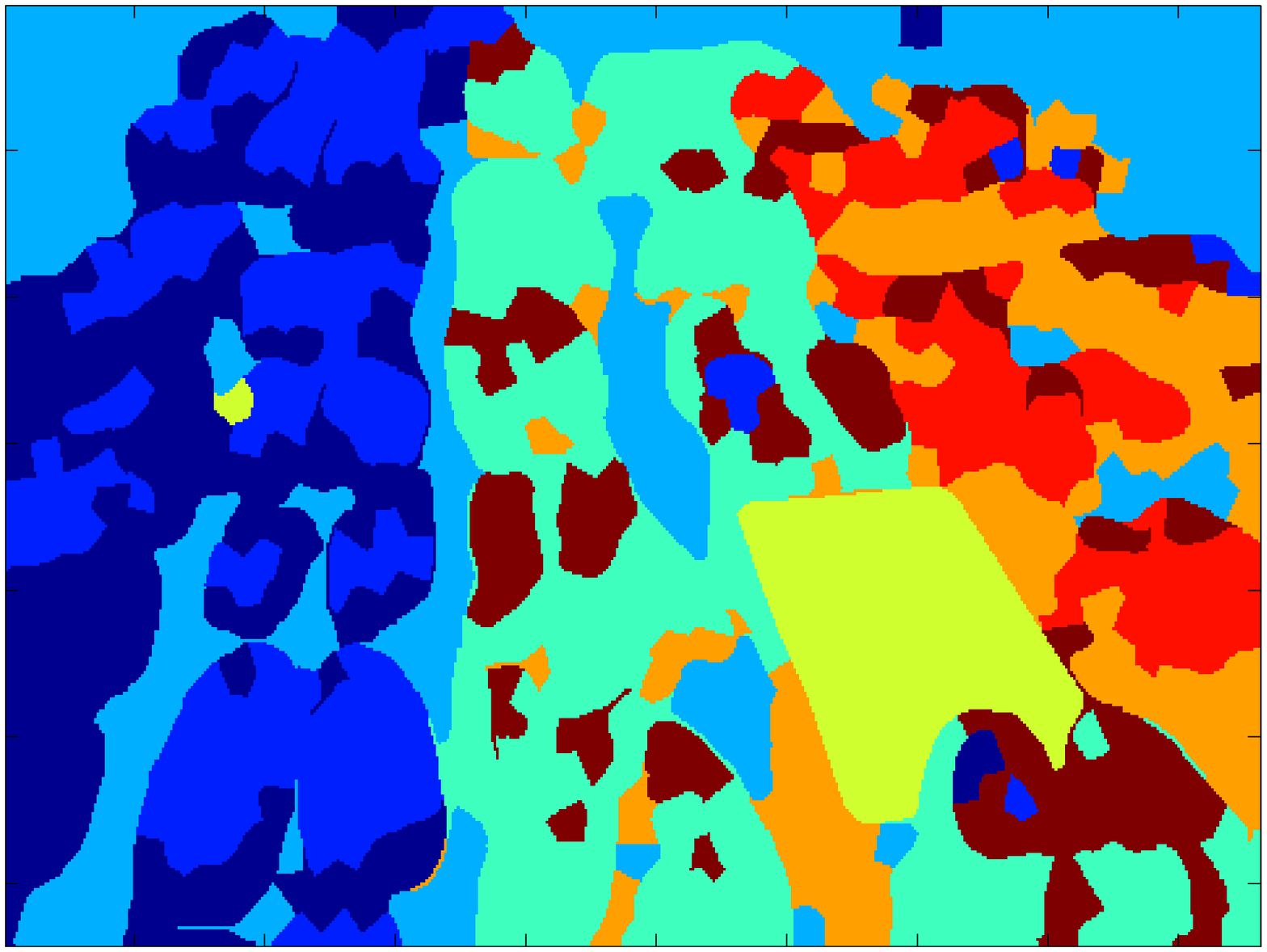}\\
 \includegraphics[width=0.15\textwidth]{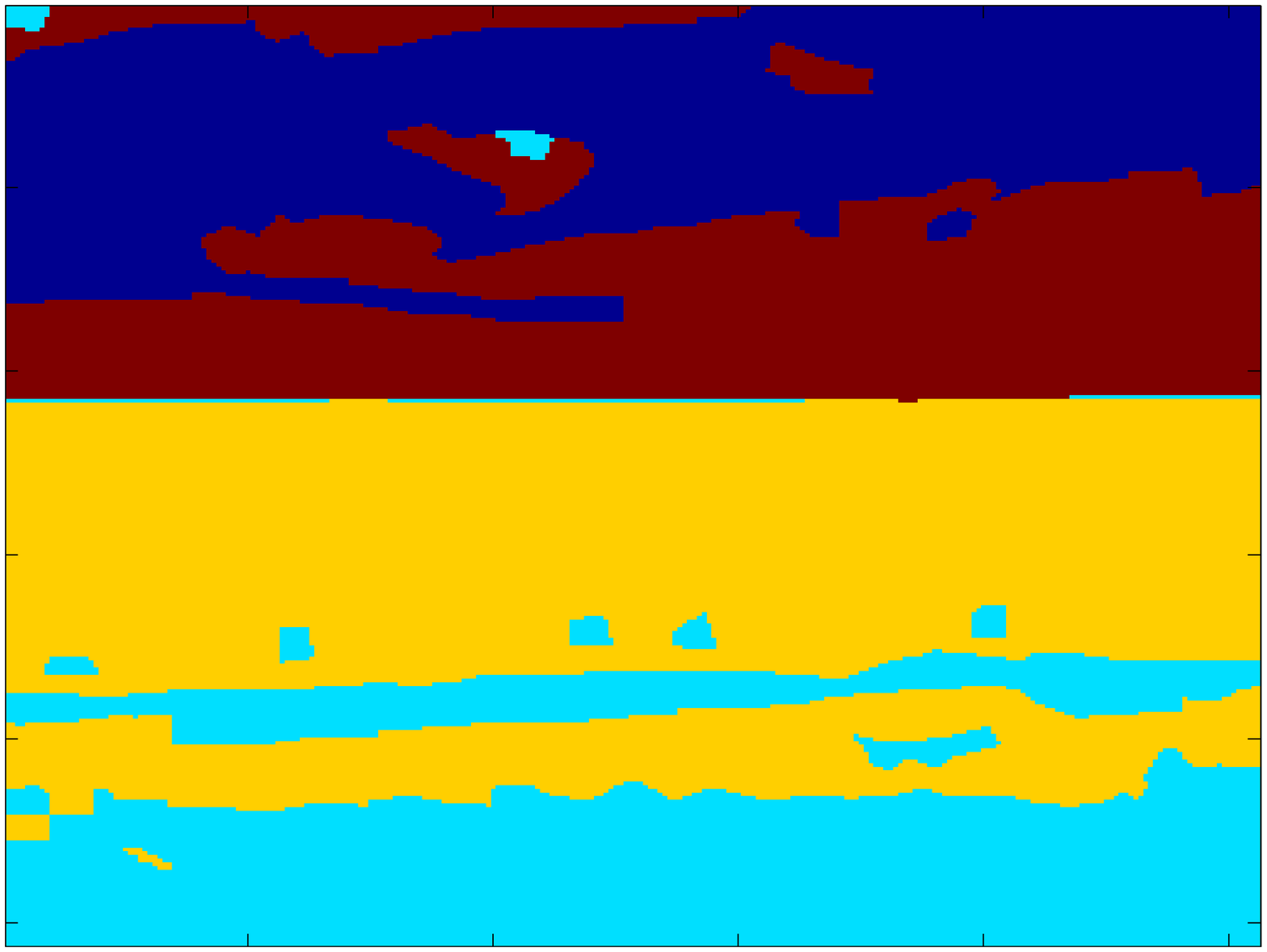}
\includegraphics[width=0.15\textwidth]{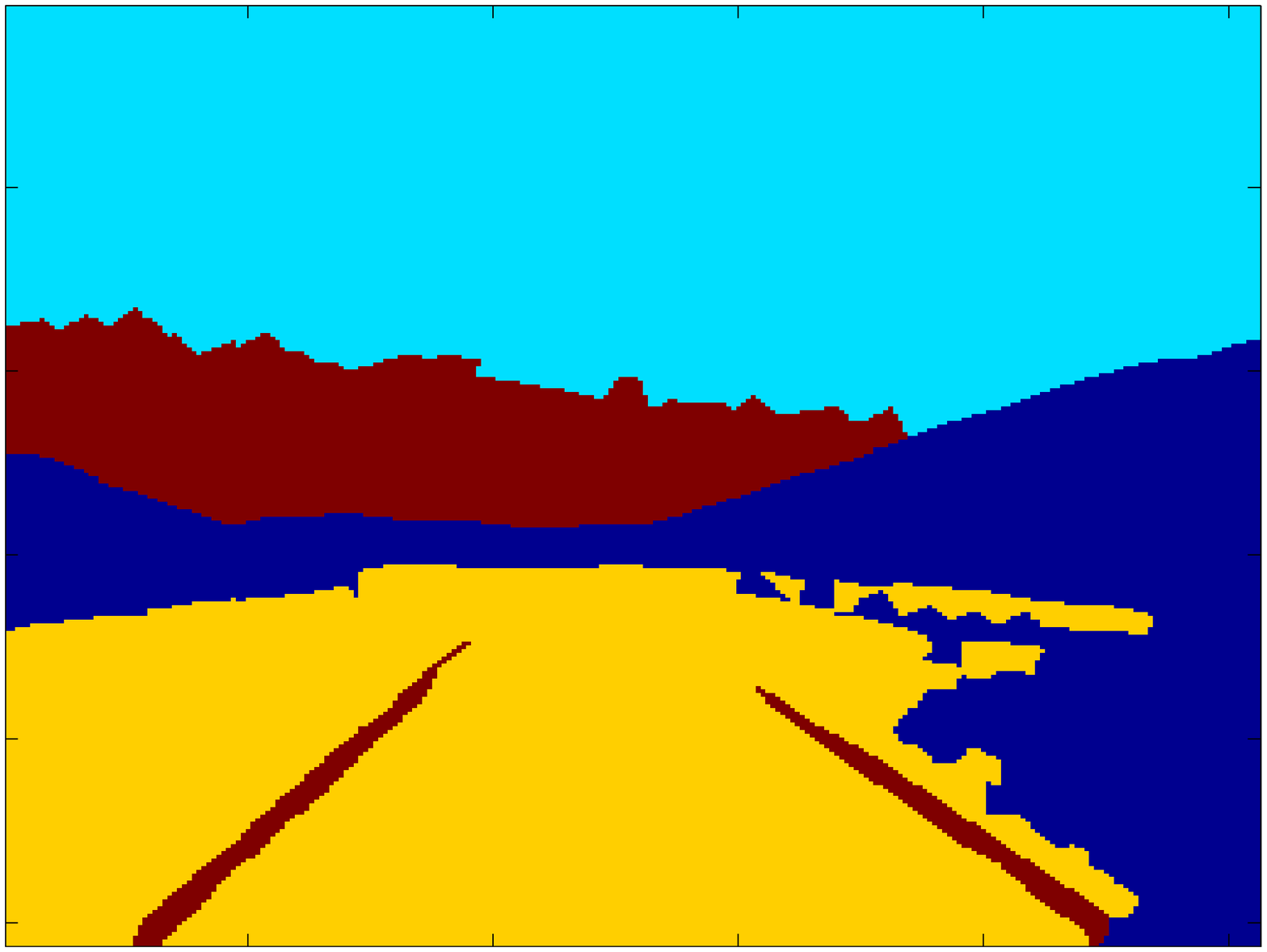}
\includegraphics[width=0.15\textwidth]{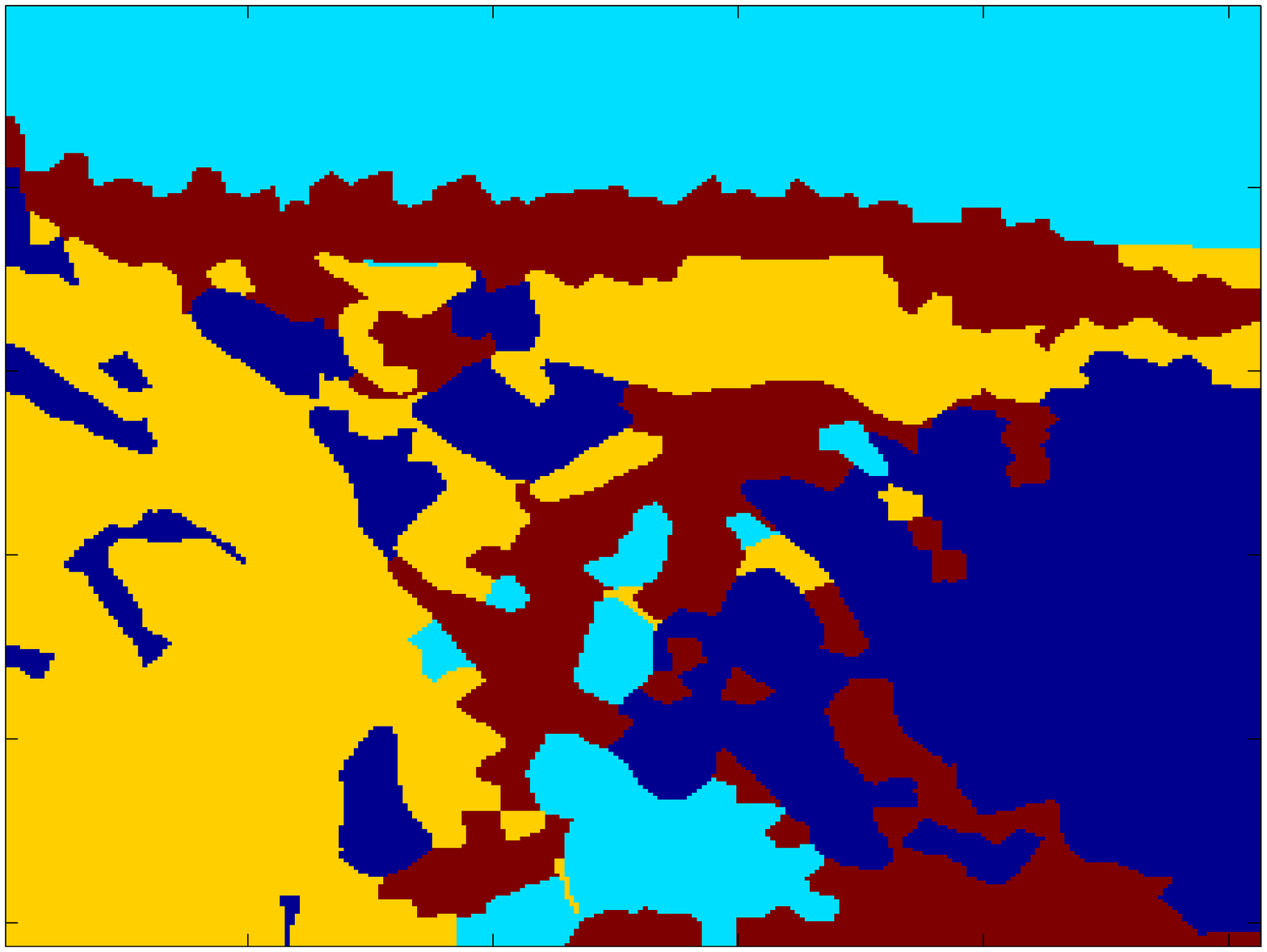}
\includegraphics[width=0.15\textwidth]{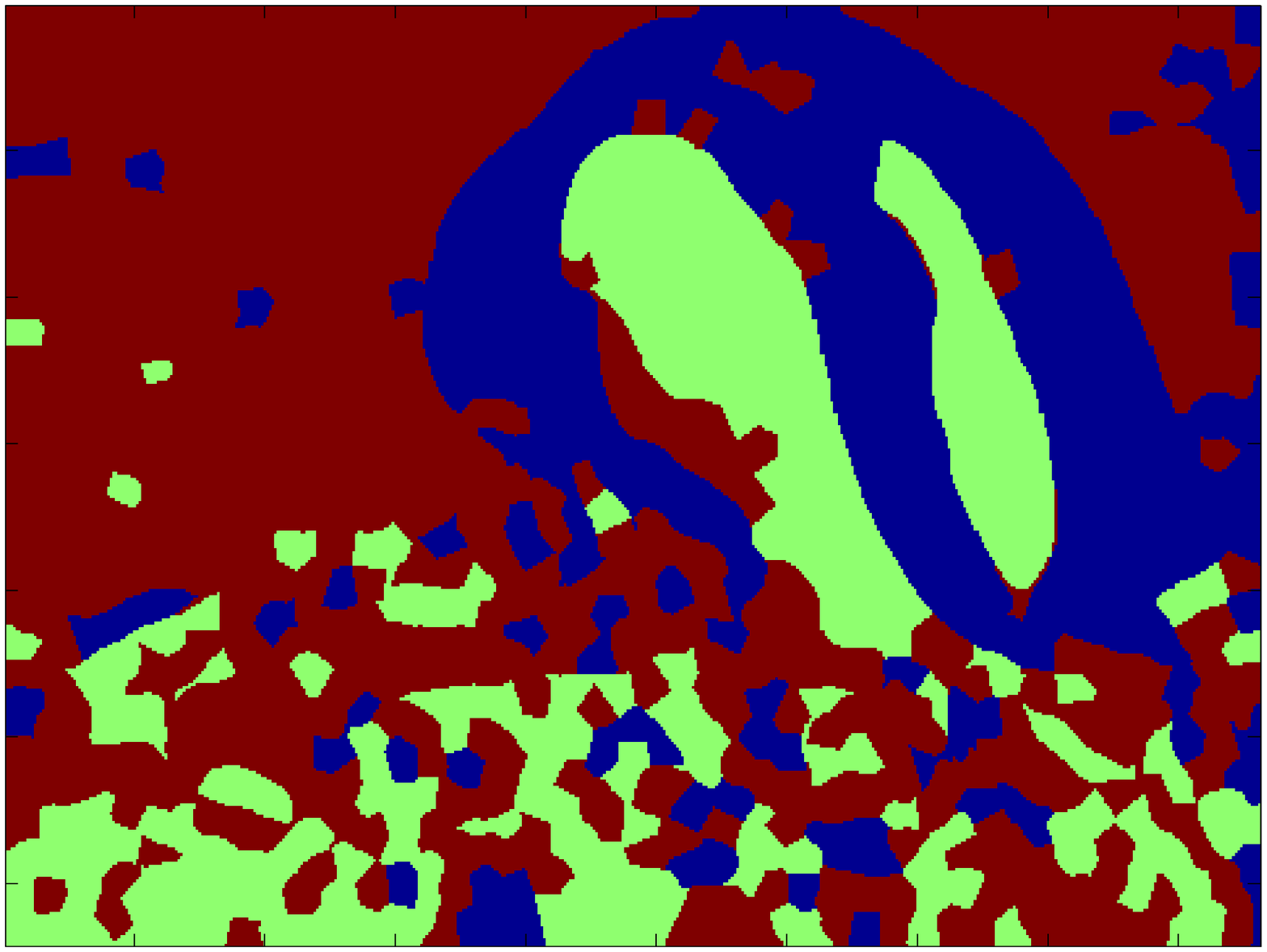}
\includegraphics[width=0.15\textwidth]{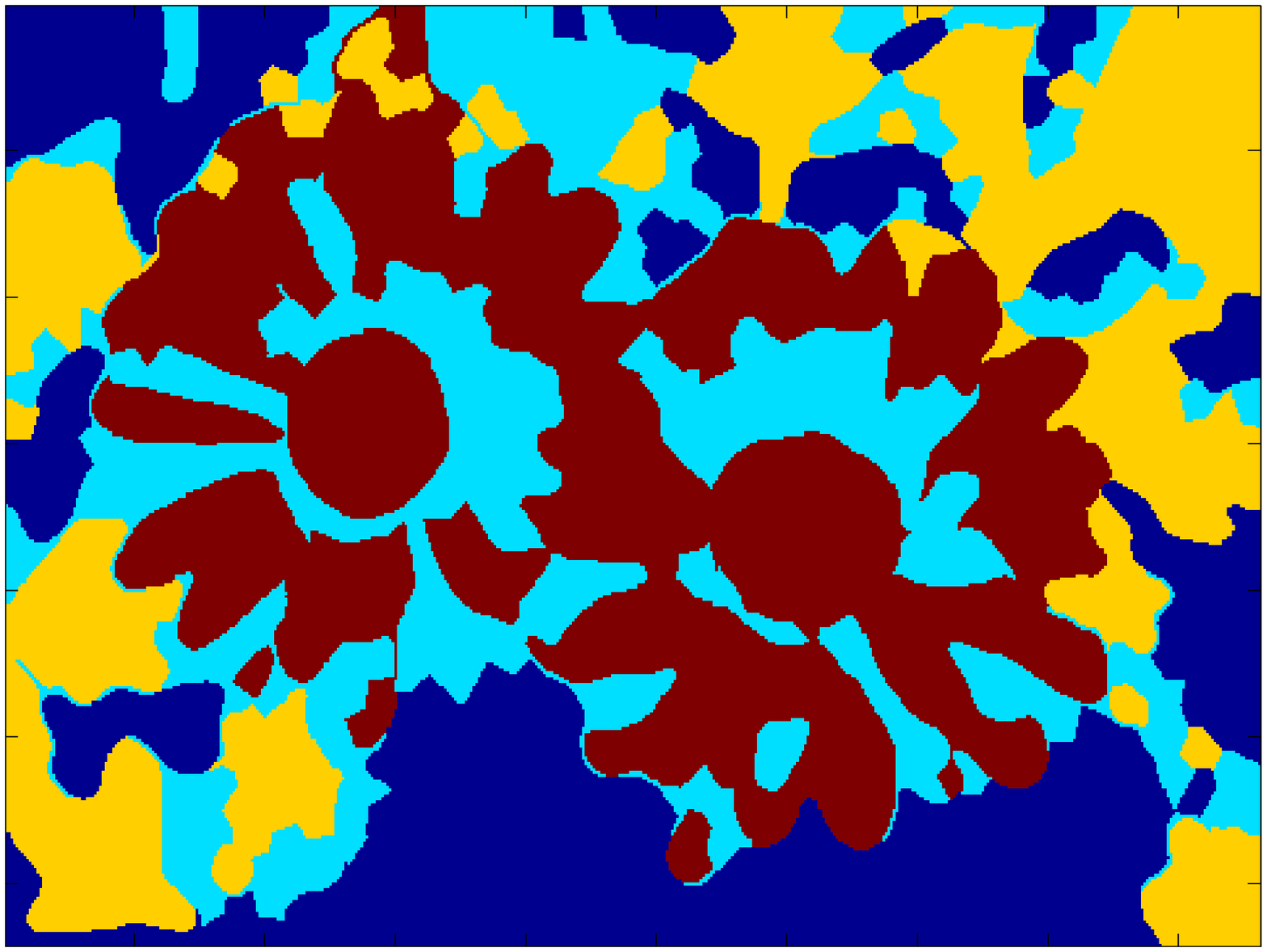}
\includegraphics[width=0.15\textwidth]{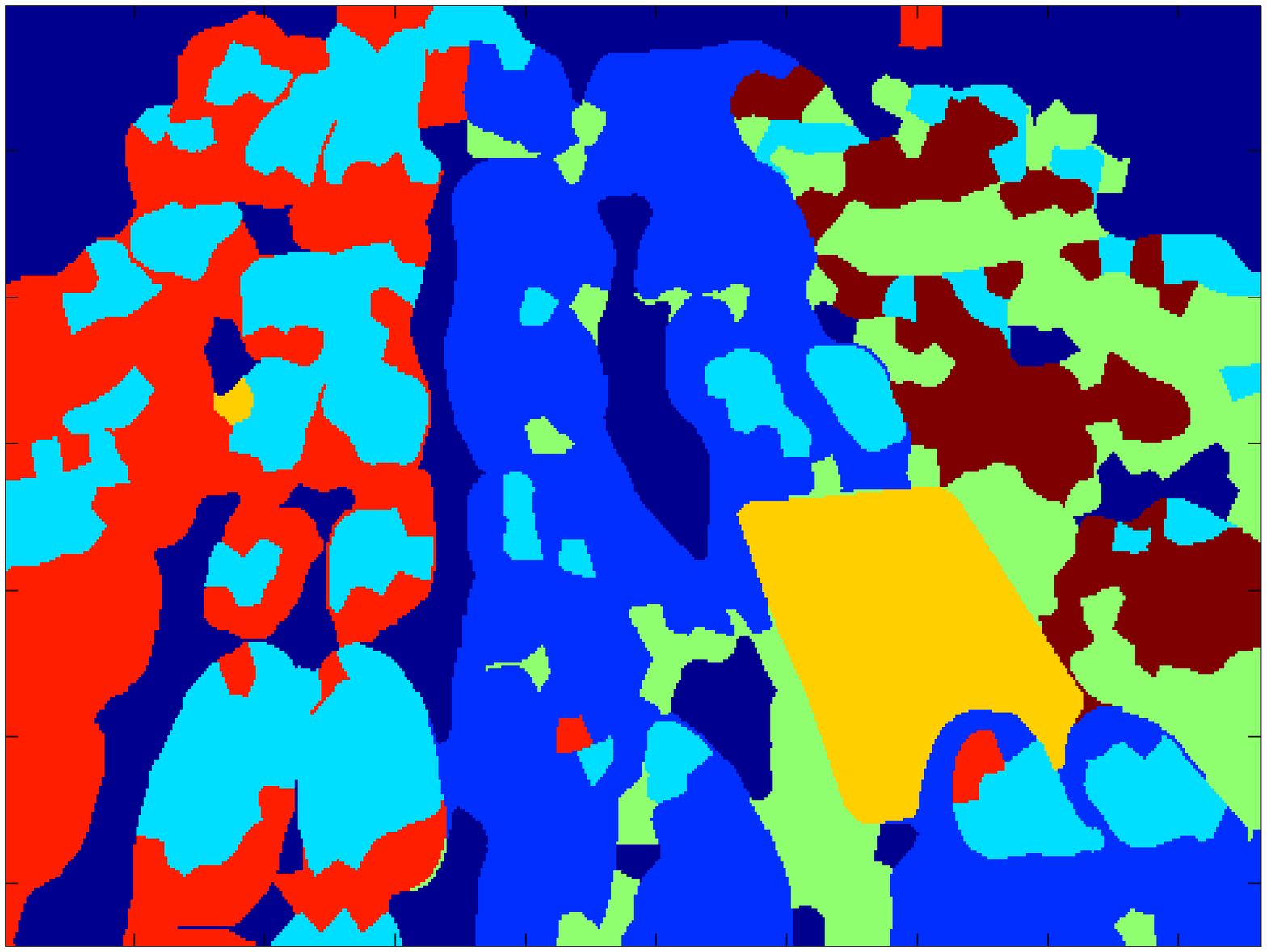}\\
 \includegraphics[width=0.15\textwidth]{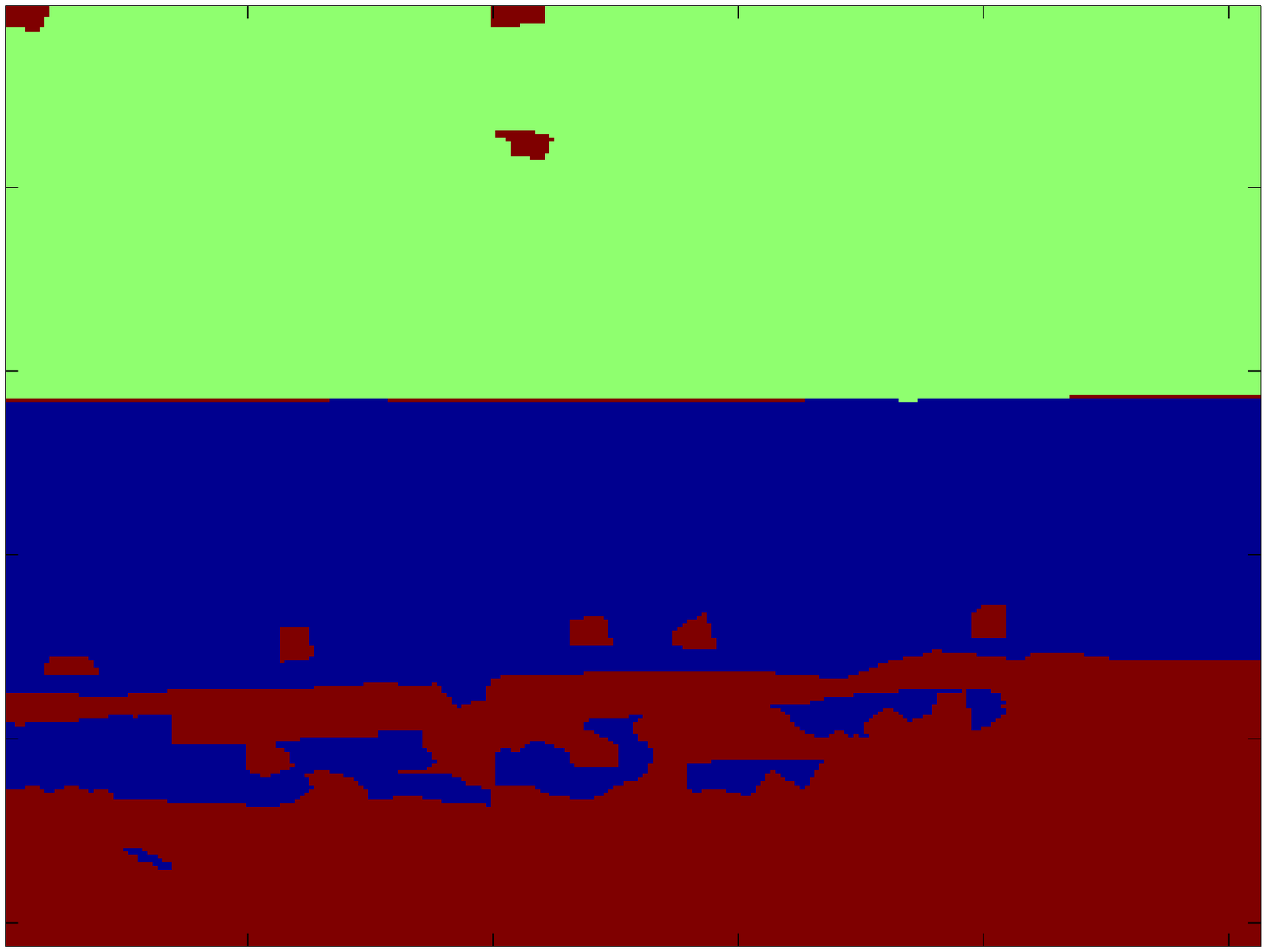}
\includegraphics[width=0.15\textwidth]{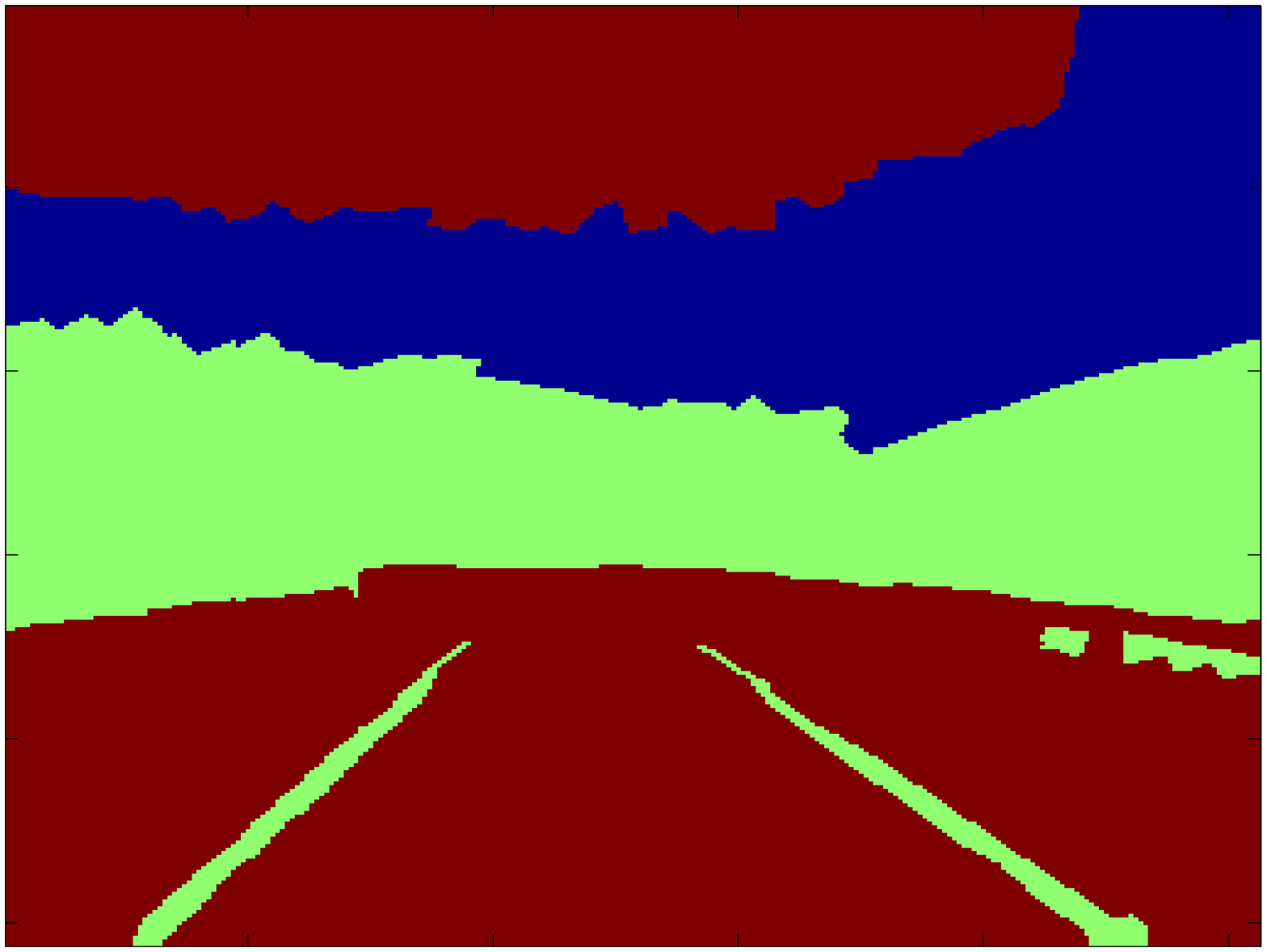}
\includegraphics[width=0.15\textwidth]{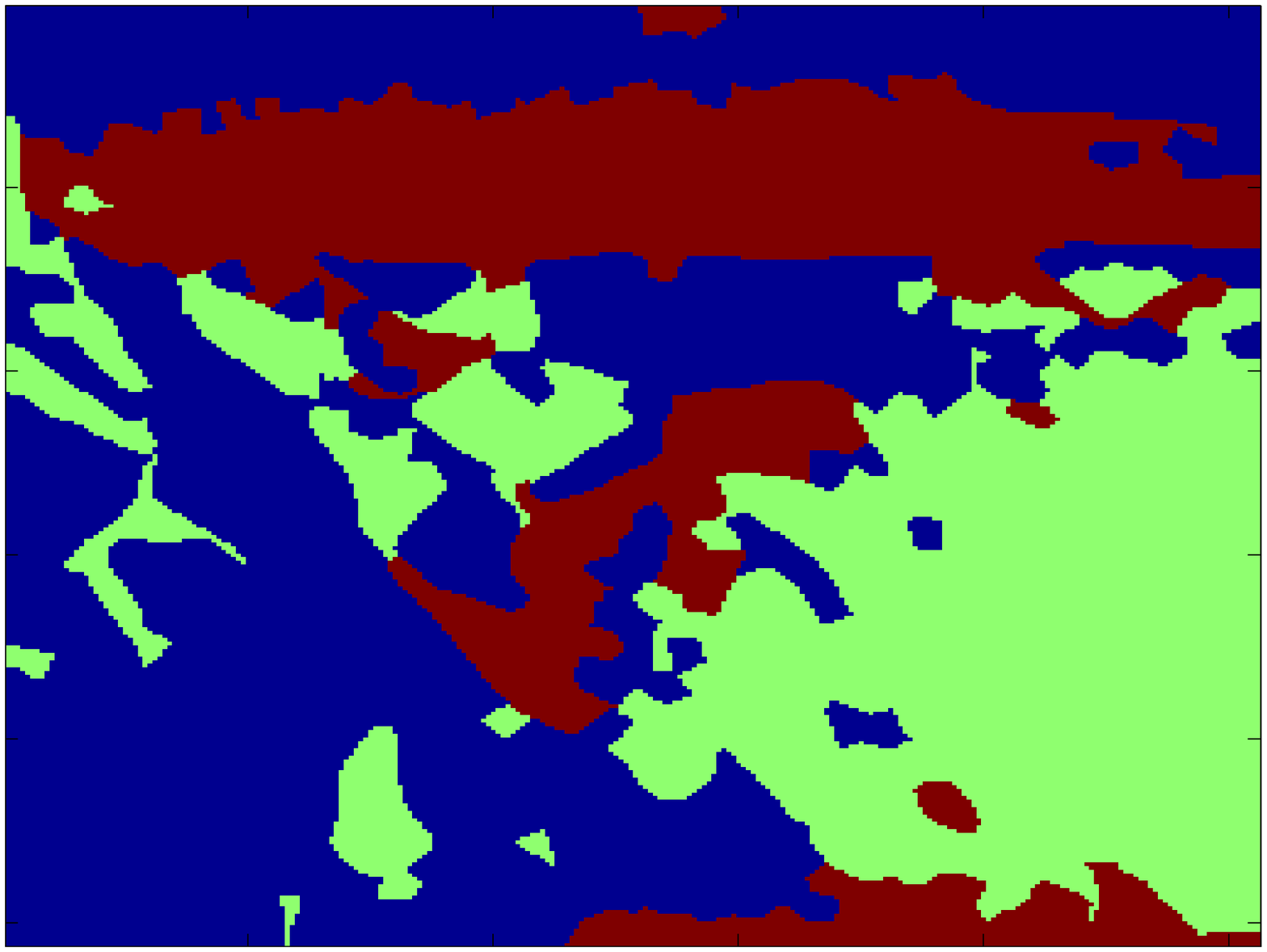}
\includegraphics[width=0.15\textwidth]{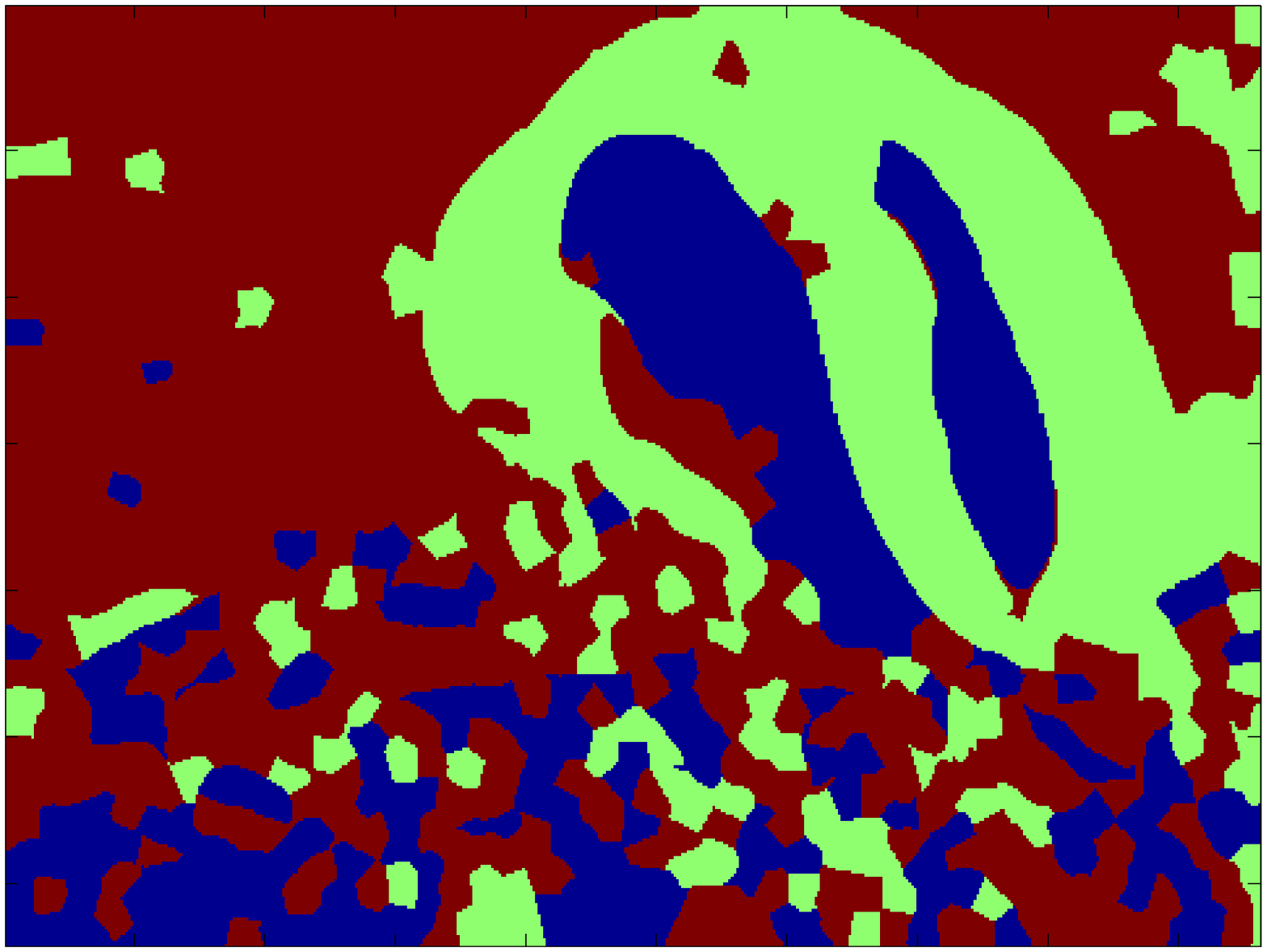}
\includegraphics[width=0.15\textwidth]{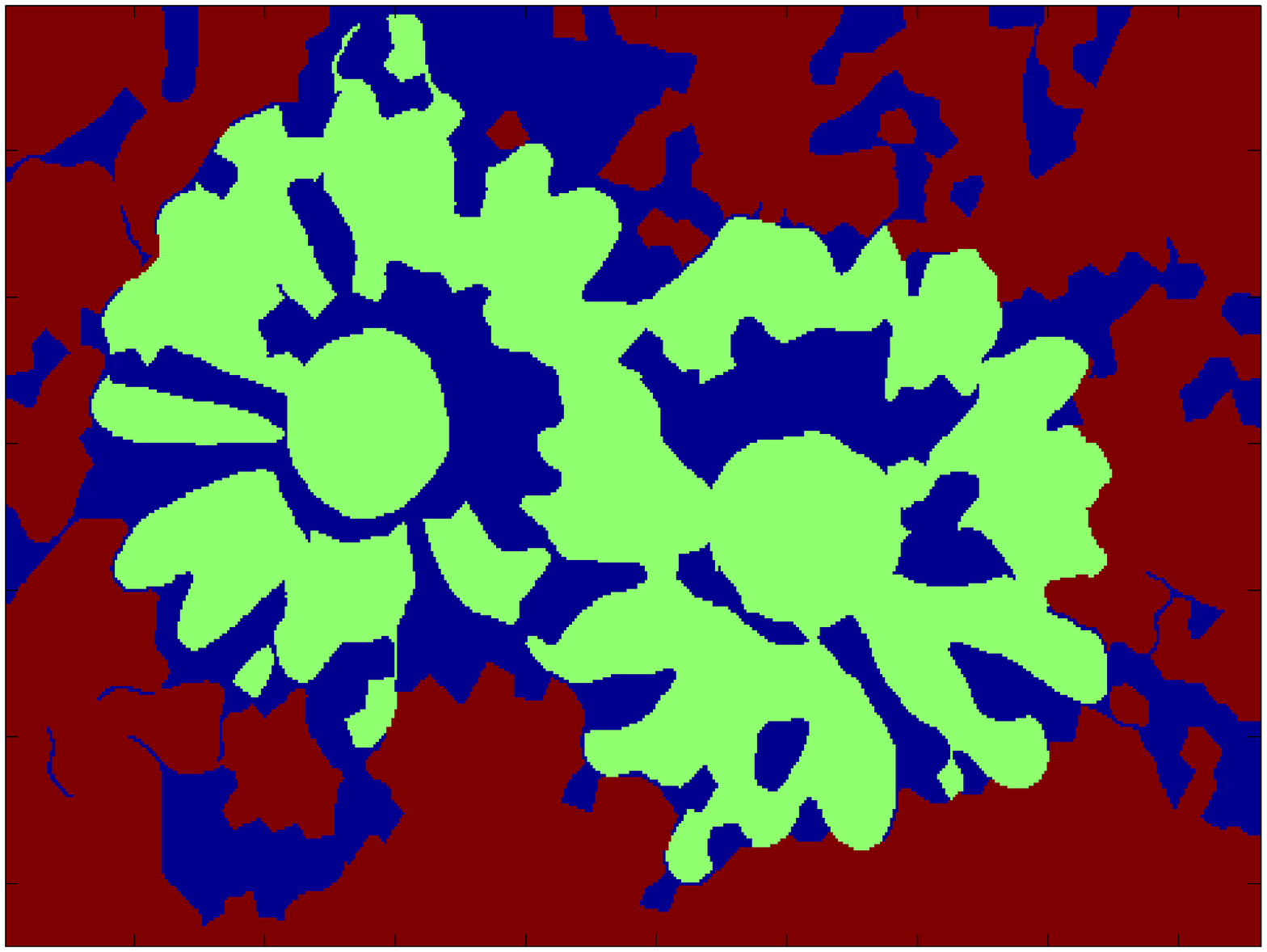}
\includegraphics[width=0.15\textwidth]{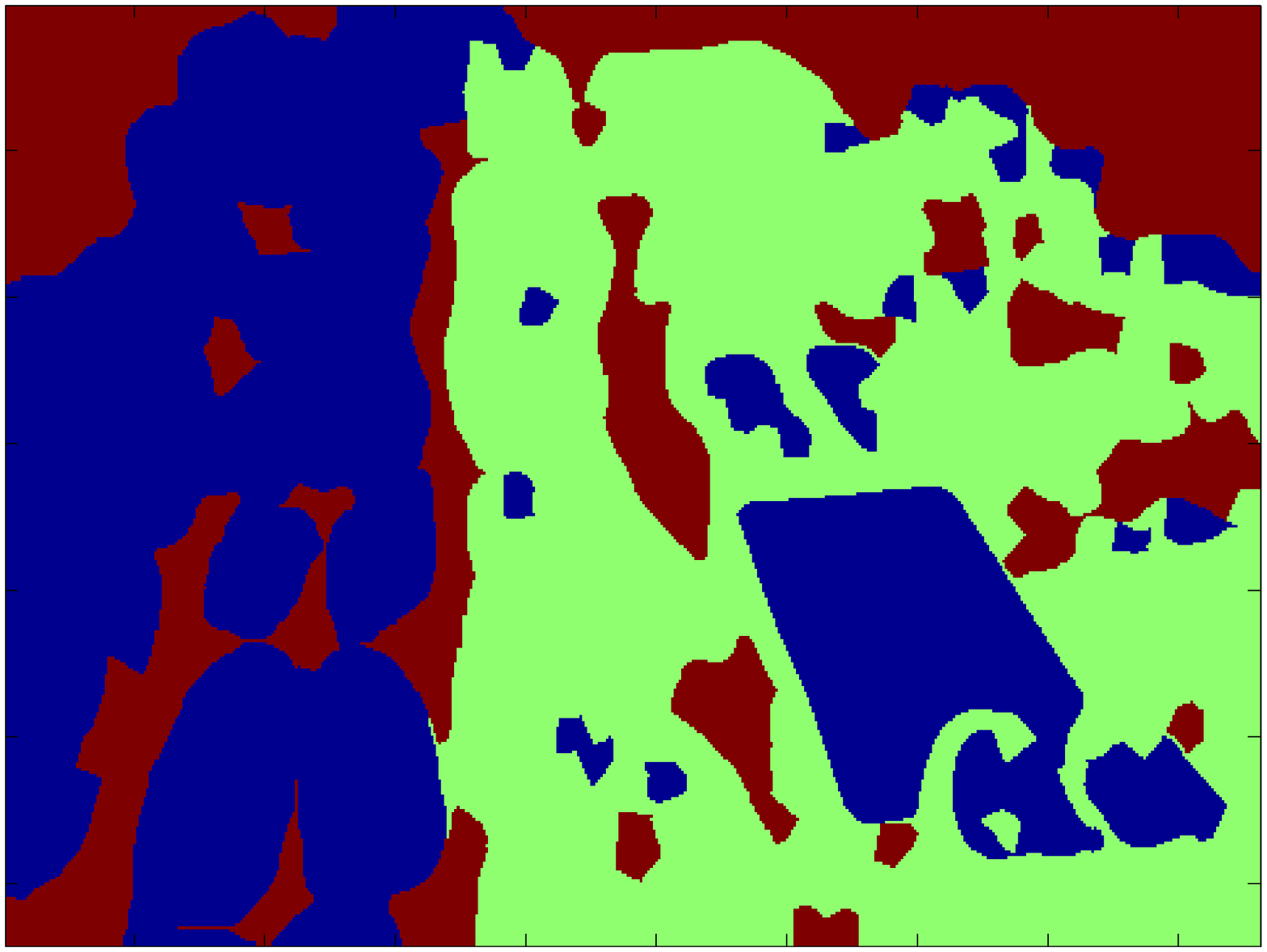}\\

\par\end{centering}

\caption{Segmentation results on six different images from the LabelMe and
Berkeley datasets. From top to bottom row: Image, Human segmentation,
Mean-shift, Hierarchical distance-dependent Chinese Restaurant Process,
Potts-Dirichlet Process, Truncated Potts-Dirichlet Process with $T_{\text{min}}=3,6,9$.}

\label{fig:compare_image2}
\end{figure}

As we expect the number of clusters to be around ten, we evaluate
the Potts-DP~\footnote{Experiments were also performed with the two parameter Poisson-Dirichlet, but without observed improvement on the performances.} and the truncated Potts-DP models with $\alpha=3$ and
$\beta=0.02$ in agreement with Figure \ref{fig:simus}. We compare
our results to mean-shift~\cite{comaniciu2002mean} and rdd-CRP~\cite{Ghosh2011}.
We tested all these methods by randomly selecting 50 images from each
of 8 categories in the LabelMe dataset (i.e. on a total of 400 images)
and 200 images from the Berkeley dataset. We display the obtained
results on six particular images in Figure \ref{fig:compare_image2}.
As expected the number of clusters decreases as $T_{\text{min}}$
increases.

\begin{table}
\caption{{\new Comparison of the Potts-DP model for various values of $T_{\text{min}}$
with the benchmark algorithms: mean-shift and rdd-CRP in terms of
the average rand index between the ground-truth label for 400 images
on the LabelMe dataset and 200 images on the Berkeley dataset. The
parameters are set to $\alpha=3$, $\beta=0.02$, $\lambda=10$, and
different threshold values. For each dataset, the first line corresponds
to the median value of the rand index for each model. The second line
represents the p-value of the Wilcoxon sign rank test that the dataset
with higher value has a median value different from the others. In
bold are shown the methods with the best results, at a 1\% significance
level. The third line represents the median number of clusters.}}

\label{fig:compare_rand}

\centering{}%
\begin{tabular}{cc||c||c||cccc}
\hline
 &  & MS  & rdd-CRP  & \multicolumn{4}{c}{(Truncated) Potts-DP}\tabularnewline
 &  &  &  & $T_{\text{min}}=0$  & $T_{\text{min}}=3$  & $T_{\text{min}}=6$  & $T_{\text{min}}=9$ \tabularnewline
\hline
\hline
LabelMe  & Med. RI  & 0.7623  & \textbf{\textcolor{red}{0.7759}}  & \textbf{\textcolor{red}{0.7712}}  & \textbf{\textcolor{red}{0.7692}}  & 0.7483  & 0.7235 \tabularnewline
 & p-value  & .0011  & --  & .0108  & .0965  & $<.001$  & $<.001$ \tabularnewline
 & Med. Nb clust.  & 12  & 6  & 9  & 6  & 4  & 3 \tabularnewline
\hline
Berkeley  & Med. RI  & \textbf{\textcolor{red}{0.7988}}  & 0.7748  & \textbf{\textcolor{red}{0.7882}}  & 0.7797  & 0.7291  & 0.6881 \tabularnewline
 & p-value  & --  & .0066  & .0566  & .0052  & $<.001$  & $<.001$ \tabularnewline
 & Med. Nb clust.  & 23  & 9  & 12  & 8  & 4  & 3 \tabularnewline
\hline
\end{tabular}
\end{table}


To assess the quality of the image segmentation results, we use the
rand index \cite{hubert1985comparing} computed using the ``ground-truth\textquotedblright{}\ which
is obtained through a manual labelling \cite{russell2008labelme}. This comparison was also performed
in \cite{Ghosh2011} and the results are presented in Table \ref{fig:compare_rand}.
To evaluate the statistical significance of the results, we performed
a Wilcoxon signed-rank test between the method with highest rand index
(rdd-CRP for LabelMe and Mean-Shift for Berkeley dataset) and the
other methods. We found no statistically significant difference (at
the 1\% level) between the performances of rdd-CRP, Potts-DP and truncated
Potts-DP on the LabelMe dataset, and no statistical difference between
Mean-Shift and Potts-DP on the Berkeley dataset.

Despite the significant differences observed visually when increasing
the truncation threshold, e.g. see Figure~\ref{fig:compare_image2},
this does not translate in any improvement from the rand index point
of view. However the manual labelling appears fairly subjective, so
the rand index and the associated results have to be interpreted carefully.

\section{Discussion}

This paper has introduced an original BNP image segmentation model
that allows us to easily introduce prior information so as to penalize
the overall number and size of clusters while preserving a spatial
smoothing component. Computationally we have shown that Bayesian inference
can be carried out using a GSW sampler which explores the posterior
distribution of interest by splitting and merging clusters. Experimentally,
the image segmentation results we obtained using a truncated Potts-DP\ model
are competitive to mean-shift and rdd-CRP. We believe that it is a
promising approach that deserves further investigation even if the
model has limits inherent to the use of a spatial Potts prior: it
penalizes the overall number and size of clusters but not connected
components, and may end up with some isolated components. Nonetheless,
there is always a tradeoff between goodness of fit of the model and
computational tractability: the proposed BNP model has the ability
to control the overall clustering structure, while the associated
GSW sampler is easy to put in practice and allows experimentally a
good exploration of the posterior. Furthermore, in the context of
the standard Potts model, various improvements over the algorithm
of \cite{Higdon1998} have been proposed by \cite{Barbu2005,Barbu2007}.
In particular \cite{Barbu2007} propose a careful selection of the
auxiliary parameters (\ref{eq:auxiliaryparameters}), various sophisticated
reversible jump MCMC moves to swap labels and multi-level approaches.
They report visually impressive segmentation results and it is likely
that developing similar type ideas for the BNP segmentation model
proposed here would further improve performance.

\bibliographystyle{plain}
\bibliography{SWPottsDP.bbl}

\begin{thebibliography}{10}

\bibitem{Barbu2005}
A.~Barbu and S.C. Zhu.
\newblock Generalizing {Swendsen-Wang} to sampling arbitrary posterior
  probabilities.
\newblock {\em IEEE Transactions on Pattern Analysis and Machine Intelligence},
  27(8):1239--1253, 2005.

\bibitem{Barbu2007}
A.~Barbu and S.C. Zhu.
\newblock Generalizing {Swendsen-Wang} for image analysis.
\newblock {\em Journal of Computational and Graphical Statistics},
  16(4):877--900, 2007.

\bibitem{comaniciu2002mean}
D.~Comaniciu and P.~Meer.
\newblock Mean shift: A robust approach toward feature space analysis.
\newblock {\em IEEE Transactions on Pattern Analysis and Machine Intelligence},
  24(5):603--619, 2002.

\bibitem{Dahl2005}
David~B Dahl.
\newblock Sequentially-allocated merge-split sampler for conjugate and
  nonconjugate {D}irichlet process mixture models.
\newblock {\em Journal of Computational and Graphical Statistics}, 11, 2005.

\bibitem{Du2009}
L.~Du, L.~Ren, D.~Dunson, and L.~Carin.
\newblock A {B}ayesian model for simultaneous image clustering, annotation and
  object segmentation.
\newblock In {\em Advances in Neural Information Processing Systems}, 2009.

\bibitem{Duan2007}
J.A. Duan, M.~Guindani, and A.E. Gelfand.
\newblock Generalized spatial {D}irichlet process models.
\newblock {\em Biometrika}, 94(4):809--825, 2007.

\bibitem{Edwards1988}
R.~E. Edwards and A.~D. Sokal.
\newblock Generalization of the {Fortuin-Kasteleyn-Swendsen-Wang}
  representation and {M}onte {C}arlo algorithm.
\newblock {\em Physical review D}, 38(6):2009--2012, 1988.

\bibitem{Fortuin1972}
C.M. Fortuin and P.W. Kasteleyn.
\newblock On the random-cluster model:: I. introduction and relation to other
  models.
\newblock {\em Physica}, 57(4):536--564, 1972.

\bibitem{Geman1984}
S.~Geman and D.~Geman.
\newblock Stochastic relaxation, {G}ibbs distributions, and the {B}ayesian
  restoration of images.
\newblock {\em IEEE Transactions on Pattern Analysis and Machine Intelligence},
  (6):721--741, 1984.

\bibitem{Ghosh2012}
S.~Ghosh and E.B. Sudderth.
\newblock Nonparametric learning for layered segmentation of natural images.
\newblock In {\em Computer Vision and Pattern Recognition (CVPR), 2012 IEEE
  Conference on}, pages 2272--2279. IEEE, 2012.

\bibitem{Ghosh2011}
S.~Ghosh, A.B. Ungureanu, E.B. Sudderth, and D.M. Blei.
\newblock Spatial distance dependent chinese restaurant processes for image
  segmentation.
\newblock In {\em Advances in Neural Information Processing Systems}, 2011.

\bibitem{Green1995}
P.J. Green.
\newblock Reversible jump {M}arkov chain {M}onte {C}arlo computation and
  {B}ayesian model determination.
\newblock {\em Biometrika}, 82:711--732, 1995.

\bibitem{Green2002}
P.J. Green and S.~Richardson.
\newblock Hidden {M}arkov models and disease mapping.
\newblock {\em Journal of the American Statistical Association}, 97:1055--1070,
  2002.

\bibitem{Higdon1998}
D.M. Higdon.
\newblock Auxiliary variable methods for markov chain monte carlo with
  applications.
\newblock {\em Journal of the American Statistical Association}, pages
  585--595, 1998.

\bibitem{hubert1985comparing}
L.~Hubert and P.~Arabie.
\newblock Comparing partitions.
\newblock {\em Journal of classification}, 2(1):193--218, 1985.

\bibitem{Hughes2012}
Michael Hughes, Emily Fox, and Erik Sudderth.
\newblock Effective split-merge monte carlo methods for nonparametric models of
  sequential data.
\newblock In {\em Advances in Neural Information Processing Systems}, pages
  1304--1312, 2012.

\bibitem{Jain2004}
S.~Jain and R.~Neal.
\newblock A split-merge markov chain monte carlo procedure for the dirichlet
  process mixture model.
\newblock {\em Journal of Computational and Graphical Statistics}, 13(1), 2004.

\bibitem{Lau2006}
J.~Lau and P.J. Green.
\newblock {B}ayesian model based clustering procedures.
\newblock {\em Journal of Computational and Graphical Statistics}, 16:526--558,
  2007.

\bibitem{MartinFTM01}
D.~Martin, C.~Fowlkes, D.~Tal, and J.~Malik.
\newblock A database of human segmented natural images and its application to
  evaluating segmentation algorithms and measuring ecological statistics.
\newblock In {\em Proc. 8th Int'l Conf. Computer Vision}, volume~2, pages
  416--423, July 2001.

\bibitem{mori2004recovering}
Greg Mori, Xiaofeng Ren, Alexei~A Efros, and Jitendra Malik.
\newblock Recovering human body configurations: Combining segmentation and
  recognition.
\newblock In {\em Computer Vision and Pattern Recognition, 2004. CVPR 2004.
  Proceedings of the 2004 IEEE Computer Society Conference on}, volume~2, pages
  II--326. IEEE, 2004.

\bibitem{Orbanz2008}
P.~Orbanz and J.~M. Buhmann.
\newblock Nonparametric {B}ayesian image segmentation.
\newblock {\em International Journal of Computer Vision}, 77:25--45, 2008.

\bibitem{Pitman1995}
J.~Pitman.
\newblock Exchangeable and partially exchangeable random partitions.
\newblock {\em Probability theory and related fields}, 102:145--158, 1995.

\bibitem{Ren2003}
X.~Ren and J.~Malik.
\newblock Learning a classification model for segmentation.
\newblock In {\em IEEE International Conference on Computer Vision, 2003},
  2003.

\bibitem{russell2008labelme}
B.C. Russell, A.~Torralba, K.P. Murphy, and W.T. Freeman.
\newblock Labelme: a database and web-based tool for image annotation.
\newblock {\em International journal of computer vision}, 77(1):157--173, 2008.

\bibitem{Sudderth2009}
E.B. Sudderth and M.I. Jordan.
\newblock Shared segmentation of natural scenes using dependent {Pitman-Yor}
  processes.
\newblock In {\em Advances in Neural Information Processing Systems},
  volume~21, pages 1585--1592, 2009.

\bibitem{Swendsen1987}
R.H. Swendsen and J.-S. Wang.
\newblock Nonuniversal critical dynamics in {M}onte {C}arlo simulations.
\newblock {\em Physical Review Letters}, 58:86--88, 1987.

\bibitem{Winkler2003}
G.~Winkler.
\newblock {\em Image analysis, random fields and Markov chain Monte Carlo
  methods: a mathematical introduction}, volume~27.
\newblock Springer Verlag, 2003.

\end{thebibliography}

\end{document}